\documentclass[10pt,twocolumn,letterpaper]{article}

\usepackage[pagenumbers]{wacv} 

\usepackage{multirow}
\usepackage{colortbl}

\newcommand{\ppm}{\,\scriptsize$\pm$}

\definecolor{wacvblue}{rgb}{0.21,0.49,0.74}
\usepackage[pagebackref,breaklinks,colorlinks,allcolors=wacvblue]{hyperref}

\def\wacvPaperID{2622} 
\def\confName{WACV}
\def\confYear{2027}

\title{Towards Continual Test-Time Adaptation of Vision-Language Models in Open-Vocabulary Semantic Segmentation}

\author{Chandler Timm C. Doloriel$^{1}$, Yunbei Zhang$^{2}$, Sarthak Kumar Maharana$^{3}$, Muhammad Salman Siddiqui$^{1}$,\\
Tor Kristian Stevik$^{1}$, Fadi Al Machot$^{1}$, Kristian Hovde Liland$^{1}$, Habib Ullah$^{1}$\\
$^{1}$Faculty of Science and Technology (REALTEK), Norwegian University of Life Sciences (NMBU)\\
$^{2}$Tulane University, $^{3}$University of Texas at Dallas\\
{\tt\small chandler.timm.cagmat.doloriel@nmbu.no}
}

\begin{document}
\maketitle

\begin{abstract}
Open-vocabulary semantic segmentation (OVSS) relies on vision-language alignment to recognize arbitrary text-defined categories, yet this alignment is fragile under continual test-time distribution shift. Our diagnostic analysis reveals that entropy minimization drives patch-level class collapse, continual updates erode vision-language alignment, and redundant gradients from low-shift samples waste computation. We propose Diversify, Anchor, and Filter (DAF), a stabilization framework that augments entropy-based adaptation with a marginal diversity loss that resists collapse, a cross-modal anchor consistency loss that constrains feature drift relative to a frozen source model, and feature salience filtering that skips low-value backward passes to offset part of the source-anchor overhead. We evaluate on five datasets spanning natural scenes, autonomous driving, underwater imagery, and remote sensing with their corrupted variants. Across the evaluated continual shifts, DAF remains stable where entropy minimization collapses, improving mIoU by over 8 points on Pascal VOC20-C, over 9 points on LoveDA, and over 3 points on Foggy Cityscapes compared to the source model, and is robust to aggressive adaptation and learning rate choices.
\end{abstract}

\section{Introduction}
\label{sec:intro}
Contrastive vision-language models (VLMs) such as CLIP~\cite{Radford2021LearningTV} have become strong backbones for open-vocabulary semantic segmentation (OVSS) because they align dense visual features with text embeddings of class prompts. Recent OVSS methods such as SCLIP~\cite{Wang2023SCLIPRS} and NACLIP~\cite{Hajimiri2024PayAT} show that broad category vocabularies, extended beyond base classes with synonyms and subcategories, can be supported without task-specific retraining. Yet, this open-vocabulary generalization is fragile at test-time. Real deployments face weather, sensor variation, image corruptions, and underwater domain shifts~\cite{Konovalov2019UnderwaterFD,Zhao2022UnsupervisedAD,Zhao2023DMDnetAD} that distort patch-to-text similarities for dense prediction. Consider an autonomous driving model deployed across cities with varying weather and sensor conditions, where the vocabulary is defined by text prompts rather than a fixed set. Because collecting labeled target data or retraining offline is impractical, the model must adapt on-the-fly from incoming unlabeled samples alone.

Real deployment rarely involves a single isolated shift. OVSS systems face evolving domains, making continual test-time adaptation (CTTA) a natural setting~\cite{Wangetal2022cotta,maharana2026continual}. Unlike TTA, CTTA carries model state across domains without access to past data. This is especially difficult for OVSS because standard entropy-based methods such as TENT~\cite{Wang2021TentFT} and CoTTA~\cite{Wangetal2022cotta} were designed for closed-vocabulary settings and can erode the patch-to-text alignment needed for open-vocabulary prediction.

\begin{figure*}[htbp]
    \centering
    \begin{subfigure}{0.24\textwidth}
        \centering
        \includegraphics[width=\textwidth]{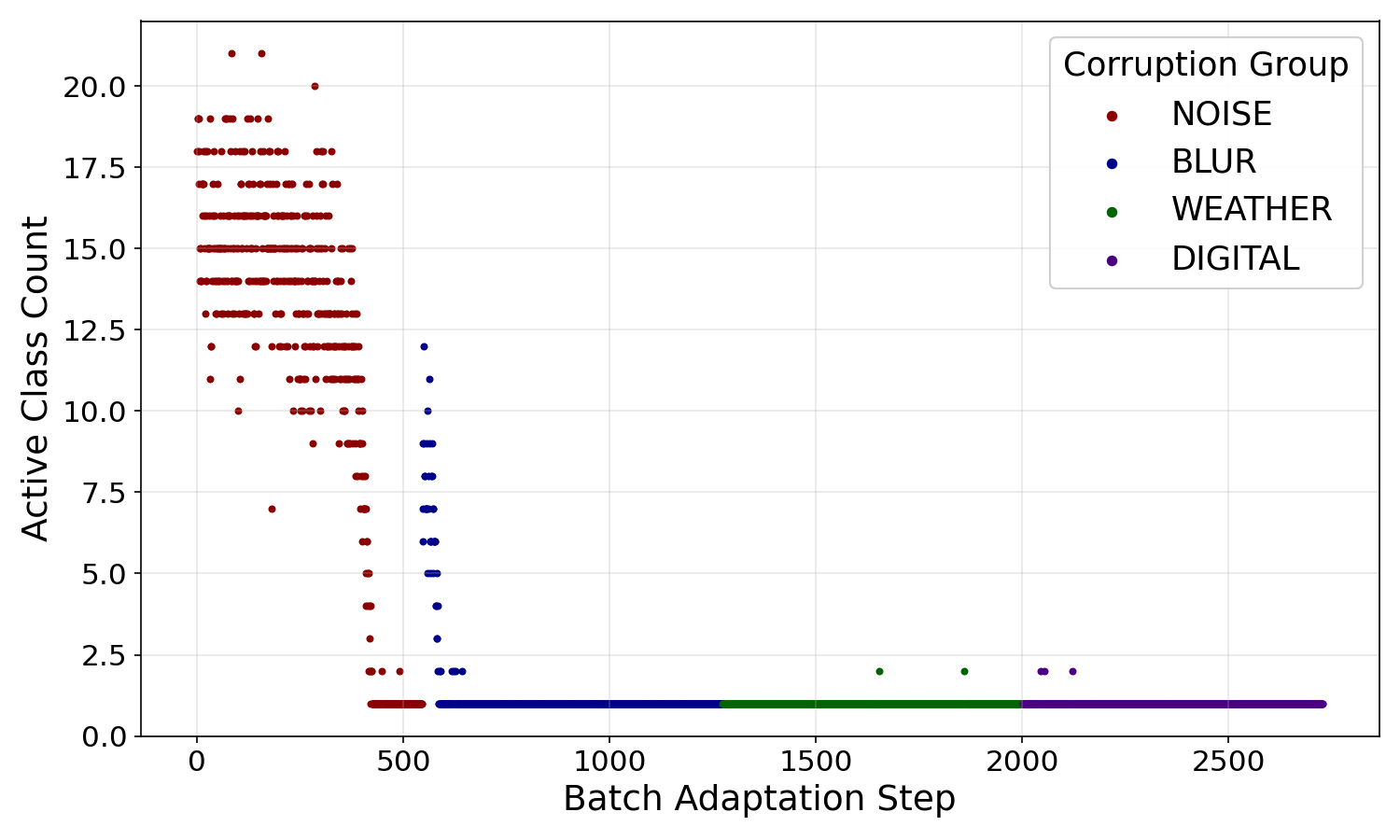}
        \caption{Class Collapse}
    \end{subfigure}
    \hfill
    \begin{subfigure}{0.24\textwidth}
        \centering
        \includegraphics[width=\textwidth]{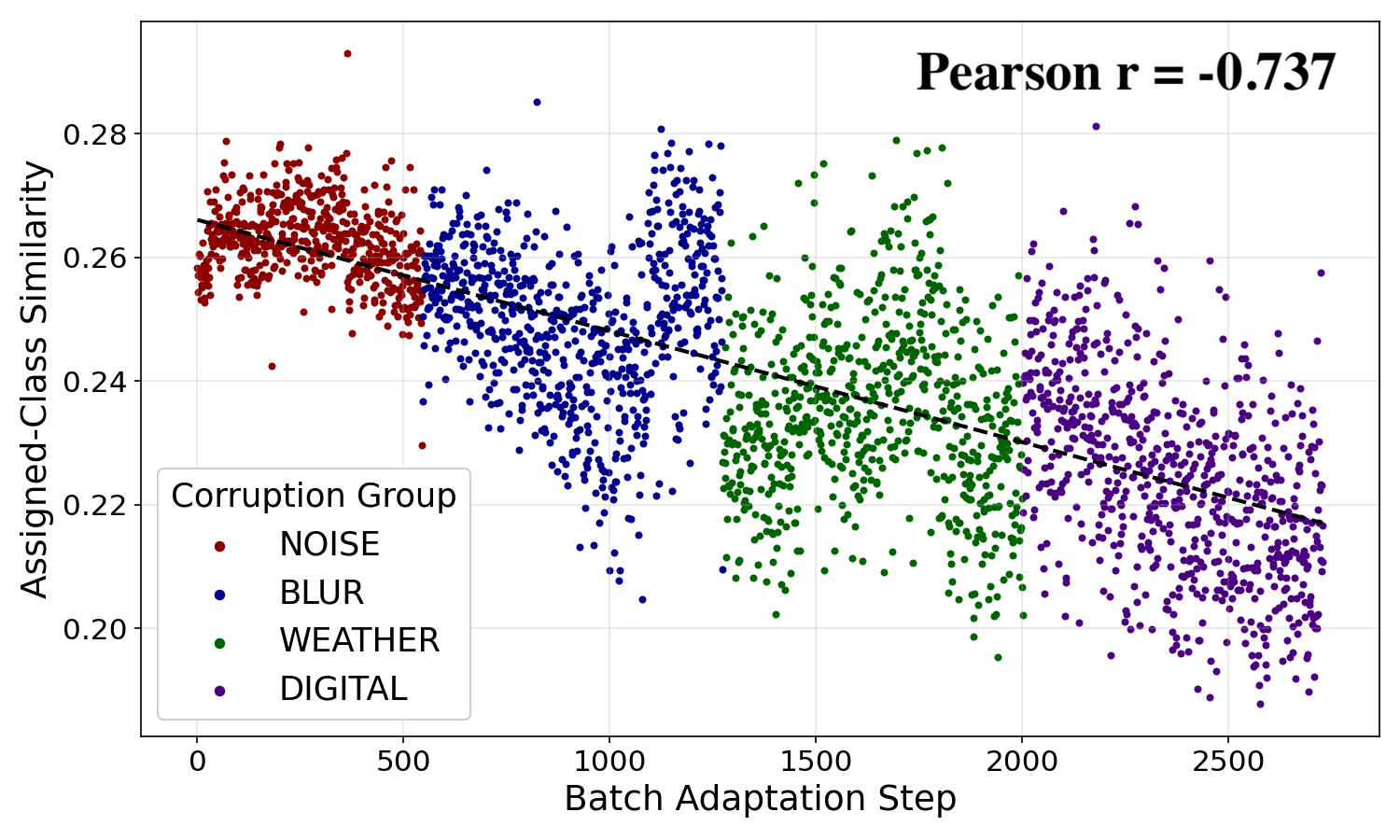}
        \caption{Assigned-Class Drift}
    \end{subfigure}
    \hfill
    \begin{subfigure}{0.24\textwidth}
        \centering
        \includegraphics[width=\textwidth]{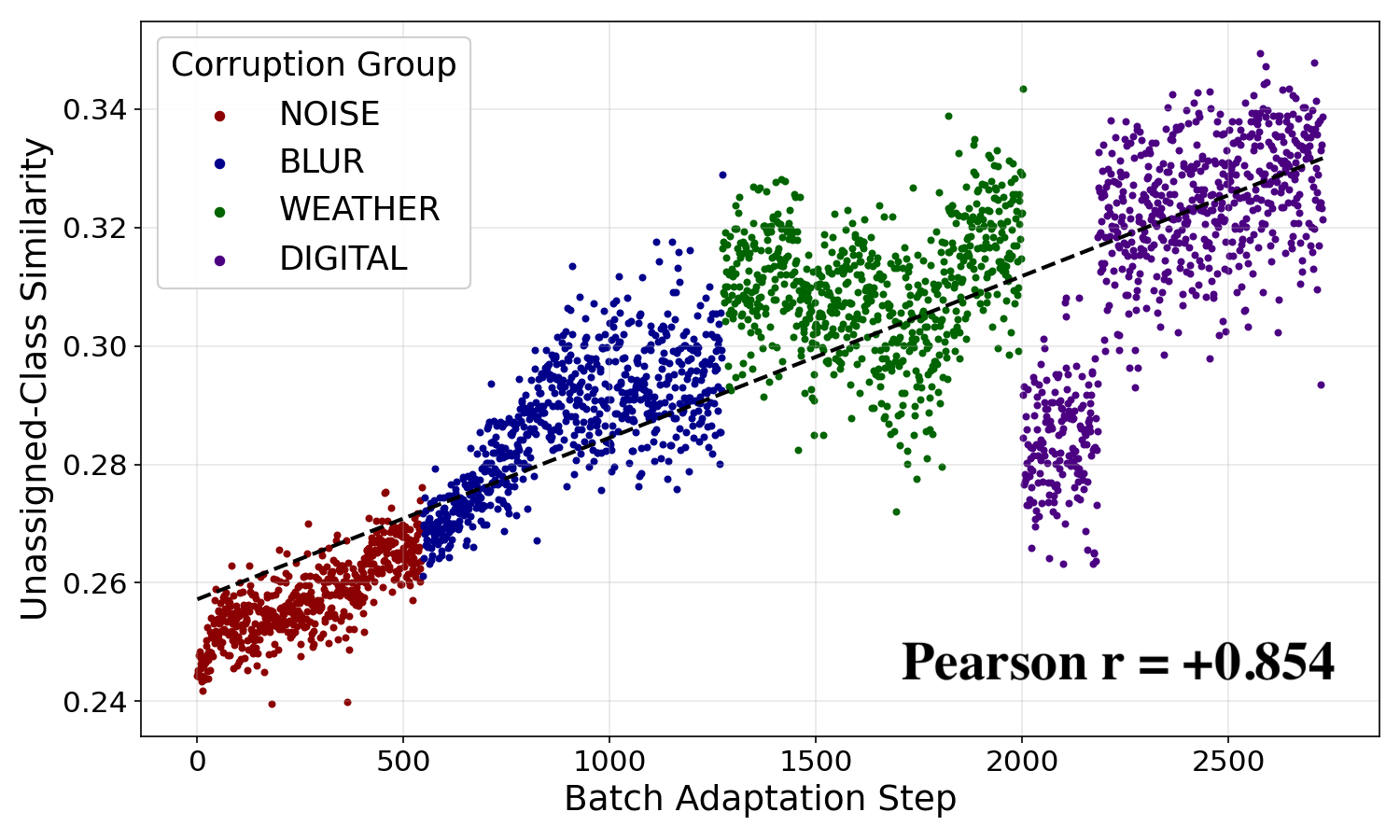}
        \caption{Unassigned-Class Drift}
    \end{subfigure}
    \hfill
    \begin{subfigure}{0.192\textwidth}
        \centering
        \includegraphics[width=\textwidth]{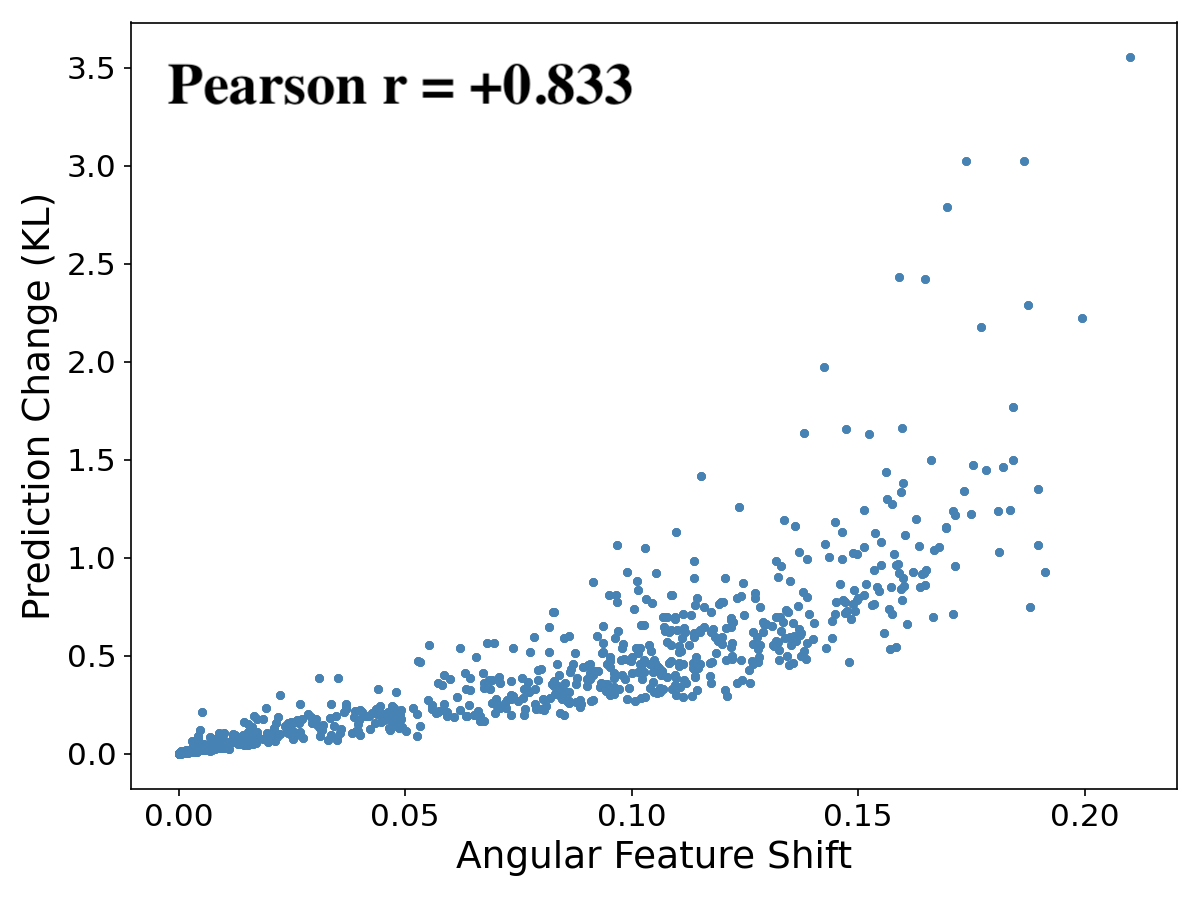}
        \caption{Shift vs Prediction}
    \end{subfigure}
    \caption{Diagnostic analysis of MLMP~\cite{Noori2025TestTimeAO} under CTTA for OVSS with NACLIP~\cite{Hajimiri2024PayAT} ViT-B/32 on PASCAL VOC20-C~\cite{Everingham2014ThePV}. (a) Entropy minimization progressively shrinks the active class vocabulary, measured by counting classes that receive more than 1\% of patch-token argmax predictions. (b) The cosine similarity between adapted image patch features and their source-assigned text prototypes decreases ($r = -0.737$). (c) Similarity to unassigned prototypes increases ($r = +0.854$). (d) Per-sample image feature shift, measured as average cosine distance between adapted and source patch features, correlates with prediction change ($r = +0.833$), indicating that source-relative feature change can identify samples whose representations have changed little. These observations motivate our diversity and anchor consistency components for stability, and our filtering component for adaptation efficiency. Full diagnostic methodology and Cityscapes~\cite{Cordts2016TheCD} diagnostics are in the supplementary material.}
    \label{fig:voc20_diagnostics_pre}
\end{figure*}

Recent VLM TTA methods use weight averaging, prototype shifting, contrastive objectives, and Bayesian updates~\cite{Osowiechi2024WATTWA,Sui2024JustSI,Hakim2024CLIPArTTAO,Lafon2025CLIPTTARC,Zhou2025BayesianTA,Cui2025BayesTTACT,Maharana2024BATCLIPBO}, but they remain largely classification-oriented. Traditional entropy-based adaptation such as TENT, and its continual extension CoTTA, can be instantiated for OVSS by updating normalization parameters over patch-token logits, while MLMP~\cite{Noori2025TestTimeAO} extends episodic TTA to OVSS. What remains missing is a continual OVSS treatment that explicitly preserves vision-language alignment rather than only reducing prediction entropy, which is the gap we study.

To understand why continual adaptation is unstable for OVSS, we diagnose MLMP on PASCAL VOC20-C~\cite{Everingham2014ThePV,hendrycks2019benchmarking} (\cref{fig:voc20_diagnostics_pre}), where VOC20-C applies 15 standard image corruptions~\cite{hendrycks2019benchmarking} at five severity levels to the PASCAL VOC validation set, revealing two failure modes and an adaptation inefficiency. First, entropy minimization shrinks the active class vocabulary, defined as classes receiving more than 1\% of patch-token predictions under the argmax of the patch-level softmax, from near 20 to roughly 1 to 3 by step 500, as large regions collapse to a few dominant concepts. Second, the cosine similarity between adapted image patch features and their source-assigned text prototypes decreases ($r = -0.737$) while similarity to unassigned prototypes increases ($r = +0.854$), eroding vision-language alignment. Third, per-sample image feature shift, measured as the average cosine distance between adapted and source patch features, correlates with prediction change ($r = +0.833$), so many samples undergo minimal prediction change yet still consume backward passes. Since dense predictions depend on patch-to-text comparisons, drift degrades a full segmentation map rather than a single output. The supplementary material confirms these patterns on Cityscapes-C and provides full diagnostic methodology.

These three findings motivate three targeted remedies: a marginal diversity (MDIV) loss to resist class collapse, a cross-modal anchor consistency (CMAC) loss to constrain feature drift, and source-anchored feature salience (SAFS) filtering to skip low-value backward passes and recover part of the source-anchoring cost. Together they form Diversify, Anchor, and Filter (DAF), a simple stabilization framework, with the following contributions:
\begin{itemize}[itemsep=0em,leftmargin=2em]
\item \textbf{CTTA for OVSS:} While MLMP~\cite{Noori2025TestTimeAO} introduced episodic TTA for OVSS, we address the continual setting where model state carries across domains without resetting, preserving vision-language alignment over sequential shifts.
\item \textbf{Diagnostic Analysis:} We characterize why CTTA is unusually unstable for OVSS, revealing class collapse and cross-modal drift as failure modes tied to degradation of vision-language structure, and low-shift redundancy as a motivation for selective updating.
\item \textbf{Stabilization Framework:} We design DAF as a source-anchored regularization around these failure modes, with MDIV and CMAC as complementary core mechanisms and SAFS as a compute-balancing companion to source anchoring, where CMAC is the OVSS-specific component, instantiated on TENT~\cite{Wang2021TentFT} and MLMP~\cite{Noori2025TestTimeAO}.
\item \textbf{Robustness and Evaluation:} We evaluate across five datasets, showing that DAF is robust to aggressive adaptation and learning rate choices across the evaluated continual shifts where entropy minimization collapses, with stable performance under long horizon shifts.
\end{itemize}

\section{Related Work}
\label{sec:relatedwork}

\paragraph{Test-time adaptation.}
Traditional TTA improves robustness through self-supervision~\cite{sun2020test} and entropy minimization, with TENT as a standard baseline~\cite{Wang2021TentFT}, while CTTA extends adaptation to non-stationary streams through teacher averaging~\cite{Wangetal2022cotta}, self-learning~\cite{Rusak2021IfYD}, masked self-supervision~\cite{liu2024continual,doloriel2026family}, prompt memories~\cite{zhang2024dpcore}, subspace constrained online learning~\cite{Duan2025}, and related stabilizers surveyed in~\cite{maharana2026continual}. Sample selection methods such as entropy filtering~\cite{Niu2022EfficientTM,niu2023sar} and object-sensitive selection~\cite{Lee2024EntropyIN} further study which samples to trust. In parallel, VLM TTA has emerged through weight averaging~\cite{Osowiechi2024WATTWA}, prototype shifting~\cite{Sui2024JustSI}, contrastive objectives~\cite{Hakim2024CLIPArTTAO,Lafon2025CLIPTTARC}, Bayesian formulations~\cite{Zhou2025BayesianTA,Cui2025BayesTTACT}, and bimodal alignment~\cite{Maharana2024BATCLIPBO}, but these methods remain largely episodic, single-domain, and classification-oriented.

\paragraph{Segmentation under distribution shift.}
Source-free segmentation methods preserved source knowledge through self-training on unlabeled target images~\cite{Liu2021SourceFreeDA,Ye2021SourceDU}, and subsequent test-time adaptation for segmentation introduced structural priors~\cite{Bateson2022TestTimeAW}, active feedback~\cite{Yuan2023FewCS}, visual prompts~\cite{Chen2023EachTI}, and distribution sensitive parameter adaptation~\cite{Ni2023DistributionAwareCT}. Still, these methods assume fixed label spaces and task-specific components, precluding their use as baselines for open-vocabulary inference where classes are defined by text prompts at test time.

\paragraph{Open-vocabulary dense prediction.}
Open-vocabulary semantic segmentation builds on CLIP~\cite{Radford2021LearningTV}. Early methods turned image-level alignment into dense prediction through mask representations~\cite{Ghiasi2021ScalingOI}, region-text alignment~\cite{Cha2022LearningTG}, and knowledge distillation~\cite{Chen2023ExploringOS,Cai2023MixReorgCM}. Training-free OVSS methods then adapted foundation models for dense inference through correlative self-attention~\cite{Wang2023SCLIPRS}, local patch interactions~\cite{Hajimiri2024PayAT}, and class purification~\cite{Chen2025TrainingFreeCP}. MLMP brought episodic TTA to OVSS~\cite{Noori2025TestTimeAO}, but a continual treatment that preserves vision-language alignment over time remains missing, which DAF addresses with lightweight source-anchored regularization.

\section{Methodology}
\label{sec:methodology}
Our diagnostic analysis in \cref{sec:intro} revealed three failure modes under continual adaptation: patch-level class collapse, vision-language alignment erosion, and redundant gradients from low-shift samples. In OVSS, these are amplified because predictions depend on patch-to-text comparisons across an open label space, where absent classes can absorb probability mass and drift degrades an entire segmentation map rather than a single output. We address them with a marginal diversity (MDIV) loss as an anti-collapse regularizer, a cross-modal anchor consistency (CMAC) loss as the OVSS-specific component, and source-anchored feature salience (SAFS) filtering as a selective update strategy. Each operates without source training data, labels, or architectural changes, though CMAC and SAFS require a frozen copy of the source model for anchored features. We instantiate these on top of TENT~\cite{Wang2021TentFT} and MLMP~\cite{Noori2025TestTimeAO}, yielding DAF-T and DAF-M. We first establish the necessary background and notation, then describe each component.

\paragraph{Preliminaries.} CLIP-based OVSS models~\cite{Hajimiri2024PayAT} consist of an image encoder $g_\theta$ and a text encoder $h_\phi$ that map images and class names into a shared embedding space of dimension $D$. We adopt NACLIP~\cite{Hajimiri2024PayAT} with a ViT-B/32 backbone. Each test image is decomposed into overlapping $224{\times}224$ crops, which we call \emph{samples} and process in a batch of size $Q$. The image encoder outputs $\mathbf{F} \in \mathbb{R}^{(1{+}N) \times D}$, whose patch-token subset $\mathbf{F}_{1:N} \in \mathbb{R}^{N \times D}$ excludes the CLS token and has row $\mathbf{f}_{q,i}$ for patch $i$ of sample $q$, and the text encoder maps $C$ class names, extended with synonyms following standard OVSS practice~\cite{Wang2023SCLIPRS,Noori2025TestTimeAO}, into prototypes $\mathbf{T}^{(r)} \in \mathbb{R}^{C \times D}$ using $R$ prompt templates. Per-patch class probabilities are obtained by softmax over scaled cosine-similarity logits, $p_r(c \mid \mathbf{x}_{q,i}) = \mathrm{softmax}(\tau \cdot \mathrm{cos}(\mathbf{f}_{q,i}, \mathbf{T}^{(r)}_c))_c$. The baseline adaptation signal is template-averaged entropy minimization $\mathcal{L}_{\text{ent}}$ over patch-grid logits. Following TENT~\cite{Wang2021TentFT}, we update only the LayerNorm parameters of $g_\theta$ across a sequence of corruption domains without resetting, making the model susceptible to error accumulation and alignment erosion. A frozen source copy $g_{\theta_s}$ provides anchored features $\mathbf{F}^{\text{src}}$ for CMAC and SAFS. Full details are in the supplementary material.

\subsection{Marginal Diversity Loss}
\label{sec:diversity}

The first failure mode is patch-level class collapse, where the model degenerates from using the full inference-time vocabulary of $C$ candidate text classes to a small subset. We address this with a marginal diversity (MDIV) loss, an anti-collapse regularizer that assumes the source vocabulary remains appropriate for the target domain, an assumption not stressed by the ImageNet-C~\cite{hendrycks2019benchmarking} protocol since corruptions preserve the label space.

The MDIV loss acts on the aggregate class distribution across the batch, unlike the baseline entropy objective which minimizes uncertainty for each patch independently. Given the per-patch softmax probabilities $p_r(c \mid \mathbf{x}_{q,i})$, we compute a marginal class distribution for each prompt template by averaging over all samples in the current adaptation step and over all patch tokens within each sample:
\begin{equation}
\label{eq:marginal}
\bar{p}_r(c) = \frac{1}{Q \cdot N} \sum_{q=1}^{Q} \sum_{i=1}^{N} p_r(c \mid \mathbf{x}_{q,i}).
\end{equation}
The marginal diversity loss maximizes the entropy of these marginal distributions and then averages over templates:
\begin{equation}
\label{eq:div_loss}
\begin{split}
\mathcal{L}_{\text{div}} &= -\frac{1}{R} \sum_{r=1}^{R} H(\bar{p}_r) \\
&= \frac{1}{R} \sum_{r=1}^{R} \sum_{c=1}^{C} \bar{p}_r(c) \log \bar{p}_r(c).
\end{split}
\end{equation}
Minimizing $\mathcal{L}_{\text{div}}$ maximizes the batch-level marginal class entropy for each template. $\mathcal{L}_{\text{div}}$ discourages degenerate concentration of the aggregate prediction distribution rather than enforcing uniform per-image class occurrence, acting as a soft anti-collapse regularizer whose strength is controlled by $\lambda_{\text{div}}$. When only one prompt template is used, $R = 1$ and the expressions reduce to the simpler single-template form. Although $\mathcal{L}_{\text{div}}$ involves entropy, it is not redundant with $\mathcal{L}_{\text{ent}}$: $\mathcal{L}_{\text{ent}}$ applies the logarithm inside the patch-level summation (average per-patch entropy), whereas $\mathcal{L}_{\text{div}}$ applies it outside (entropy of the average prediction).

\subsection{Cross-Modal Anchor Consistency}
\label{sec:cmac}

While entropy minimization and marginal diversity (MDIV) loss operate at the prediction level, they do not explicitly constrain the feature space. As the OVSS-specific component, our cross-modal anchor consistency (CMAC) loss directly regularizes the vision-language alignment that underpins open-vocabulary prediction. Our diagnostics revealed two drift behaviors harmful to segmentation. \textbf{Drift-Away} occurs when the adapted image feature moves away from its assigned text prototype, weakening the correct class association. \textbf{Drift-Toward} occurs when the adapted image feature moves closer to an unassigned text prototype, increasing the risk of misclassification. CMAC addresses both behaviors by regularizing feature-level drift using the frozen source model as an anchor. By design, CMAC prioritizes preservation of source-relative geometry over unconstrained correction, which may become conservative when source prototype assignments are unreliable under large domain gaps.

\paragraph{Source-assigned prototypes.}
For each patch token $\mathbf{f}^{\text{src}}_{q,i}$ from sample $q$ in the frozen source model, we compute its similarity to all prompt-integrated text prototypes and assign it to the nearest one:
\begin{equation}
\label{eq:assignment}
a_{q,i} = \arg\max_{c} \; \tau \cdot \mathrm{cos}(\mathbf{f}^{\text{src}}_{q,i}, \mathbf{T}_c),
\end{equation}
where $\tau$ is the CLIP logit scale. When multiple prompt templates are available, we average their text embeddings before computing these assignments, so $\mathbf{T}_c$ denotes the prompt-integrated prototype for class $c$.

\paragraph{Anchor consistency.}
We define $s^{\text{adapt}}_{q,i,c} = \tau \cdot \mathrm{cos}(\mathbf{f}^{\text{adapt}}_{q,i}, \mathbf{T}_c)$ and $s^{\text{src}}_{q,i,c}$ analogously. The CMAC loss combines two one-sided hinge terms into a single objective. The first penalizes drift away from the assigned prototype, active only when adapted similarity drops below source similarity, $\mathrm{ReLU}(s^{\text{src}}_{q,i,a_{q,i}} - s^{\text{adapt}}_{q,i,a_{q,i}})$. The second penalizes drift toward unassigned prototypes, averaged over all $C{-}1$ wrong classes and active only when adapted similarity exceeds source similarity, $\mathrm{ReLU}(s^{\text{adapt}}_{q,i,c} - s^{\text{src}}_{q,i,c})$ for $c \neq a_{q,i}$. Introducing the Kronecker delta $\delta_{c,a_{q,i}}$ (equal to 1 if $c = a_{q,i}$ and 0 otherwise), the sign function $\xi_{q,i,c} = 2\,\delta_{c,a_{q,i}} - 1$, and per-class weight $w_{q,i,c} = \frac{1}{1 + (C{-}2)(1 - \delta_{c,a_{q,i}})}$, which equals $1$ for the assigned prototype and $\frac{1}{C{-}1}$ for each unassigned one, both terms unify into:
\begin{equation}
\label{eq:cmac_loss}
\begin{split}
\mathcal{L}_{\text{cmac}} &= \frac{1}{Q \cdot N} \sum_{q=1}^{Q} \sum_{i=1}^{N} \sum_{c=1}^{C} w_{q,i,c} \\
&\phantom{=} \cdot \mathrm{ReLU}\!\left(\xi_{q,i,c}\,(s^{\text{src}}_{q,i,c} - s^{\text{adapt}}_{q,i,c})\right),
\end{split}
\end{equation}
where $w_{q,i,c}$ matches the $\frac{1}{C{-}1}$ averaging in the drift-toward term.

\subsection{Source-Anchored Feature Salience Filtering}
\label{sec:safs}

Continual adaptation need not backpropagate through every incoming sample equally. Because DAF already computes source and adapted features for CMAC, their discrepancy provides a cheap salience signal for identifying samples whose representations have changed little and whose backward computation can be skipped. Source-anchored feature salience (SAFS) filtering uses this signal to recover part of the adaptation time introduced by CMAC's extra source-model forward pass. While MDIV and CMAC are responsible for stability and accuracy, SAFS mainly trades a small amount of peak accuracy for fewer backward passes and lower adaptation time. SAFS is optional and does not alter the stabilization mechanism.

For each sample $q$, we measure the angular feature shift between the adapted and source models by averaging cosine distance over all ViT patch tokens:
\begin{equation}
\label{eq:shift}
\Delta_q = 1 - \frac{1}{N} \sum_{i=1}^{N} \mathrm{cos}\!\left(\mathbf{f}^{\text{adapt}}_{q,i},\; \mathbf{f}^{\text{src}}_{q,i}\right).
\end{equation}
Rather than using a fixed threshold, we compute an adaptive threshold $\rho_{\text{safs}} = \mu(\boldsymbol{\Delta}) - \alpha_{\text{safs}} \cdot \sigma(\boldsymbol{\Delta})$, where $\mu(\boldsymbol{\Delta})$ and $\sigma(\boldsymbol{\Delta})$ are the mean and standard deviation of per-sample shifts and $\alpha_{\text{safs}}$ is a margin hyperparameter. Samples with $\Delta_q > \rho_{\text{safs}}$ are retained for the loss computation and gradient update, while the remaining samples are filtered out. If all samples fall below the threshold, no filtering is applied to avoid an empty batch.

\paragraph{Integration.}
SAFS filtering is applied before computing any loss, restricting the summation over $q$ in the entropy, MDIV, and CMAC losses to SAFS-retained samples. Because DAF augments rather than replaces the baseline objective, the complete adaptation loss combines the chosen baseline loss $\mathcal{L}_{\text{base}}$ with our two regularization terms: $\mathcal{L} = \mathcal{L}_{\text{base}} + \lambda_{\text{div}} \, \mathcal{L}_{\text{div}} + \lambda_{\text{cmac}} \, \mathcal{L}_{\text{cmac}}$, where $\mathcal{L}_{\text{base}}$ denotes the native objective of the chosen baseline, such as TENT's template-averaged prediction entropy loss or MLMP's multi-layer, multi-prompt entropy objective with its additional classification term.

\section{Experiments}
\label{sec:experiments}

\subsection{Dataset}

We evaluate on five semantic segmentation datasets spanning diverse visual domains: natural scenes (PASCAL VOC20/VOC21~\cite{Everingham2014ThePV}), autonomous driving (Foggy Cityscapes\cite{Sakaridis2017SemanticFS}, Cityscapes~\cite{Cordts2016TheCD}), underwater imagery (SUIM~\cite{islam2020semantic}), and remote sensing (LoveDA~\cite{Wang2021LoveDAAR}). For each dataset, we generate corrupted variants following the ImageNet-C protocol~\cite{hendrycks2019benchmarking} at the highest severity (level~5), covering 15 corruption types across noise, blur, weather, and digital categories. The resulting corrupted variants are denoted with a ``\texttt{-C}'' suffix (\eg, VOC20-C, LoveDA-C). To leverage the open-vocabulary capability of CLIP, we use class extensions for VOC20 and VOC21 whenever available, augmenting the base vocabulary with synonyms and subcategory names. VOC20 extends from 20 to 30 classes and VOC21 from 21 to 56, following the standard protocol of SCLIP~\cite{Wang2023SCLIPRS}, NACLIP~\cite{Hajimiri2024PayAT}, and MLMP~\cite{Noori2025TestTimeAO}. Further dataset details are in the supplementary.

\subsection{Implementation Details}

We compare against baseline methods for CTTA, grouping non-VLM (Vision-Language Model) adaptations (TENT~\cite{Wang2021TentFT}, COTTA\cite{Wangetal2022cotta}, SAR\cite{niu2023sar}) with VLM adaptations (WATT~\cite{Osowiechi2024WATTWA}, CLIPArTT~\cite{Hakim2024CLIPArTTAO}, SegTTO\cite{Silva2025TestTimeOF}, MLMP~\cite{Noori2025TestTimeAO}). Segmentation-specific CTTA methods~\cite{Ni2023DistributionAwareCT,Chen2023EachTI} assume fixed label spaces and task-specific components, precluding open-vocabulary use. RPL\cite{Rusak2021IfYD}, DEYO\cite{Lee2024EntropyIN}, M2A\cite{doloriel2026family} are in the supplementary. WATT~\cite{Osowiechi2024WATTWA}, CLIPArTT~\cite{Hakim2024CLIPArTTAO}, and MLMP~\cite{Noori2025TestTimeAO} were designed for episodic TTA with 10 optimization steps per batch, but we evaluate them with 1 step per batch for CTTA consistency. We restrict adaptation to LayerNorm parameters of the CLIP image encoder, keeping the text encoder fixed. Our method is applied on top of TENT (denoted DAF-T) and MLMP (denoted DAF-M). We build upon the public MLMP codebase. 

For all OVSS baselines, we use NACLIP~\cite{Hajimiri2024PayAT} with a ViT-B/32 backbone as the open-vocabulary semantic segmentation (OVSS) model. Following MLMP~\cite{Noori2025TestTimeAO}, each input image is divided into overlapping $224{\times}224$ crops, which are independently passed through the CLIP image encoder. Unless otherwise stated, we use a dataloader batch size of~8 input images before crop extraction, SGD optimizer with a learning rate of~$1{\times}10^{-3}$, perform a single optimization step per batch, and set DAF hyperparameters to $\lambda_{\text{div}} = 2.0$, $\lambda_{\text{cmac}} = 0.5$, and $\alpha_{\text{safs}} = 0.5$. TENT and DAF-T use a single prompt template, while MLMP and DAF-M use 7 prompt templates following the original MLMP setup. All experiments use seed~0. However, the main benchmark results in \cref{tab:ctta_benchmark_fine_vitb32} are averaged over three seeds (0, 1, 2). Additional results including backbone generalization (ViT-B/16, ViT-L/14), forward transfer, OVSS variant analysis, and cross-dataset ablations are provided in the supplementary material.

\subsection{Results}

\begin{table*}[tb]
    \centering
    \begin{small}
    \caption{Benchmark results on segmentation datasets under CTTA using NACLIP\cite{Hajimiri2024PayAT} ViT-B/32. Adaptation is continuous between domains, without resetting the model. Entries report mIoU (\%), averaged over seeds 0/1/2. "--" indicates OOM (Out-Of-Memory) error. Results on NACLIP\cite{Hajimiri2024PayAT} ViT-B/16 and ViT-L/14 are in the supplementary material.}
    \label{tab:ctta_benchmark_fine_vitb32}
    \resizebox{0.95\textwidth}{!}{
    \begin{tabular}{ll|c|ccc|cccccc}
    \toprule
    \multicolumn{2}{c|}{\multirow{3}{*}{Dataset}} & \multicolumn{10}{c}{Adaptation Method} \\
    \cmidrule(lr){3-12}
    \multicolumn{2}{c|}{} & \multirow{2}{*}{SOURCE} & \multicolumn{3}{c|}{Non-VLM} & \multicolumn{6}{c}{VLM} \\
    \cmidrule(lr){4-6} \cmidrule(lr){7-12}
    \multicolumn{2}{c|}{} &  & TENT\cite{Wang2021TentFT} & CoTTA\cite{Wangetal2022cotta} & SAR\cite{niu2023sar} & WATT\cite{Osowiechi2024WATTWA} & CLIPArTT\cite{Hakim2024CLIPArTTAO} & SegTTO\cite{Silva2025TestTimeOF} & MLMP\cite{Noori2025TestTimeAO} & DAF-T & DAF-M \\
    \midrule

    \multicolumn{2}{l|}{VOC20\cite{Everingham2014ThePV} (Original)} & 72.4 & 74.2\ppm0.06 & 72.3\ppm0.02 & 73.6\ppm0.01 & 66.6\ppm0.11 & 19.1\ppm0.51 & 39.1\ppm0.63 & 76.6\ppm0.01 & 73.1\ppm0.07 & \textbf{78.2\ppm0.02} \\
    \rowcolor{gray!15}\multicolumn{2}{l|}{VOC20-C\cite{Everingham2014ThePV}} & 54.4 & 3.3\ppm0.03 & 44.2\ppm1.58 & 47.9\ppm0.14 & 45.5\ppm0.12 & 2.0\ppm0.06 & 2.4\ppm0.12 & 9.7\ppm3.85 & 57.3\ppm0.06 & \textbf{62.7\ppm0.03} \\
    \midrule
    
    \multicolumn{2}{l|}{VOC21\cite{Everingham2014ThePV} (Original)} & 41.3 & 42.0\ppm0.02 & 41.3\ppm0.00 & 41.9\ppm0.05 & 41.0\ppm0.03 & 17.5\ppm1.07 & 41.2\ppm0.38 & 42.9\ppm0.01 & 41.6\ppm0.02 & \textbf{43.5\ppm0.02} \\
    \rowcolor{gray!15}\multicolumn{2}{l|}{VOC21-C\cite{Everingham2014ThePV}} & 33.9 & 5.0\ppm0.76 & 29.4\ppm0.79 & 33.7\ppm0.03 & 28.7\ppm0.01 & 3.5\ppm0.25 & 1.0\ppm0.05 & 9.4\ppm0.96 & 35.1\ppm0.02 & \textbf{36.1\ppm0.01} \\
    \midrule
    
    \parbox[t]{8mm}{\multirow{3}{*}{\rotatebox[origin=c]{0}{\parbox{8mm}{Foggy\newline Cityscapes\cite{Sakaridis2017SemanticFS}}}}}
    & Light Fog     & 27.5 & 27.8\ppm0.01 & 27.5\ppm0.00 & 27.9\ppm0.01 & 24.6\ppm0.03 & -- & \textbf{31.4\ppm0.14} & 30.7\ppm0.00 & 27.8\ppm0.01 & 30.6\ppm0.02 \\
    & Medium Fog    & 27.0 & 27.6\ppm0.01 & 26.9\ppm0.00 & 27.9\ppm0.01 & 24.5\ppm0.19 & -- & 26.5\ppm0.84 & 28.9\ppm0.06 & 27.5\ppm0.00 & \textbf{30.0\ppm0.02} \\
    & Dense Fog     & 25.7 & 26.2\ppm0.02 & 25.4\ppm0.03 & 26.7\ppm0.01 & 19.2\ppm0.13 & -- & 21.9\ppm1.19 & 27.4\ppm0.04 & 26.3\ppm0.01 & \textbf{28.7\ppm0.01} \\
    \cmidrule{2-12}
    \rowcolor{gray!15} & Mean   & 26.7 & 27.2\ppm0.01 & 26.6\ppm0.01 & 27.5\ppm0.01 & 22.8\ppm0.12 & -- & 26.6\ppm0.63 & 29.0\ppm0.03 & 27.2\ppm0.01 & \textbf{29.8\ppm0.00} \\
    \midrule
    \rowcolor{gray!15}\multicolumn{2}{l|}{Cityscapes-C\cite{Cordts2016TheCD}} & 17.8 & 4.5\ppm0.01 & 16.7\ppm0.52 & 17.1\ppm0.02 & 7.7\ppm0.00 & -- & 2.9\ppm0.01 & 1.6\ppm0.01 & \textbf{19.0\ppm0.01} & 18.5\ppm0.01 \\
    \midrule
    
    \multicolumn{2}{l|}{SUIM\cite{islam2020semantic} (Original)} & 22.4 & 22.5\ppm0.02 & 22.4\ppm0.00 & 22.5\ppm0.01 & 21.9\ppm0.36 & 22.5\ppm0.01 & \textbf{23.9\ppm0.20} & 12.9\ppm0.02 & 22.5\ppm0.02 & 13.0\ppm0.04 \\
    \rowcolor{gray!15}\multicolumn{2}{l|}{SUIM-C\cite{islam2020semantic}} & 19.6 & 19.5\ppm0.08 & 19.4\ppm0.01 & 19.1\ppm0.07 & 18.5\ppm0.07 & 12.7\ppm0.15 & 12.0\ppm0.02 & 10.6\ppm0.02 & \textbf{20.3\ppm0.00} & 11.9\ppm0.02 \\
    \midrule

    \parbox[t]{15mm}{\multirow{2}{*}{\rotatebox[origin=c]{0}{LoveDA\cite{Wang2021LoveDAAR}}}}
    & Rural                & 16.3 & 16.8\ppm0.12 & 16.1\ppm0.00 & 17.2\ppm0.12 & 19.2\ppm0.07 & -- & 11.8\ppm0.03 & 22.0\ppm0.30 & 17.2\ppm0.02 & \textbf{28.1\ppm0.11} \\
    & Urban                & 29.2 & 23.8\ppm0.27 & 28.1\ppm0.04 & 23.4\ppm0.25 & 25.6\ppm0.06 & -- & 3.8\ppm0.02 & 11.4\ppm2.69 & 31.0\ppm0.03 & \textbf{35.5\ppm0.08} \\
    \cmidrule{2-12}
    \rowcolor{gray!15} & Mean    & 22.7 & 20.3\ppm0.19 & 22.1\ppm0.02 & 20.3\ppm0.18 & 22.4\ppm0.06 & -- & 7.8\ppm0.02 & 16.7\ppm1.50 & 24.1\ppm0.00 & \textbf{31.8\ppm0.02} \\
    \midrule
    \rowcolor{gray!15}\multicolumn{2}{l|}{LoveDA-C\cite{Wang2021LoveDAAR}} & 9.2 & 1.2\ppm0.00 & -- & 7.6\ppm0.15 & -- & -- & 5.2\ppm0.00 & 5.2\ppm0.00 & 12.3\ppm0.01 & \textbf{14.9\ppm0.01} \\

    \bottomrule    
    \end{tabular}
    }
    \end{small}
\end{table*}

\paragraph{Benchmark results.}
\cref{tab:ctta_benchmark_fine_vitb32} reports mIoU under continual adaptation across five datasets and their corrupted variants. On clean VOC20 and VOC21, entropy-based methods already improve over source, so DAF yields smaller but consistent gains, showing that stabilization need not sacrifice peak accuracy when the domain is benign. The benefit is larger under corrupted or larger-gap settings. On VOC20-C and VOC21-C, TENT and MLMP collapse below source, consistent with the class collapse and alignment erosion diagnosed in \cref{sec:intro}, whereas DAF-T and DAF-M remain stable and exceed source. The same pattern appears as the shift strengthens on Foggy Cityscapes, where DAF-M achieves the best mean and gains about 3 points over source, and on LoveDA and LoveDA-C, where DAF-M achieves large gains over both source and MLMP. On Cityscapes-C, entropy-based methods also collapse while DAF stays above source. On SUIM-C, DAF-T improves over source and TENT, while DAF-M improves over MLMP but remains below source. Given SUIM's 110 validation images, we interpret this setting cautiously, though it still reflects MLMP degradation and CMAC conservatism under large domain gaps (\cref{sec:cmac}). ImageNet-C corruptions preserve the label space by design, so these results do not stress the diversity loss assumption that the source vocabulary remains appropriate under shifts, and evaluating under vocabulary mismatch remains future work. Among VLM-specific baselines, CLIPArTT and SegTTO struggle under continual adaptation, indicating that these tested VLM-specific objectives do not inherently prevent alignment erosion over sequential domains. The stability of DAF-T and DAF-M across these settings aligns with the complementary roles of the MDIV loss and the CMAC loss described in \cref{sec:methodology}, and the component ablation in \cref{tab:voc20_efficiency} supports their joint contribution to stability.

\begin{table}[htbp]
    \caption{Domain-shuffled multi-pass CTTA on Cityscapes-C\cite{Cordts2016TheCD} and LoveDA \cite{Wang2021LoveDAAR} using NACLIP\cite{Hajimiri2024PayAT} ViT-B/32. In each pass, we shuffled the domain order without resetting the model. Entries report mIoU (\%). "-" indicates OOM. Results on NACLIP\cite{Hajimiri2024PayAT} ViT-B/16 are in the supplementary material.}
    \label{tab:ctta_shuffle_pround_vitb32}
    \begin{center}
    \footnotesize
    \setlength\tabcolsep{4pt}
    \begin{tabular}{l|ccc|ccc}
        \toprule
        \multirow{2}{*}{Variant} & \multicolumn{3}{c|}{Cityscapes-C} & \multicolumn{3}{c}{LoveDA} \\
        \cline{2-7}
         & Pass 1 & Pass 2 & Pass 3 & Pass 1 & Pass 2 & Pass 3 \\
        \midrule
        SOURCE      & 17.8 & 17.8 & 17.8    & 22.7 & 22.7 & 22.7  \\
        TENT        & 5.3 & 2.0 & 2.0       & 20.1 & 13.7 & 5.5  \\
        SAR         & 16.2 & 15.9 & 16.6    & 20.2 & 20.6 & 21.4  \\
        SegTTO      & 2.6 & 2.0 & 2.0       & 3.4 & 2.4 & 1.9  \\
        MLMP        & 1.2 & 0.2 & 0.2       & 15.2 & 4.9 & 4.9  \\
        \midrule
        \rowcolor{gray!15}DAF-T & \textbf{19.0} & 19.0 & \textbf{18.8}        & 24.1 & 24.8 & 25.1  \\
        \rowcolor{gray!15}DAF-M & 18.2 & \textbf{19.1} & 18.3        & \textbf{31.8} & \textbf{33.5} & \textbf{33.0}  \\
        \bottomrule
    \end{tabular}
    \end{center}
\end{table}

\paragraph{Domain-shuffled CTTA.}
\cref{tab:ctta_shuffle_pround_vitb32} evaluates long-horizon stability under domain-shuffled multi-pass CTTA on Cityscapes-C and LoveDA, where the domain order is reshuffled each pass without resetting the model. This setting stresses error accumulation: repeated exposure to shifting domains should reveal whether adaptation compounds or stabilizes. On both datasets, TENT and MLMP degrade progressively across passes, falling far below source. SAR holds near source but never exceeds it. DAF-T and DAF-M stay above source on Cityscapes-C across all three passes, and DAF-M improves on LoveDA across passes, gaining over 10 points above source by the final pass. The progressive degradation of entropy-based methods reflects the error accumulation and alignment erosion diagnosed in \cref{sec:intro}. That DAF-T and DAF-M avoid this fate, with DAF-M improving on LoveDA across passes, points to the joint stabilization of MDIV and the source-anchored CMAC loss (\cref{sec:cmac}) in limiting drift accumulation.

\begin{table}[t!]
    \caption{Learning rate sweep on PASCAL VOC20-C \cite{Everingham2014ThePV} with NACLIP\cite{Hajimiri2024PayAT} ViT-B/32. Entries report mIoU (\%). Results on LoveDA\cite{Wang2021LoveDAAR} are in the supplementary material.}
    \label{tab:voc20_lr_sweep}
    \begin{center}
    \footnotesize
    \setlength\tabcolsep{4pt}
    \begin{tabular}{l|cccc|c}
        \toprule
        \multirow{2}{*}{METHOD} & \multicolumn{4}{c|}{Learning Rate} & \multirow{2}{*}{Mean$\uparrow$} \\
        \cline{2-5}
         & 1e-2 & 1e-3 & 1e-4 & 1e-5 & \\
        \midrule
        SOURCE      & 54.4 & 54.4 & 54.4 & 54.4 & 54.4 \\
        TENT        & 0.7 & 3.3 & 55.1 & 55.0 & 28.5 \\
        CoTTA       & 45.9 & 42.6 & 53.7 & 54.4 & 49.2 \\
        SAR         & 33.9 & 48.1 & 55.0 & 54.7 & 47.9 \\
        WATT        & 3.5 & 45.7 & 56.8 & 57.9 & 41.0 \\
        CLIPArTT    & 1.4 & 2.1 & 10.2 & 52.0 & 16.4 \\
        SegTTO      & 0.6 & 2.3 & 8.4 & 32.4 & 10.9 \\
        MLMP        & 1.7 & 13.5 & 57.0 & \textbf{58.6} & 32.7 \\
        \midrule
        \rowcolor{gray!15}DAF-T & \textbf{53.1} & 57.4 & 55.5 & 54.6 & 55.2 \\
        \rowcolor{gray!15}DAF-M & 41.0 & \textbf{62.7} & \textbf{61.5} & \textbf{58.6} & \textbf{56.0} \\
        \bottomrule
    \end{tabular}
    \end{center}
\end{table}

\paragraph{Learning rate robustness.}
\cref{tab:voc20_lr_sweep} sweeps the learning rate from $1{\times}10^{-2}$ to $1{\times}10^{-5}$ on VOC20-C to test how each method handles aggressive adaptation. Entropy-based methods are highly sensitive: TENT and MLMP collapse at higher learning rates and only remain stable at $1{\times}10^{-4}$ and below. CoTTA and SAR are more resilient but remain below the source mIoU of 54.4. VLM-specific baselines struggle across all learning rates. DAF-T and DAF-M exceed the source and maintain stability even at $1{\times}10^{-2}$ where TENT and MLMP collapse to near-zero. This robustness allows DAF to operate at learning rates that destabilize entropy-based methods, a practical advantage that stems from the joint stabilization of the MDIV loss and the CMAC loss.

\begin{table*}[htbp]
    \caption{Longterm recurring CTTA on Cityscapes-C\cite{Cordts2016TheCD} and LoveDA\cite{Wang2021LoveDAAR} with NACLIP\cite{Hajimiri2024PayAT} ViT-B/32. In each pass, we follow the same domain order without resetting the model. Entries report mIoU (\%).}
    \label{tab:voc_recur_pround}
    \begin{center}
    \footnotesize
    \setlength\tabcolsep{3pt}
    \begin{tabular}{l|ccccc|c|ccccc|c}
        \toprule
        \multirow{2}{*}{METHOD} & \multicolumn{6}{c|}{Cityscapes-C} & \multicolumn{6}{c}{LoveDA} \\
        \cline{2-13}
         & Pass 1 & Pass 2 & Pass 3 & Pass 4 & Pass 5 & Mean$\uparrow$ & Pass 1 & Pass 2 & Pass 3 & Pass 4 & Pass 5 & Mean$\uparrow$ \\
        \midrule
        SOURCE      & 17.8 & 17.8 & 17.8 & 17.8 & 17.8 & 17.8       & 22.7 & 22.7 & 22.7 & 22.7 & 22.7 & 22.7 \\
        TENT        & 4.5 & 2.0 & 2.0 & 2.0 & 2.0 & 2.5             & 20.1 & 13.7 & 5.5 & 4.5 & 4.4 & 9.6 \\
        SAR         & 17.1 & 16.8 & 16.5 & 16.6 & 16.6 & 16.7       & 20.2 & 20.6 & 21.4 & 22.5 & 23.0 & 21.5 \\
        SegTTO      & 2.9 & 2.1 & 2.1 & 2.1 & 2.1 & 2.2             & 3.4 & 2.4 & 1.9 & 1.6 & 1.5 & 2.2 \\
        MLMP        & 1.6 & 0.2 & 0.2 & 0.2 & 0.2 & 0.5             & 15.2 & 4.9 & 4.9 & 4.9 & 4.9 & 7.0 \\
        \midrule
        \rowcolor{gray!15}DAF-T & \textbf{19.0} & 19.0 & \textbf{18.8} & \textbf{18.6} & \textbf{18.5} & \textbf{18.8}       & 24.1 & 24.8 & 25.1 & 25.2 & 25.3 & 24.9 \\
        \rowcolor{gray!15}DAF-M & 18.5 & \textbf{19.2} & 18.6 & 18.1 & 17.6 & 18.4       & \textbf{31.8} & \textbf{33.5} & \textbf{33.0} & \textbf{32.1} & \textbf{31.3} & \textbf{32.3} \\
        \bottomrule
    \end{tabular}
    \end{center}
\end{table*}

\paragraph{Recurring adaptation.}
\cref{tab:voc_recur_pround} extends the multi-pass evaluation to five passes with a fixed domain order. As in the domain-shuffled setting, TENT and MLMP collapse by the second or third pass and never recover, while DAF-T and DAF-M remain stable across all five passes, staying above source on Cityscapes-C. On LoveDA, DAF-M achieves a mean of 32.3, peaking mid-sequence before settling without catastrophic degradation. These results confirm that the stability observed under domain-shuffled CTTA persists over longer horizons with a fixed order, further supporting the joint role of MDIV and CMAC in limiting drift accumulation.

\begin{table}[htbp]
    \caption{Ablation study on DAF components and adaptation efficiency on PASCAL VOC20-C \cite{Everingham2014ThePV} using NACLIP\cite{Hajimiri2024PayAT} ViT-B/32. Each image is divided into $224{\times}224$ crops as input to the CLIP model. Each domain has $S{=}1456$ samples with batch size of 8, evaluated on an AMD Instinct MI200 GPU. \#Fward and \#Bward denote the total number of forward and backward passes per domain, expressed in terms of $S$. Time (s) is the per-domain adaptation time in seconds. Results on LoveDA \cite{Wang2021LoveDAAR} dataset are in the supplementary material.}
    \label{tab:voc20_efficiency}
    \begin{center}
    \footnotesize
    \setlength\tabcolsep{2pt}
    \begin{tabular}{l|ccc|ccc|c}
        \toprule
         Method & $\mathcal{L}_{\text{div}}$ & $\mathcal{L}_{\text{cmac}}$ & $\rho_{\text{safs}}$ & \#Fward$\downarrow$ & \#Bward$\downarrow$ & Time(s)$\downarrow$ & mIoU$\uparrow$ \\
        \midrule
        SOURCE      & $\times$ & $\times$ & $\times$ & $1.0S$ & $0$       & 48.0  & 54.4 \\
        TENT        & $\times$ & $\times$ & $\times$ & $1.0S$ & $1.0S$    & 48.4  & 3.3 \\
        MLMP        & $\times$ & $\times$ & $\times$ & $1.0S$ & $1.0S$    & 67.4  & 9.7 \\
        \midrule
        \rowcolor{gray!15}DAF-T      & $\checkmark$ & $\checkmark$ & $\checkmark$ & $2.0S$ & $0.7S$     & 69.4  & 57.3 \\
        \rowcolor{gray!15}DAF-M      & $\checkmark$ & $\times$ & $\times$ & $2.0S$ & $1.0S$             & 68.0  & 44.4 \\
        \rowcolor{gray!15}DAF-M      & $\times$ & $\checkmark$ & $\times$ & $2.0S$ & $1.0S$             & 50.6  & 39.1 \\
        \rowcolor{gray!15}DAF-M      & $\checkmark$ & $\checkmark$ & $\times$ & $2.0S$ & $1.0S$         & 68.4  & 63.4 \\
        \rowcolor{gray!15}DAF-M      & $\checkmark$ & $\checkmark$ & $\checkmark$ & $2.0S$ & $0.7S$     & 50.8  & 62.7 \\
        \bottomrule
    \end{tabular}
    \end{center}
\end{table}

\paragraph{Component analysis.}
\cref{tab:voc20_efficiency} ablates the three DAF components on VOC20-C. The MDIV and CMAC are the core stabilization mechanisms: MDIV alone reaches 44.4 and CMAC alone 39.1, both improving substantially over MLMP's 9.7 but neither matching the combined 63.4 without SAFS. This confirms they address complementary failure modes, with MDIV resisting class collapse and CMAC constraining feature drift, as designed in \cref{sec:methodology}. Comparing MDIV+CMAC with and without SAFS shows that filtering reduces backward passes by approximately 30\% (from $1.0S$ to $0.7S$ per domain) and lowers DAF-M adaptation time from 68.4 s to 50.8 s, at the cost of less than one mIoU point (62.7 vs 63.4), showing that SAFS partly offsets the added adaptation time introduced by source anchoring rather than contributing directly to stability. Notably, DAF-T increases time over TENT (69.4 s vs 48.4 s) because TENT's forward and backward costs are roughly equal (we doubled the forward pass in DAF), so the additional source forward pass is not offset by SAFS backward savings, whereas MLMP's more expensive multi-layer backward makes SAFS savings outweigh the source forward, yielding DAF-M a net decrease from 67.4 s to 50.8 s. SAFS does not reduce the additional source forward pass required by CMAC, so its net benefit depends on how much backward cost can be saved.

\begin{figure*}[htbp]
    \centering
    \begin{subfigure}{0.24\textwidth}
        \centering
        \includegraphics[width=\textwidth]{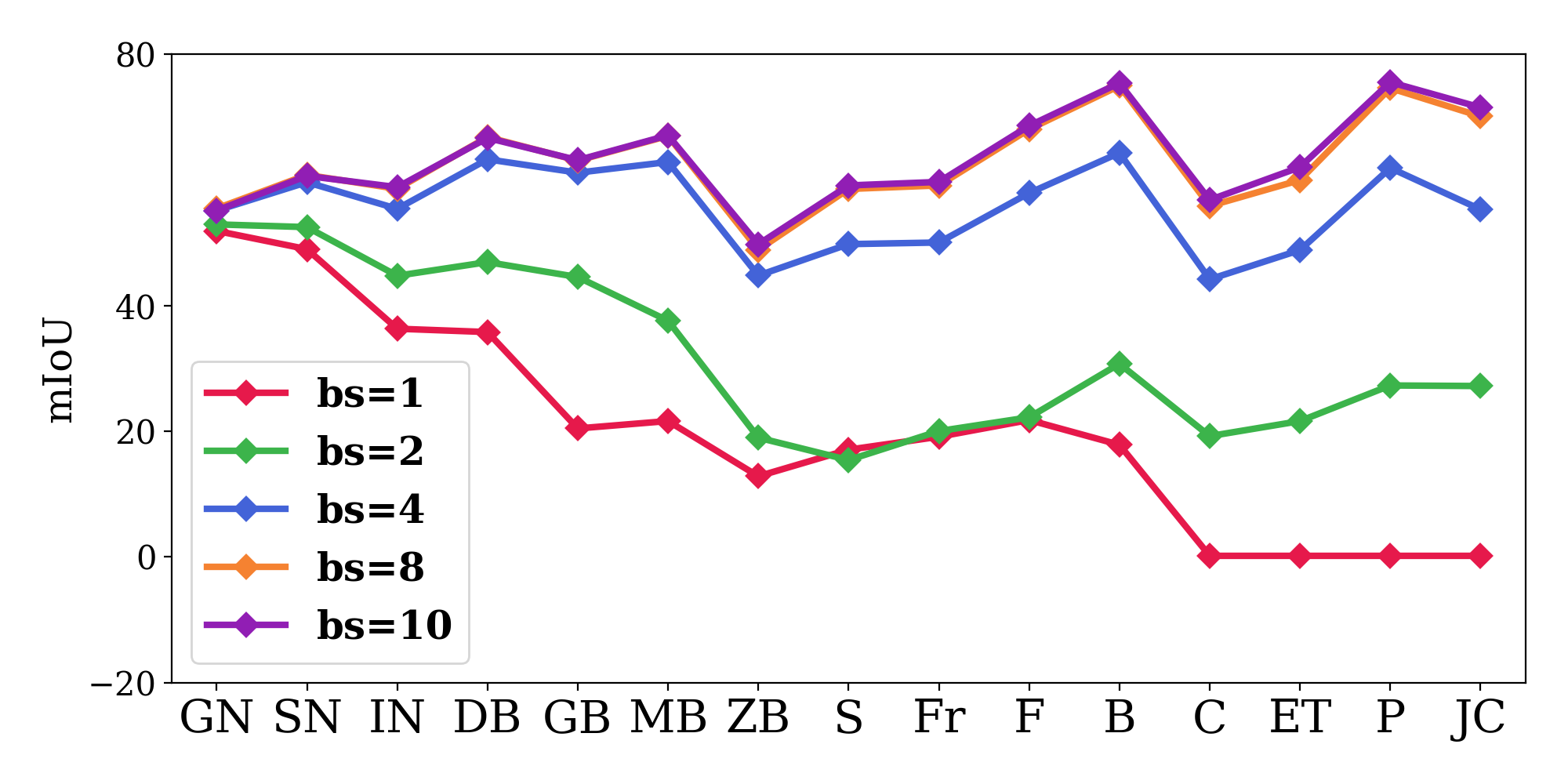}
        \caption{Batch Size}
    \end{subfigure}
    \hfill
    \begin{subfigure}{0.24\textwidth}
        \centering
        \includegraphics[width=\textwidth]{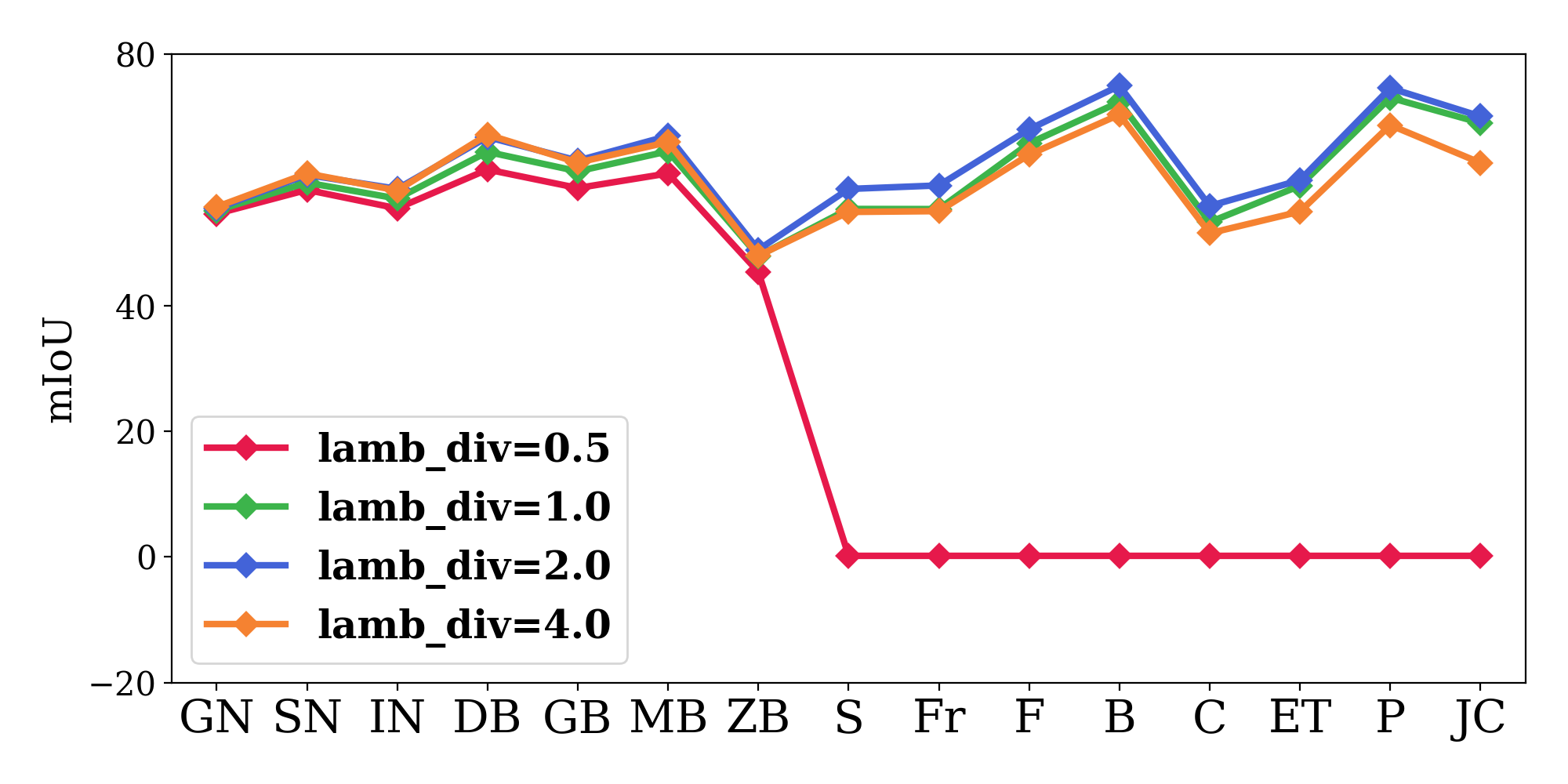}
            \caption{$\lambda_{\text{div}}$}
    \end{subfigure}
    \hfill
    \begin{subfigure}{0.24\textwidth}
        \centering
        \includegraphics[width=\textwidth]{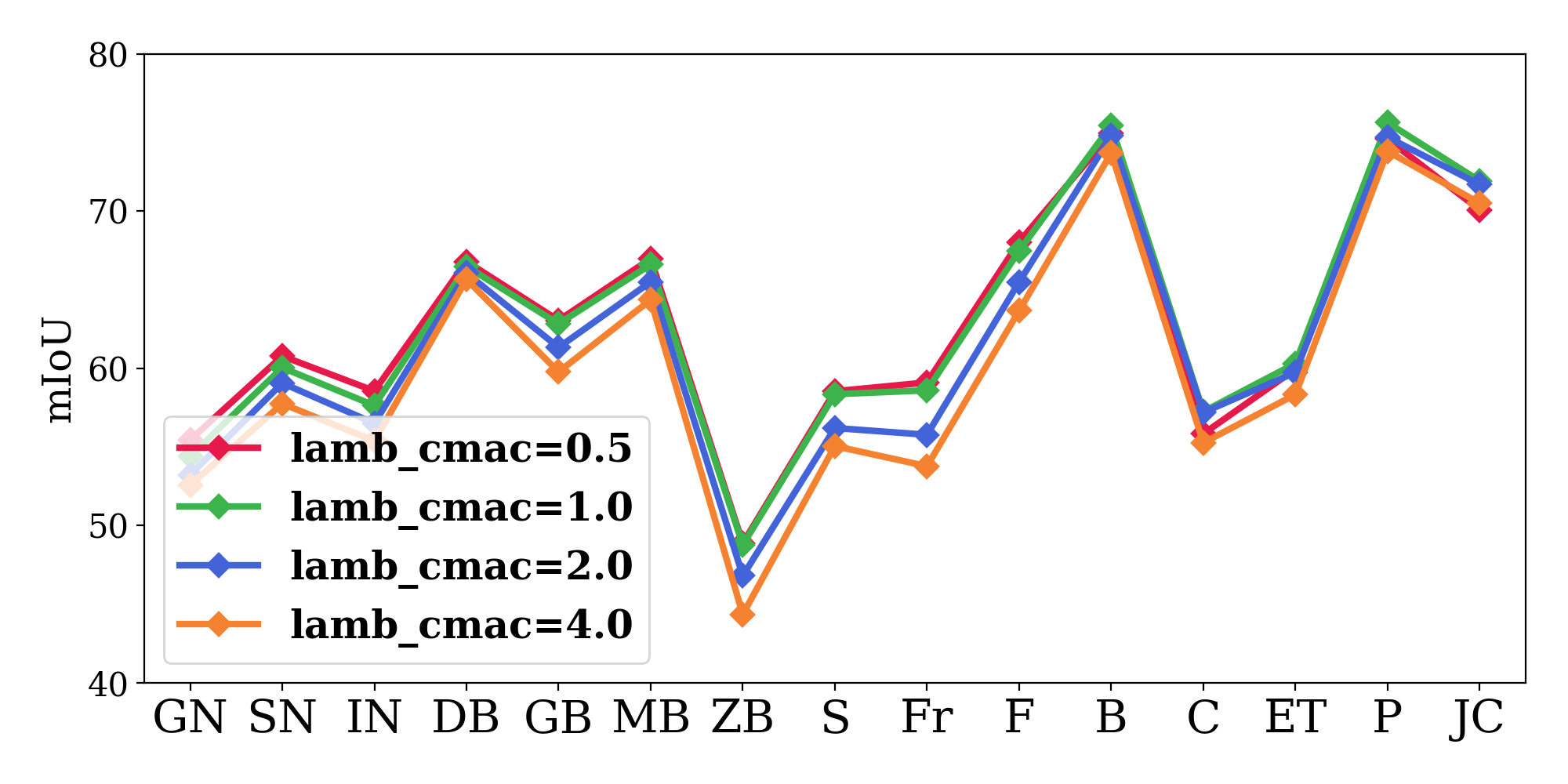}
        \caption{$\lambda_{\text{cmac}}$}
    \end{subfigure}
    \hfill
    \begin{subfigure}{0.24\textwidth}
        \centering
        \includegraphics[width=\textwidth]{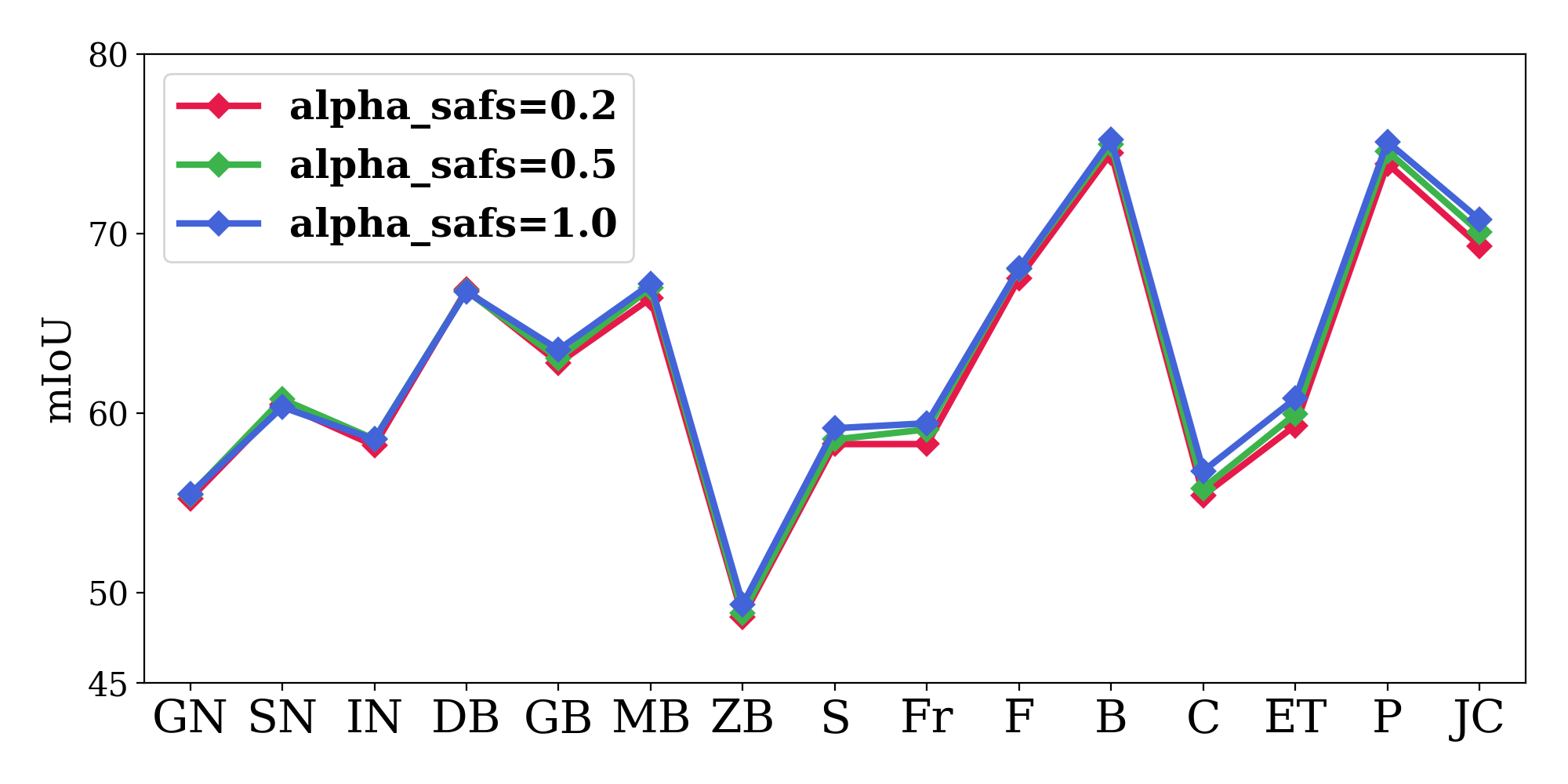}
        \caption{$\alpha_{\text{safs}}$}
    \end{subfigure}
    \caption{Ablation results of DAF-M with NACLIP~\cite{Hajimiri2024PayAT} ViT-B/32 under CTTA in VOC20-C~\cite{Everingham2014ThePV}. Entries show mIoU (\%). Each panel shows the effect of a single hyperparameter: (a) batch size, (b) $\lambda_{\text{div}}$, (c) $\lambda_{\text{cmac}}$, and (d) $\alpha_{\text{safs}}$. Results for LoveDA-C\cite{Wang2021LoveDAAR} are in the supplementary material.}
    \label{fig:voc20_hyperparams}
\end{figure*}

\paragraph{Hyperparameter analysis.}
\cref{fig:voc20_hyperparams} sweeps four hyperparameters of DAF-M on VOC20-C: batch size, $\lambda_{\text{div}}$, $\lambda_{\text{cmac}}$, and $\alpha_{\text{safs}}$. For batch size, values of 1 and 2 are unstable but 4 begins to stabilize and 8 through 10 produce nearly identical stable curves, so we use 8 as a balance between stability and memory. For $\lambda_{\text{div}}$, a value of 0.5 collapses while 1, 2, and 4 produce stable and nearly identical curves, with 2.0 achieving slightly higher mIoU, indicating that MDIV is robust to tuning once the collapse threshold is crossed. For $\lambda_{\text{cmac}}$, values of 0.5, 1, 2, and 4 are all stable with nearly identical curves, confirming that CMAC does not require careful tuning, and we default to 0.5. For $\alpha_{\text{safs}}$, values of 0.2, 0.5, and 1.0 are all stable, and we choose 0.5 as a balance between filtering redundant samples and retaining informative ones. Across all four hyperparameters, the stable ranges are broad, meaning DAF does not require careful tuning, supporting the robustness claims in \cref{sec:intro}.

\begin{figure*}[h!]
    \centering
    \begin{subfigure}{0.135\textwidth}
        \centering
        \includegraphics[width=\textwidth]{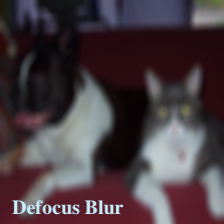}
        \caption{INPUT}
    \end{subfigure}
    \hfill
    \begin{subfigure}{0.135\textwidth}
        \centering
        \includegraphics[width=\textwidth]{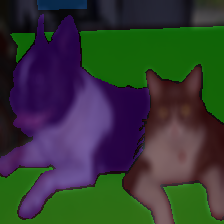}
        \caption{GT}
    \end{subfigure}
    \hfill
    \begin{subfigure}{0.135\textwidth}
        \centering
        \includegraphics[width=\textwidth]{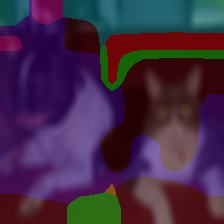}
        \caption{SOURCE}
    \end{subfigure}
    \hfill
    \begin{subfigure}{0.135\textwidth}
        \centering
        \includegraphics[width=\textwidth]{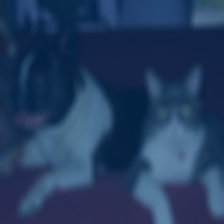}
        \caption{TENT}
    \end{subfigure}
    \hfill
    \begin{subfigure}{0.135\textwidth}
        \centering
        \includegraphics[width=\textwidth]{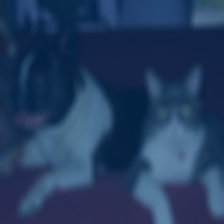}
        \caption{MLMP}
    \end{subfigure}
    \hfill
    \begin{subfigure}{0.135\textwidth}
        \centering
        \includegraphics[width=\textwidth]{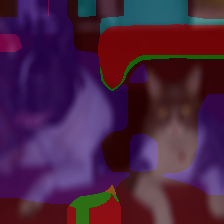}
        \caption{DAF-T}
    \end{subfigure}
    \hfill
    \begin{subfigure}{0.135\textwidth}
        \centering
        \includegraphics[width=\textwidth]{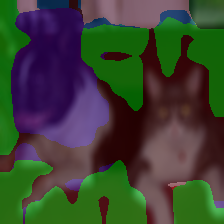}
        \caption{DAF-M}
    \end{subfigure}
    \caption{We visualize the prediction of CTTA for OVSS methods using NACLIP~\cite{Hajimiri2024PayAT} ViT-B/32 on PASCAL VOC20-C~\cite{Everingham2014ThePV}. Classes shown: dog in dark purple/blue, cat in dark maroon/brown, and sofa in green.}
    \label{fig:voc20_predictions}
\end{figure*}

\paragraph{Qualitative predictions.}
\cref{fig:voc20_predictions} visualizes segmentation maps on a VOC20-C sample under defocus blur. After continual adaptation, TENT and MLMP collapse to predicting predominantly a single class across the image, providing a direct visual confirmation of the patch-level class collapse diagnosed in \cref{sec:intro}. DAF-T and DAF-M preserve the multi-class structure of the source prediction, illustrating how the MDIV loss maintains the active class vocabulary while CMAC preserves the vision-language alignment needed for correct patch-to-text associations.

\begin{figure}[htbp]
    \centering
    \begin{subfigure}{0.48\columnwidth}
        \centering
        \includegraphics[width=\columnwidth]{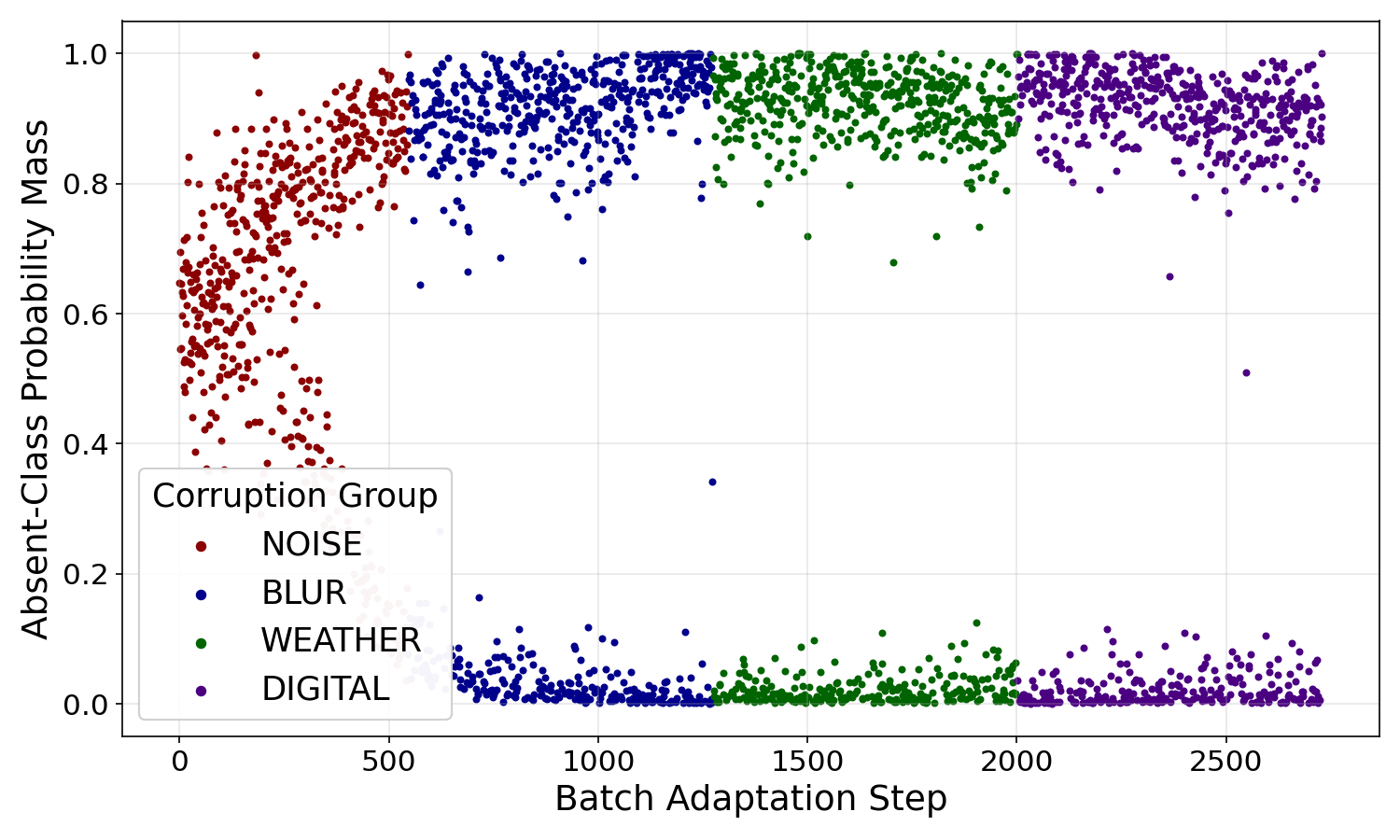}
        \caption{$\mathcal{L}_{\text{div}} = \text{False}$}
    \end{subfigure}
    \hfill
    \begin{subfigure}{0.48\columnwidth}
        \centering
        \includegraphics[width=\columnwidth]{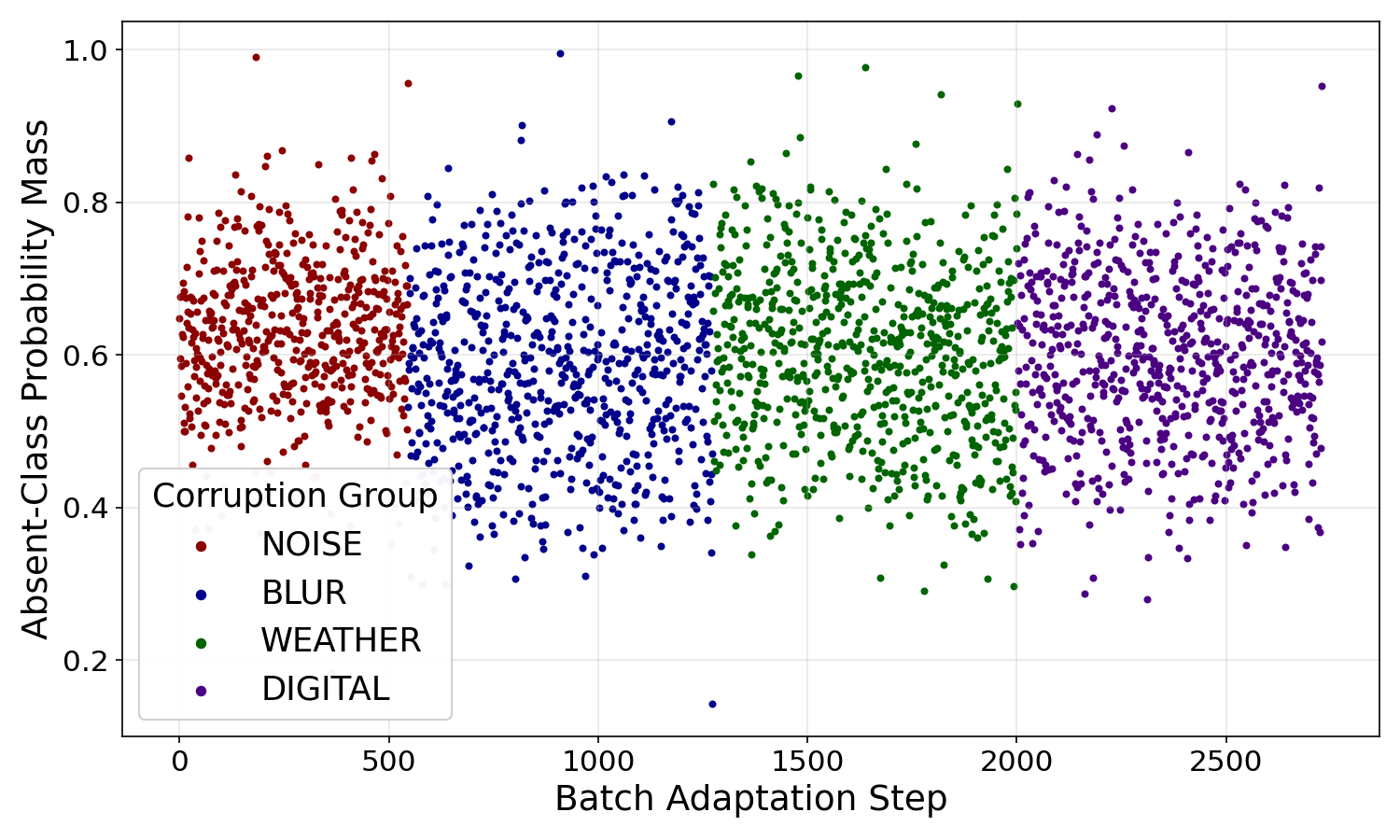}
        \caption{$\mathcal{L}_{\text{div}} = \text{True}$}
    \end{subfigure}
    \caption{Absent-class probability mass on PASCAL VOC20-C~\cite{Everingham2014ThePV} with NACLIP~\cite{Hajimiri2024PayAT} ViT-B/32, comparing adaptation without~(a) and with~(b) $\mathcal{L}_{\text{div}}$. Results on PASCAL VOC21-C~\cite{Everingham2014ThePV} are in the supplementary material.}
    \label{fig:voc20_probability_mass}
\end{figure}

\paragraph{MDIV effect.}
\cref{fig:voc20_probability_mass} examines the absent-class probability mass metric on VOC20-C, which measures how much softmax probability the model allocates to classes absent from the current batch. Without $\mathcal{L}_{\text{div}}$, the model exhibits bimodal behavior, swinging between stable predictions and near-collapse where almost all probability shifts to absent classes. With $\mathcal{L}_{\text{div}}$ enabled, these extreme swings are reduced and the diversity loss does not increase absent-class probability mass on average, confirming its role as a stabilizer of the active class vocabulary rather than a source of hallucination.

\begin{table}[htbp]
    \caption{Comparison of feature consistency loss variants under CTTA on Cityscapes-C\cite{Cordts2016TheCD} and VOC20-C \cite{Everingham2014ThePV} using NACLIP\cite{Hajimiri2024PayAT} ViT-B/32. Entries report mIoU (\%) and mDice (\%). $\mathcal{L}_{\text{cmac}}$ is our proposed Cross-modal Anchor Consistency loss (\cref{sec:cmac}). $\mathcal{L}_{\text{cosine}}$ and $\mathcal{L}_{\text{L2}}$ are baselines that penalize angular and Euclidean drift between adapted and source patch tokens, respectively. Implementation details are in supplementary material.}
    \label{tab:ctta_consistency_type_vitb32}
    \begin{center}
    \footnotesize
    \setlength\tabcolsep{4pt}
    \begin{tabular}{l|cc|cc}
        \toprule
        \multirow{2}{*}{METHOD} & \multicolumn{2}{c|}{Cityscapes-C} & \multicolumn{2}{c}{VOC20-C} \\
        \cline{2-5}
         & mIoU & mDice & mIoU & mDice \\
        \midrule
        $\mathcal{L}_{\text{cosine}}$   & 15.2 & 23.0    & 51.7 & 66.7  \\
        $\mathcal{L}_{\text{L2}}$       & 15.2 & 23.0    & 40.8 & 55.2  \\
        $\mathcal{L}_{\text{cmac}}$     & \bf18.5 & \bf27.5    & \bf62.7 & \bf76.0  \\
        \bottomrule
    \end{tabular}
    \end{center}
\end{table}

\paragraph{Consistency loss variants.}
\cref{tab:ctta_consistency_type_vitb32} compares CMAC against two consistency losses that penalize drift between adapted and source patch tokens without text anchoring. On VOC20-C, CMAC reaches 62.7 mIoU versus 51.7 for $\mathcal{L}_{\text{cosine}}$ and 40.8 for $\mathcal{L}_{\text{L2}}$. On Cityscapes-C, CMAC still leads at 18.5 versus 15.2 for both alternatives. Because CLIP features are L2-normalized, cosine and L2 distances are proportional on the unit hypersphere, so both constrain angular drift but penalize all directions equally. They cannot distinguish harmful drift away from the assigned text prototype from benign reorientation that preserves vision-language alignment. CMAC overcomes this with a one-sided hinge anchored to text prototypes, separately penalizing drift away from the assigned class and toward unassigned classes. The 11-point gap between $\mathcal{L}_{\text{cmac}}$ and $\mathcal{L}_{\text{cosine}}$ on VOC20-C confirms that text-anchored directionality, not feature alignment alone, preserves segmentation accuracy under continual adaptation.
\section{Conclusion}
\label{sec:conclusion}
We introduced CTTA for OVSS, a setting where standard entropy-based updates erode the vision-language alignment that enables flexible segmentation. Our diagnostic analysis revealed class collapse and cross-modal drift as failure modes tied to degradation of vision-language structure, along with adaptation inefficiency from redundant gradients. The proposed DAF framework addresses each through a marginal diversity (MDIV) loss, a cross-modal anchor consistency (CMAC) loss, and source-anchored feature salience (SAFS) filtering. Across the evaluated continual shifts, DAF remains stable where entropy minimization collapses, is robust to aggressive learning rates, and sustains performance under long-horizon multi-pass adaptation. DAF's stabilization relies on two assumptions: that the source vocabulary remains appropriate for the target domain, and that source-relative semantic geometry stays informative under shift. When these assumptions are violated, such as under severe vocabulary mismatch, MDIV loss may resist beneficial adaptation and CMAC loss may become overly conservative, as reflected by the weaker gains under larger domain gaps. Future work could explore source-anchored regularization for other vision-language tasks with vocabulary mismatch.

{
    \small
    \bibliographystyle{ieeenat_fullname}
    \bibliography{main}
}

\end{document}


\maketitle

\tableofcontents

\section{Preliminaries}
\label{sec:preliminaries_full}

CLIP-based OVSS models~\cite{Hajimiri2024PayAT} consist of an image encoder $g_\theta$ and a text encoder $h_\phi$ that map images and class names into a shared embedding space of dimension $D$. We adopt NACLIP~\cite{Hajimiri2024PayAT} with a ViT-B/32 backbone as our OVSS model. The inference and adaptation setup is summarized below.

\subsection{Image encoding.}
Each original test image is resized and decomposed into overlapping $224{\times}224$ crops with stride $112$. Let $M$ denote the number of original images in the current test batch and $P_m$ the number of crops extracted from image $m$. The crops from all $M$ images are concatenated into a single sample batch of size $Q = \sum_{m=1}^{M} P_m$ and processed together in one forward pass. We refer to each crop as a \emph{sample} in the remainder of this paper. Given one sample $\mathbf{x}$, the image encoder $g_\theta$ produces a sequence of $1{+}N$ token embeddings, where the first token is the CLS token and the remaining $N$ are ViT patch tokens arranged on a spatial grid of size $h{\times}w = N$. For ViT-B/32 at $224{\times}224$ input, $N = 7 \times 7 = 49$. We denote the full image feature output as $\mathbf{F} \in \mathbb{R}^{(1{+}N) \times D}$ and its patch-token subset as $\mathbf{F}_{1:N} \in \mathbb{R}^{N \times D}$, whose $i$-th row for sample $q$ we write as $\mathbf{f}_{q,i}$.

\subsection{Text encoding.}
The text encoder $h_\phi$ encodes $C$ candidate class names, where $C$ denotes the size of the inference-time text vocabulary. Because adaptation is unlabeled, these class names are supplied by the OVSS benchmark or deployment vocabulary and remain fixed during adaptation. Each class name is encoded with $R$ prompt templates, producing text prototypes $\mathbf{T}^{(r)} \in \mathbb{R}^{C \times D}$ for template $r$. Both the text encoder and the resulting text features remain frozen throughout adaptation. In MLMP, the stored text tensor also includes one prompt-averaged prototype per class, which is used in text-integration mode and during evaluation.

\subsection{Prediction logits.}
The OVSS forward pass L2-normalizes both image features and text features, then forms scaled cosine-similarity logits. For patch token $i$ of sample $q$ under template $r$, the logit for class $c$ is
\begin{equation}
s_r(q, i, c) = \tau \cdot \mathrm{cos}(\mathbf{f}_{q,i}, \mathbf{T}^{(r)}_c),
\end{equation}
where $\tau$ is the learned CLIP logit scale. The image feature tensor returned by the model still contains the CLS token, but segmentation logits are formed only from the patch tokens after removing that first token. Reshaping the remaining $N = h \times w$ logits back to the ViT grid yields a patch-grid prediction tensor of shape $(R, Q, C, h, w)$. The per-patch class probability under template $r$ is
\begin{equation}
p_r(c \mid \mathbf{x}_{q,i}) = \mathrm{softmax}(s_r(q, i, \cdot))_c.
\end{equation}
In MLMP, the CLS token can additionally produce a classification-level entropy term, but it does not enter the patch-grid segmentation logits.

\subsection{Adaptation versus evaluation.}
During adaptation, all losses are computed on these patch-grid logits before any interpolation or crop recombination. The primary training signal is template-averaged entropy minimization,
\begin{equation}
\begin{split}
\mathcal{L}_{\text{ent}} &= -\frac{1}{R \cdot Q \cdot N} \sum_{r=1}^{R} \sum_{q=1}^{Q} \sum_{i=1}^{N} \\
&\quad \times \sum_{c=1}^{C} p_r(c \mid \mathbf{x}_{q,i}) \log p_r(c \mid \mathbf{x}_{q,i}).
\end{split}
\end{equation}
In TENT, this entropy is computed for the current prompt set and then averaged over templates. In MLMP~\cite{Noori2025TestTimeAO}, the default prompt-integration mode likewise keeps the template dimension explicit, applies the segmentation entropy to each template, averages over templates, and adds a classification-level entropy term computed from CLS-token logits. In text-integration mode, and also during evaluation, MLMP instead uses the prompt-averaged text prototype for each class. Hence, the adaptation step operates on the sample batch indexed by $q = 1, \ldots, Q$, whereas image-level predictions are formed only afterward by bilinearly interpolating each sample score map to $224{\times}224$ and averaging overlapping crops back into each original resized image.

\subsection{Continual setting.}
Following TENT~\cite{Wang2021TentFT}, we adapt the model at test-time by updating only the LayerNorm parameters $\{\gamma, \beta\}$ of the image encoder $g_\theta$, while keeping all other parameters frozen. This restricts adaptation to low dimensional affine transformations, though alignment erosion remains a risk under continual adaptation over many domains. In the CTTA setting, the model encounters a sequence of corruption domains $(d_1, d_2, \ldots, d_K)$ without access to source data or labels. Adaptation is continual: the model state is carried over between domains without resetting, meaning that adaptation to domain $d_k$ starts from the model state after adapting to $d_{k-1}$. This setting is particularly challenging because error accumulation across domains can lead to severe performance degradation.

A key ingredient of our method is a frozen copy of the source model $g_{\theta_s}$, created as a deep copy of the initial model before any adaptation. We denote the adapted and source image features as $\mathbf{F}^{\text{adapt}}$ and $\mathbf{F}^{\text{src}}$, respectively. CMAC and SAFS operate on their patch-token subsets $\mathbf{F}^{\text{adapt}}_{1:N}$ and $\mathbf{F}^{\text{src}}_{1:N}$, excluding the CLS token. We write $\mathbf{f}^{\text{adapt}}_{q,i}$ and $\mathbf{f}^{\text{src}}_{q,i}$ for the $i$-th patch token of sample $q$ from these subsets.

\subsection{Per-dataset settings.}
\cref{tab:per_dataset_settings} lists the dataset-specific parameters used in all experiments. All datasets share the following common settings: NACLIP ViT-B/32 as the OVSS model, a dataloader batch size of 8 input images, one optimization step per batch, SGD optimizer with learning rate $1{\times}10^{-3}$, continual reset mode (the model state carries over between domains without resetting), and 7 prompt templates for text encoding. TENT and DAF-T use a single prompt template (\texttt{a photo of a \{\}}), while MLMP and DAF-M use 7 prompt templates following the original MLMP setup. The 15 standard corruption types follow the ImageNet-C protocol~\cite{hendrycks2019benchmarking} at severity level 5: \texttt{gaussian\_noise}, \texttt{shot\_noise}, \texttt{impulse\_noise}, \texttt{defocus\_blur}, \texttt{glass\_blur}, \texttt{motion\_blur}, \texttt{zoom\_blur}, \texttt{snow}, \texttt{frost}, \texttt{fog}, \texttt{brightness}, \texttt{contrast}, \texttt{elastic\_transform}, \texttt{pixelate}, and \texttt{jpeg\_compression}. Clean variants use a single \texttt{original} domain except where noted. The \texttt{init\_resize} column gives the height and width to which each original image is resized before crop extraction. Validation set sizes are 500 per domain for Cityscapes-C, 500 per fog level (1500 total) for Foggy Cityscapes, 992 rural and 677 urban for LoveDA, 1669 per domain for LoveDA-C, 110 for SUIM, 110 per domain for SUIM-C, 1449 for VOC20/VOC21, and 1449 per domain for VOC20-C/VOC21-C. Class extensions augment the base vocabulary with synonyms and subcategory names: VOC20 extends the person class with six attire descriptors (person in shirt, person in jeans, person in dress, person in sweater, person in skirt, person in jacket), and VOC21 adds a background class with 26 scene element synonyms (sky, wall, tree, grass, road, building, etc.) and a television monitor class with five display synonyms (television monitor, tv monitor, monitor, television, screen). After class extension, VOC20 has 30 classes and VOC21 has 56 classes. Cityscapes uses 19 classes, CityscapesFoggy uses the same 19 classes with foggy variants (light, medium, dense), LoveDA uses 7 classes, and SUIM uses 6 classes.

\begin{table}[t]
\centering
\caption{Per-dataset experimental settings. All datasets use patch size $224{\times}224$ and patch stride 112.}
\label{tab:per_dataset_settings}
\footnotesize
\begin{tabular}{lccc}
\toprule
Dataset & Init resize & Domains & Cls.\ ext. \\
\midrule
Cityscapes-C & $1120{\times}560$ & 15 corruptions & No \\
CityscapesFoggy & $1120{\times}560$ & 3 fog levels & No \\
LoveDA-C & $1024{\times}1024$ & 15 corruptions & No \\
LoveDA & $1024{\times}1024$ & 2 domains & No \\
SUIM-C & $320{\times}256$ & 15 corruptions & No \\
SUIM & $320{\times}256$ & 1 domain & No \\
VOC20-C & $224{\times}224$ & 15 corruptions & Yes \\
VOC20 & $224{\times}224$ & 1 domain & Yes \\
VOC21-C & $224{\times}224$ & 15 corruptions & Yes \\
VOC21 & $224{\times}224$ & 1 domain & Yes \\
\bottomrule
\end{tabular}
\end{table}

\section{Additional Results}
\label{sec:additional_results}

\begin{table*}[tb]
    \centering
    \begin{small}
    \caption{Benchmark results on segmentation datasets under CTTA using NACLIP\cite{Hajimiri2024PayAT} ViT-B/32. Adaptation is continuous between domains, without resetting the model. Entries report mIoU (\%), averaged over seeds 0/1/2. DAF-T and DAF-M denote our method (Diversify, Anchor, and Filter) applied on top of TENT and MLMP, respectively. "--" indicates OOM.}
    \label{tab:ctta_benchmark_fine_vitb32}
    \resizebox{0.95\textwidth}{!}{
    \begin{tabular}{ll|c|ccc|cccccc}
    \toprule
    \multicolumn{2}{c|}{\multirow{3}{*}{Dataset}} & \multicolumn{10}{c}{Adaptation Method} \\
    \cmidrule(lr){3-12}
    \multicolumn{2}{c|}{} & \multirow{2}{*}{SOURCE} & \multicolumn{3}{c|}{Non-VLM} & \multicolumn{6}{c}{VLM} \\
    \cmidrule(lr){4-6} \cmidrule(lr){7-12}
    \multicolumn{2}{c|}{} &  & TENT\cite{Wang2021TentFT} & COTTA\cite{Wangetal2022cotta} & SAR\cite{niu2023sar} & WATT\cite{Osowiechi2024WATTWA} & CLIPArTT\cite{Hakim2024CLIPArTTAO} & SegTTO\cite{Silva2025TestTimeOF} & MLMP\cite{Noori2025TestTimeAO} & DAF-T & DAF-M \\
    \midrule

    \multicolumn{2}{l|}{VOC20\cite{Everingham2014ThePV} (Original)} & 72.4 & 74.2\ppm0.06 & 72.3\ppm0.02 & 73.6\ppm0.01 & 66.6\ppm0.11 & 19.1\ppm0.51 & 39.1\ppm0.63 & 76.6\ppm0.01 & 73.1\ppm0.07 & 78.2\ppm0.02 \\
    \midrule
    \parbox[t]{2mm}{\multirow{16}{*}{\rotatebox[origin=c]{90}{VOC20-C\cite{Everingham2014ThePV}}}}
    & Gaussian Noise       & 47.6 & 37.2\ppm0.26 & 47.1\ppm0.06 & 40.8\ppm0.15 & 41.9\ppm0.56 & 17.3\ppm0.55 & 19.1\ppm1.22 & 52.7\ppm1.61 & 48.9\ppm0.05 & 55.1\ppm0.31 \\
    & Shot Noise           & 51.8 & 7.2\ppm0.07 & 49.5\ppm0.02 & 20.5\ppm1.81 & 44.9\ppm0.01 & 1.7\ppm0.03 & 2.5\ppm0.22 & 42.3\ppm10.37 & 54.4\ppm0.04 & 60.3\ppm0.53 \\
    & Impulse Noise        & 48.8 & 1.0\ppm0.03 & 44.6\ppm0.13 & 31.8\ppm1.69 & 42.5\ppm0.75 & 1.4\ppm0.13 & 1.9\ppm0.14 & 21.3\ppm19.30 & 51.9\ppm0.09 & 58.5\ppm0.11 \\
    & Defocus Blur         & 60.2 & 1.4\ppm0.06 & 56.6\ppm0.13 & 59.0\ppm0.62 & 49.4\ppm0.26 & 0.9\ppm0.04 & 1.5\ppm0.08 & 18.6\ppm18.28 & 62.3\ppm0.03 & 67.4\ppm0.63 \\
    & Glass Blur           & 54.6 & 0.3\ppm0.03 & 51.7\ppm0.39 & 57.9\ppm0.07 & 41.6\ppm0.12 & 0.9\ppm0.14 & 1.1\ppm0.01 & 6.6\ppm6.35 & 57.8\ppm0.07 & 63.2\ppm0.16 \\
    & Motion Blur          & 59.6 & 0.4\ppm0.03 & 53.4\ppm0.40 & 55.4\ppm1.39 & 49.3\ppm0.07 & 0.8\ppm0.09 & 1.1\ppm0.06 & 0.9\ppm0.73 & 62.3\ppm0.13 & 66.9\ppm0.08 \\
    & Zoom Blur            & 38.7 & 0.2\ppm0.00 & 30.2\ppm0.68 & 26.6\ppm0.41 & 27.8\ppm0.08 & 0.6\ppm0.12 & 1.0\ppm0.03 & 0.3\ppm0.14 & 42.0\ppm0.18 & 49.1\ppm0.22 \\
    & Snow                 & 49.2 & 0.2\ppm0.02 & 40.1\ppm0.49 & 51.4\ppm0.14 & 39.7\ppm0.56 & 0.7\ppm0.08 & 1.0\ppm0.02 & 0.3\ppm0.16 & 53.4\ppm0.07 & 57.9\ppm0.64 \\
    & Frost                & 47.6 & 0.2\ppm0.01 & 38.2\ppm0.61 & 50.1\ppm0.48 & 36.4\ppm0.26 & 0.9\ppm0.13 & 1.0\ppm0.01 & 0.4\ppm0.20 & 52.1\ppm0.24 & 58.8\ppm0.27 \\
    & Fog                  & 56.2 & 0.2\ppm0.01 & 42.1\ppm3.69 & 47.3\ppm0.43 & 50.7\ppm0.25 & 0.7\ppm0.20 & 1.0\ppm0.03 & 0.2\ppm0.09 & 60.5\ppm0.02 & 67.8\ppm0.28 \\
    & Brightness           & 68.0 & 0.5\ppm0.03 & 56.6\ppm2.94 & 46.1\ppm2.97 & 61.6\ppm0.62 & 1.0\ppm0.25 & 1.1\ppm0.02 & 0.5\ppm0.27 & 70.1\ppm0.02 & 75.0\ppm0.04 \\
    & Contrast             & 48.8 & 0.2\ppm0.01 & 29.1\ppm4.26 & 38.0\ppm5.29 & 33.5\ppm0.24 & 0.7\ppm0.09 & 1.0\ppm0.03 & 0.2\ppm0.08 & 51.9\ppm0.12 & 56.0\ppm0.22 \\
    & Elastic Transform    & 52.7 & 0.1\ppm0.01 & 32.5\ppm3.90 & 57.0\ppm0.23 & 42.3\ppm0.46 & 0.9\ppm0.21 & 1.0\ppm0.00 & 0.2\ppm0.06 & 56.1\ppm0.19 & 60.4\ppm0.45 \\
    & Pixelate             & 68.6 & 0.3\ppm0.00 & 49.5\ppm5.94 & 71.6\ppm0.18 & 63.1\ppm0.10 & 0.8\ppm0.10 & 1.0\ppm0.00 & 0.2\ppm0.02 & 70.5\ppm0.06 & 74.5\ppm0.05 \\
    & JPEG Compression     & 63.9 & 0.3\ppm0.01 & 41.8\ppm4.65 & 65.7\ppm0.80 & 58.5\ppm0.31 & 0.9\ppm0.01 & 1.0\ppm0.01 & 0.2\ppm0.06 & 65.8\ppm0.07 & 70.2\ppm0.15 \\
    \cmidrule{2-12}
    \rowcolor{gray!15} & Mean    & 54.4 & 3.3\ppm0.03 & 44.2\ppm1.58 & 47.9\ppm0.14 & 45.5\ppm0.12 & 2.0\ppm0.06 & 2.4\ppm0.12 & 9.7\ppm3.85 & 57.3\ppm0.06 & 62.7\ppm0.03 \\
    \midrule
    
    \multicolumn{2}{l|}{VOC21\cite{Everingham2014ThePV} (Original)} & 41.3 & 42.0\ppm0.02 & 41.3\ppm0.00 & 41.9\ppm0.05 & 41.0\ppm0.03 & 17.5\ppm1.07 & 41.2\ppm0.38 & 42.9\ppm0.01 & 41.6\ppm0.02 & 43.5\ppm0.02 \\
    \rowcolor{gray!15}\multicolumn{2}{l|}{VOC21-C\cite{Everingham2014ThePV}} & 33.9 & 5.0\ppm0.76 & 29.4\ppm0.79 & 33.7\ppm0.03 & 28.7\ppm0.01 & 3.5\ppm0.25 & 1.0\ppm0.05 & 9.4\ppm0.96 & 35.1\ppm0.02 & 36.1\ppm0.01 \\
    \midrule
    
    \parbox[t]{8mm}{\multirow{3}{*}{\rotatebox[origin=c]{0}{\parbox{8mm}{Foggy\newline Cityscapes\cite{Sakaridis2017SemanticFS}}}}}
    & Light Fog     & 27.5 & 27.8\ppm0.01 & 27.5\ppm0.00 & 27.9\ppm0.01 & 24.6\ppm0.03 & -- & 31.4\ppm0.14 & 30.7\ppm0.00 & 27.8\ppm0.01 & 30.6\ppm0.02 \\
    & Medium Fog    & 27.0 & 27.6\ppm0.01 & 26.9\ppm0.00 & 27.9\ppm0.01 & 24.5\ppm0.19 & -- & 26.5\ppm0.84 & 28.9\ppm0.06 & 27.5\ppm0.00 & 30.0\ppm0.02 \\
    & Dense Fog     & 25.7 & 26.2\ppm0.02 & 25.4\ppm0.03 & 26.7\ppm0.01 & 19.2\ppm0.13 & -- & 21.9\ppm1.19 & 27.4\ppm0.04 & 26.3\ppm0.01 & 28.7\ppm0.01 \\
    \cmidrule{2-12}
    \rowcolor{gray!15} & Mean   & 26.7 & 27.2\ppm0.01 & 26.6\ppm0.01 & 27.5\ppm0.01 & 22.8\ppm0.12 & -- & 26.6\ppm0.63 & 29.0\ppm0.03 & 27.2\ppm0.01 & 29.8\ppm0.00 \\
    \midrule
    \rowcolor{gray!15}\multicolumn{2}{l|}{Cityscapes-C\cite{Cordts2016TheCD}} & 17.8 & 4.5\ppm0.01 & 16.7\ppm0.52 & 17.1\ppm0.02 & 7.7\ppm0.00 & -- & 2.9\ppm0.01 & 1.6\ppm0.01 & 19.0\ppm0.01 & 18.5\ppm0.01 \\
    \midrule
    
    \multicolumn{2}{l|}{SUIM\cite{islam2020semantic} (Original)} & 22.4 & 22.5\ppm0.02 & 22.4\ppm0.00 & 22.5\ppm0.01 & 21.9\ppm0.36 & 22.5\ppm0.01 & 23.9\ppm0.20 & 12.9\ppm0.02 & 22.5\ppm0.02 & 13.0\ppm0.04 \\
    \rowcolor{gray!15}\multicolumn{2}{l|}{SUIM-C\cite{islam2020semantic}} & 19.6 & 19.5\ppm0.08 & 19.4\ppm0.01 & 19.1\ppm0.07 & 18.5\ppm0.07 & 12.7\ppm0.15 & 12.0\ppm0.02 & 10.6\ppm0.02 & 20.3\ppm0.00 & 11.9\ppm0.02 \\
    \midrule

    \parbox[t]{15mm}{\multirow{2}{*}{\rotatebox[origin=c]{0}{LoveDA\cite{Wang2021LoveDAAR}}}}
    & Rural                & 16.3 & 16.8\ppm0.12 & 16.1\ppm0.00 & 17.2\ppm0.12 & 19.2\ppm0.07 & -- & 11.8\ppm0.03 & 22.0\ppm0.30 & 17.2\ppm0.02 & 28.1\ppm0.11 \\
    & Urban                & 29.2 & 23.8\ppm0.27 & 28.1\ppm0.04 & 23.4\ppm0.25 & 25.6\ppm0.06 & -- & 3.8\ppm0.02 & 11.4\ppm2.69 & 31.0\ppm0.03 & 35.5\ppm0.08 \\
    \cmidrule{2-12}
    \rowcolor{gray!15} & Mean    & 22.7 & 20.3\ppm0.19 & 22.1\ppm0.02 & 20.3\ppm0.18 & 22.4\ppm0.06 & -- & 7.8\ppm0.02 & 16.7\ppm1.50 & 24.1\ppm0.00 & 31.8\ppm0.02 \\
    \midrule
    \rowcolor{gray!15}\multicolumn{2}{l|}{LoveDA-C\cite{Wang2021LoveDAAR}} & 9.2 & 1.2\ppm0.00 & -- & 7.6\ppm0.15 & -- & -- & 5.2\ppm0.00 & 5.2\ppm0.00 & 12.3\ppm0.01 & 14.9\ppm0.01 \\

    \bottomrule    
    \end{tabular}
    }
    \end{small}
\end{table*}

\begin{table*}[tb]
    \centering
    \begin{small}
    \caption{Benchmark results on segmentation datasets under CTTA using NACLIP\cite{Hajimiri2024PayAT} ViT-B/16. Adaptation is continuous between domains, without resetting the model. Entries report mIoU (\%), averaged over seeds 0/1/2. DAF-T and DAF-M denote our method (Diversify, Anchor, and Filter) applied on top of TENT and MLMP, respectively. "--" indicates OOM.}
    \label{tab:ctta_benchmark_fine_vitb16}
    \resizebox{0.95\textwidth}{!}{
    \begin{tabular}{ll|c|cccc|cc}
    \toprule
    \multicolumn{2}{l|}{OVSS Backbone: NACLIP\cite{Hajimiri2024PayAT} ViT-B/16} &  \multicolumn{6}{c}{Adaptation Method} \\ \cmidrule{1-9}
    \multicolumn{2}{l|}{Dataset} & SOURCE & TENT\cite{Wang2021TentFT} & WATT\cite{Osowiechi2024WATTWA} & CLIPArTT\cite{Hakim2024CLIPArTTAO} & MLMP\cite{Noori2025TestTimeAO} & DAF-T & DAF-M \\
    \midrule

    \multicolumn{2}{l|}{VOC20\cite{Everingham2014ThePV} (Original)} & 77.6 & 79.3\ppm0.8 & 47.8\ppm0.5 & 20.7\ppm0.7 & 82.3\ppm0.1 & 79.4\ppm0.1 & 82.2\ppm0.2 \\
    \midrule
    \parbox[t]{2mm}{\multirow{16}{*}{\rotatebox[origin=c]{90}{VOC20-C\cite{Everingham2014ThePV}}}}
    & Gaussian Noise       & 48.0 & 36.8\ppm1.5 & 16.5\ppm0.3 & 9.8\ppm0.3 & 59.9\ppm0.9 & 51.2\ppm0.1 & 56.4\ppm0.3 \\
    & Shot Noise           & 52.4 & 3.7\ppm1.6 & 16.1\ppm0.3 & 2.8\ppm0.9 & 63.1\ppm2.6 & 56.3\ppm0.1 & 61.9\ppm0.3 \\
    & Impulse Noise        & 49.5 & 1.0\ppm0.4 & 16.2\ppm0.2 & 1.8\ppm0.4 & 17.8\ppm5.5 & 53.5\ppm0.2 & 58.2\ppm0.3 \\
    & Defocus Blur         & 68.0 & 7.0\ppm1.6 & 41.3\ppm0.6 & 3.8\ppm0.2 & 5.3\ppm0.9 & 70.3\ppm0.1 & 73.6\ppm0.6 \\
    & Glass Blur           & 62.1 & 0.3\ppm0.2 & 32.6\ppm0.2 & 1.6\ppm0.4 & 1.0\ppm0.1 & 65.6\ppm0.2 & 71.2\ppm0.3 \\
    & Motion Blur          & 69.5 & 0.5\ppm0.2 & 41.4\ppm0.5 & 1.9\ppm0.3 & 1.0\ppm0.0 & 71.9\ppm0.2 & 74.0\ppm0.1 \\
    & Zoom Blur            & 47.3 & 0.2\ppm0.1 & 19.7\ppm0.2 & 1.4\ppm0.1 & 1.0\ppm0.0 & 52.5\ppm0.1 & 57.2\ppm0.6 \\
    & Snow                 & 60.9 & 0.3\ppm0.2 & 28.2\ppm0.3 & 1.3\ppm0.4 & 1.0\ppm0.0 & 65.6\ppm0.3 & 69.2\ppm0.5 \\
    & Frost                & 55.4 & 0.6\ppm0.3 & 23.6\ppm0.3 & 1.2\ppm0.3 & 1.0\ppm0.0 & 60.7\ppm0.4 & 65.1\ppm0.3 \\
    & Fog                  & 67.0 & 1.7\ppm1.0 & 38.4\ppm0.3 & 2.4\ppm0.6 & 1.2\ppm0.0 & 71.2\ppm0.5 & 75.1\ppm0.1 \\
    & Brightness           & 73.3 & 0.6\ppm0.3 & 39.1\ppm0.1 & 1.9\ppm0.4 & 1.1\ppm0.1 & 76.6\ppm0.4 & 78.8\ppm0.2 \\
    & Contrast             & 60.3 & 0.9\ppm0.3 & 33.3\ppm0.3 & 1.5\ppm0.5 & 1.2\ppm0.1 & 63.7\ppm0.2 & 69.4\ppm0.3 \\
    & Elastic Transform    & 50.1 & 0.1\ppm0.0 & 20.0\ppm0.6 & 0.7\ppm0.2 & 1.0\ppm0.0 & 55.1\ppm0.3 & 60.8\ppm0.7 \\
    & Pixelate             & 75.5 & 0.2\ppm0.1 & 49.3\ppm0.1 & 1.1\ppm0.4 & 1.0\ppm0.0 & 77.4\ppm0.1 & 79.0\ppm0.2 \\
    & JPEG Compression     & 69.1 & 0.2\ppm0.0 & 35.4\ppm0.1 & 0.9\ppm0.1 & 1.0\ppm0.0 & 71.2\ppm0.1 & 74.2\ppm0.3 \\ 
    \cmidrule{2-9}
    \rowcolor{gray!15} & Mean              & 60.6 & 3.6 & 30.1 & 2.3 & 10.5 & 64.2 & 68.3 \\
    \midrule

    \multicolumn{2}{l|}{VOC21\cite{Everingham2014ThePV} (Original)} & 51.0 & 49.8\ppm1.4 & 38.9\ppm9.1 & 29.0\ppm16.1 & 50.6\ppm0.9 & 50.9\ppm0.7 & 51.9\ppm0.1 \\
    \rowcolor{gray!15}\multicolumn{2}{l|}{VOC21-C\cite{Everingham2014ThePV}} & 42.4 & 16.9\ppm2.8 & 19.8\ppm0.0 & 1.2\ppm0.1 & 8.3\ppm0.3 & 42.3\ppm0.0 & 43.1\ppm0.1 \\
    \midrule
    
    \multicolumn{2}{l|}{SUIM\cite{islam2020semantic} (Original)} & 31.4 & 31.5\pm0.1 & 24.5\pm0.8 & 31.4\pm0.2 & 22.3\pm0.0 & 31.7\pm0.1 & 22.3\pm0.0 \\
    \rowcolor{gray!15}\multicolumn{2}{l|}{SUIM-C\cite{islam2020semantic} Average} & 24.1 & 21.8\ppm1.1 & 17.4\ppm0.1 & 13.2\ppm0.3 & 13.9\ppm0.1 & 25.1\ppm0.2 & 17.9\ppm0.1 \\
    \midrule

    \parbox[t]{15mm}{\multirow{2}{*}{\rotatebox[origin=c]{0}{LoveDA\cite{Wang2021LoveDAAR}}}}
    & Rural                & 18.4 & 15.3\ppm0.1 & -- & -- & 16.9\ppm0.2 & 19.3\ppm0.0 & 22.9\ppm0.0 \\
    & Urban                & 29.4 & 15.0\ppm0.5 & -- & -- & 20.4\ppm0.1 & 31.4\ppm0.0 & 35.9\ppm0.0 \\
    \cmidrule{2-9}
    \rowcolor{gray!15} & Mean    & 23.9 & 15.1\ppm0.3 & -- & -- & 18.7\ppm0.2 & 25.4\ppm0.0 & 29.4\ppm0.0 \\
    \midrule
    \rowcolor{gray!15}\multicolumn{2}{l|}{LoveDA-C\cite{Wang2021LoveDAAR}} & 11.6 & 1.1\ppm0.0 & -- & -- & 5.2\ppm0.0 & 14.7\ppm0.0 & 16.0\ppm0.0 \\
    
    \bottomrule
    
    \end{tabular}
    }
    \end{small}
\end{table*}

\begin{table*}[tb]
    \centering
    \begin{small}
    \caption{Benchmark results on segmentation datasets under CTTA using NACLIP\cite{Hajimiri2024PayAT} ViT-L/14. Adaptation is continuous between domains, without resetting the model. Entries report mIoU (\%), averaged over seeds 0/1/2. DAF-T and DAF-M denote our method (Diversify, Anchor, and Filter) applied on top of TENT and MLMP, respectively. "--" indicates OOM.}
    \label{tab:ctta_benchmark_fine_vitl14}
    \resizebox{0.95\textwidth}{!}{
    \begin{tabular}{ll|c|cccc|cc}
    \toprule
    \multicolumn{2}{l|}{OVSS Backbone: NACLIP\cite{Hajimiri2024PayAT} ViT-L/14} &  \multicolumn{6}{c}{Adaptation Method} \\ \cmidrule{1-9}
    \multicolumn{2}{l|}{Dataset} & SOURCE & TENT\cite{Wang2021TentFT} & WATT\cite{Osowiechi2024WATTWA} & CLIPArTT\cite{Hakim2024CLIPArTTAO} & MLMP\cite{Noori2025TestTimeAO} & DAF-T & DAF-M \\
    \midrule

    \multicolumn{2}{l|}{VOC20\cite{Everingham2014ThePV} (Original)} & 75.9\ppm0.0 & 77.9\ppm0.1 & 75.1\ppm0.1 & 41.1\ppm0.6 & 81.1\ppm0.1 & 76.8\ppm0.0 & 80.3\ppm0.0 \\
    \midrule
    \parbox[t]{2mm}{\multirow{16}{*}{\rotatebox[origin=c]{90}{VOC20-C\cite{Everingham2014ThePV}}}}
    & Gaussian Noise       & 62.9\ppm0.0 & 64.8\ppm0.2 & -- & 32.2\ppm10.3 & 67.3\ppm0.3 & 63.8\ppm0.0 & 67.2\ppm0.1 \\
    & Shot Noise           & 66.2\ppm0.0 & 67.4\ppm0.1 & -- & 2.7\ppm0.7 & 71.0\ppm0.0 & 68.0\ppm0.1 & 72.5\ppm0.0 \\
    & Impulse Noise        & 63.1\ppm0.0 & 59.9\ppm0.5 & -- & 3.6\ppm0.6 & 60.2\ppm0.0 & 65.4\ppm0.0 & 70.6\ppm0.0 \\
    & Defocus Blur         & 72.6\ppm0.0 & 53.4\ppm0.2 & -- & 4.1\ppm0.2 & 58.5\ppm1.8 & 75.0\ppm0.0 & 79.5\ppm0.1 \\
    & Glass Blur           & 71.4\ppm0.0 & 38.8\ppm3.1 & -- & 3.5\ppm2.0 & 25.1\ppm3.3 & 73.8\ppm0.0 & 79.1\ppm0.1 \\
    & Motion Blur          & 73.1\ppm0.0 & 20.2\ppm7.6 & -- & 1.3\ppm0.3 & 5.2\ppm0.3 & 75.5\ppm0.2 & 80.3\ppm0.1 \\
    & Zoom Blur            & 59.0\ppm0.0 & 2.2\ppm0.3 & -- & 1.3\ppm0.1 & 0.6\ppm0.1 & 61.4\ppm0.1 & 67.7\ppm0.1 \\
    & Snow                 & 71.5\ppm0.0 & 2.3\ppm1.4 & -- & 1.0\ppm0.2 & 1.6\ppm0.5 & 73.6\ppm0.0 & 80.1\ppm0.1 \\
    & Frost                & 65.4\ppm0.0 & 1.6\ppm1.1 & -- & 0.8\ppm0.1 & 0.6\ppm0.2 & 67.4\ppm0.0 & 72.0\ppm0.1 \\
    & Fog                  & 70.7\ppm0.0 & 3.4\ppm2.5 & -- & 0.2\ppm0.1 & 0.7\ppm0.3 & 73.2\ppm0.1 & 80.0\ppm0.1 \\
    & Brightness           & 74.9\ppm0.0 & 1.9\ppm1.4 & -- & 0.2\ppm0.1 & 0.4\ppm0.1 & 76.9\ppm0.0 & 83.0\ppm0.0 \\
    & Contrast             & 71.5\ppm0.0 & 0.8\ppm0.5 & -- & 0.9\ppm0.0 & 0.3\ppm0.0 & 73.7\ppm0.1 & 79.5\ppm0.0 \\
    & Elastic Transform    & 62.9\ppm0.0 & 0.4\ppm0.1 & -- & 0.6\ppm0.4 & 0.2\ppm0.0 & 65.3\ppm0.1 & 71.2\ppm0.0 \\
    & Pixelate             & 77.3\ppm0.0 & 0.3\ppm0.0 & -- & 0.5\ppm0.4 & 0.2\ppm0.0 & 79.2\ppm0.0 & 84.0\ppm0.0 \\
    & JPEG Compression     & 72.6\ppm0.0 & 0.3\ppm0.0 & -- & 0.7\ppm0.5 & 0.2\ppm0.0 & 74.5\ppm0.0 & 80.0\ppm0.1 \\
    \cmidrule{2-9}
    \rowcolor{gray!15} & Mean    & 69.0\ppm0.0 & 21.2\ppm0.3 & -- & 3.6\ppm0.7 & 19.5\ppm0.3 & 71.1\ppm0.0 & 76.4\ppm0.0 \\
    \midrule

    \multicolumn{2}{l|}{VOC21\cite{Everingham2014ThePV} (Original)} & 45.1\ppm0.0 & 45.6\ppm0.0 & 45.8\ppm0.0 & 30.1\ppm0.3 & 49.9\ppm0.1 & 45.5\ppm0.0 & 49.7\ppm0.1 \\
    \rowcolor{gray!15}\multicolumn{2}{l|}{VOC21-C\cite{Everingham2014ThePV}} & 40.8\ppm0.0 & 23.5\ppm0.6 & -- & 4.5\ppm0.4 & 26.0\ppm1.9 & 41.6\ppm0.0 & 45.7\ppm0.0 \\

    \bottomrule
    
    \end{tabular}
    }
    \end{small}
\end{table*}

\subsection{Backbone generalization.}
Tables~\ref{tab:ctta_benchmark_fine_vitb32}, \ref{tab:ctta_benchmark_fine_vitb16}, and \ref{tab:ctta_benchmark_fine_vitl14} extend the benchmark to NACLIP ViT-B/16 and ViT-L/14. On VOC20-C, the collapse pattern is consistent across all three backbones: TENT drops to 3.3, 3.6, and 21.2 respectively, while MLMP reaches 9.7, 10.5, and 19.5, all far below their corresponding source means of 54.4, 60.6, and 69.0. DAF-M remains stable at 62.7, 68.3, and 76.4, exceeding source in every case. The persistence of collapse in larger backbones confirms that class collapse and alignment erosion are fundamental to CTTA, not artifacts of limited model capacity, as diagnosed in the main paper. DAF's stabilization scales with backbone capability: larger backbones yield higher absolute mIoU, but the relative gain pattern holds, validating that the MDIV and CMAC address failure modes that are architectural-agnostic.

\begin{table*}[tb]
    \centering
    \begin{small}
    \caption{Comparison against traditional non-VLM CTTA methods on segmentation datasets using NACLIP\cite{Hajimiri2024PayAT} ViT-B/32. Adaptation is continuous between domains, without resetting the model. Entries report mIoU (\%). DAF-T and DAF-M denote our method (Diversify, Anchor, and Filter) applied on top of TENT and MLMP, respectively. "--" indicates OOM.}
    \label{tab:ctta_tradbenchmark_fine_vitb32}
    \resizebox{0.95\textwidth}{!}{
    \begin{tabular}{ll|c|cccccc|cc}
    \toprule
    \multicolumn{2}{l|}{OVSS Backbone: NACLIP\cite{Hajimiri2024PayAT} ViT-B/32} &  \multicolumn{8}{c}{Adaptation Method} \\ \cmidrule{1-11}
    \multicolumn{2}{l|}{Dataset} & SOURCE & TENT\cite{Wang2021TentFT} & COTTA\cite{Wangetal2022cotta} & RPL\cite{Rusak2021IfYD} & SAR\cite{niu2023sar} & DEYO\cite{Lee2024EntropyIN} & M2A\cite{doloriel2026family} & DAF-T & DAF-M \\
    \midrule

    \multicolumn{2}{l|}{VOC20\cite{Everingham2014ThePV} (Original)} & 72.4 & 74.2 & 72.3 & 73.4 & 73.6 & 73.8 & 73.5 & 73.1 & 78.2 \\
    \midrule
    \parbox[t]{2mm}{\multirow{16}{*}{\rotatebox[origin=c]{90}{VOC20-C\cite{Everingham2014ThePV}}}}
    & Gaussian Noise       & 47.6 & 37.2 & 47.1 & 43.9 & 40.8 & 28.0 & 47.7 & 48.9 & 55.1 \\
    & Shot Noise           & 51.8 & 7.2 & 49.5 & 39.1 & 44.9 & 5.9 & 31.8 & 54.4 & 60.3 \\
    & Impulse Noise        & 48.8 & 1.0 & 44.6 & 23.6 & 42.5 & 0.5 & 1.5 & 51.9 & 58.5 \\
    & Defocus Blur         & 60.2 & 1.4 & 56.6 & 23.2 & 49.4 & 0.5 & 0.3 & 62.3 & 67.4 \\
    & Glass Blur           & 54.6 & 0.3 & 51.7 & 2.0 & 41.6 & 0.1 & 0.2 & 57.8 & 63.2 \\
    & Motion Blur          & 59.6 & 0.4 & 53.4 & 0.5 & 49.3 & 0.1 & 0.2 & 62.3 & 66.9 \\
    & Zoom Blur            & 38.7 & 0.2 & 30.2 & 0.2 & 27.8 & 0.1 & 0.2 & 42.0 & 49.1 \\
    & Snow                 & 49.2 & 0.2 & 40.1 & 0.5 & 39.7 & 0.1 & 0.2 & 53.4 & 57.9 \\
    & Frost                & 47.6 & 0.2 & 38.2 & 0.4 & 36.4 & 0.1 & 0.1 & 52.1 & 58.8 \\
    & Fog                  & 56.2 & 0.2 & 42.1 & 0.2 & 50.7 & 0.4 & 0.2 & 60.5 & 67.8 \\
    & Brightness           & 68.0 & 0.5 & 56.6 & 1.5 & 61.6 & 0.3 & 0.4 & 70.1 & 75.0 \\
    & Contrast             & 48.8 & 0.2 & 29.1 & 0.3 & 33.5 & 0.2 & 0.2 & 51.9 & 56.0 \\
    & Elastic Transform    & 52.7 & 0.1 & 32.5 & 0.1 & 42.3 & 0.1 & 0.2 & 56.1 & 60.4 \\
    & Pixelate             & 68.6 & 0.3 & 49.5 & 0.2 & 63.1 & 0.1 & 0.2 & 70.5 & 74.5 \\
    & JPEG Compression     & 63.9 & 0.3 & 41.8 & 0.4 & 58.5 & 0.3 & 0.2 & 65.8 & 70.2 \\
    \cmidrule{2-11}
    \rowcolor{gray!15} & Mean    & 54.4 & 3.3 & 44.2 & 9.1 & 45.5 & 2.5 & 5.6 & 57.3 & 62.7 \\
    \midrule

    \parbox[t]{15mm}{\multirow{2}{*}{\rotatebox[origin=c]{0}{LoveDA\cite{Wang2021LoveDAAR}}}}
    & Rural                & 16.3 & 16.8 & 16.1 & 17.5 & 19.2 & 13.4 & 16.7 & 17.2 & 28.1 \\
    & Urban                & 29.2 & 23.8 & 28.1 & 27.8 & 25.6 & 7.1 & 21.2 & 31.0 & 35.5 \\
    \cmidrule{2-11}
    \rowcolor{gray!15} & Mean    & 22.7 & 20.3 & 22.1 & 22.6 & 22.4 & 10.3 & 19.0 & 24.1 & 31.8 \\
    \bottomrule
    
    \end{tabular}
    }
    \end{small}
\end{table*}

\subsection{Non-VLM baselines.}
\cref{tab:ctta_tradbenchmark_fine_vitb32} compares DAF against traditional CTTA methods originally designed for closed-vocabulary settings. On VOC20-C, COTTA and SAR are the most resilient with mean mIoU of 42.6 and 48.1, though both remain below the source of 54.4. RPL, DEYO, and M2A collapse to 9.1, 2.5, and 5.6, similar to TENT's 3.3. On LoveDA, COTTA and SAR barely match the source of 22.7, while DAF-M reaches 31.8. Their collapse confirms that the class collapse mechanism is distinct to OVSS and requires explicit stabilization of the active class vocabulary.

\begin{table*}[htbp]
    \caption{Domain generalization (forward transfer) using NACLIP\cite{Hajimiri2024PayAT} ViT-B/32. We perform CTTA on PASCAL VOC20-C \cite{Everingham2014ThePV} and Cityscapes-C \cite{Cordts2016TheCD} domains except for the last 5 domains. We directly evaluate on them using the adapted weights. Entries report mIoU (\%). "-" indicates OOM.}
    \label{tab:ctta_domain_gen_vitb32}
    \begin{center}
    \footnotesize
    \setlength\tabcolsep{4pt}
    \begin{tabular}{l|ccccc|c|ccccc|c}
        \toprule
        \multirow{3}{*}{Method} & \multicolumn{6}{c|}{VOC20-C} & \multicolumn{6}{c}{Cityscapes-C} \\
        \cline{2-13}
         & \rotatebox[origin=c]{0}{B} & \rotatebox[origin=c]{0}{C} & \rotatebox[origin=c]{0}{ET} & \rotatebox[origin=c]{0}{P} & \rotatebox[origin=c]{0}{JC} & Mean$\uparrow$ & \rotatebox[origin=c]{0}{B} & \rotatebox[origin=c]{0}{C} & \rotatebox[origin=c]{0}{ET} & \rotatebox[origin=c]{0}{P} & \rotatebox[origin=c]{0}{JC} & Mean$\uparrow$ \\
        \midrule
        SOURCE      & 68.0 & 48.8 & 52.7 & 68.6 & 63.9 & 60.4       & 26.9 & 10.7 & 27.2 & 27.3 & 21.5 & 22.7 \\
        TENT        & 0.7 & 0.2 & 0.1 & 0.5 & 0.5 & 0.4             & 2.7 & 2.0 & 2.3 & 2.2 & 2.2 & 2.3 \\
        COTTA       & 60.2 & 36.1 & 40.9 & 61.9 & 57.4 & 51.3       & 24.1 & 4.6 & 25.6 & 24.8 & 20.8 & 20.0 \\
        SAR         & 58.0 & 27.4 & 39.9 & 57.3 & 52.2 & 46.9       & 27.7 & 9.6 & 27.1 & 27.4 & 21.3 & 22.6 \\
        WATT        & 69.9 & 50.3 & 53.9 & 70.7 & 66.2 & 62.2       & 18.2 & 1.1 & 20.9 & 19.7 & 14.7 & 14.9 \\
        SegTTO      & 1.1 & 1.1 & 1.0 & 1.0 & 1.1 & 1.1             & 2.7 & 2.1 & 2.6 & 2.7 & 2.5 & 2.5 \\
        MLMP        & 0.2 & 0.1 & 0.1 & 0.2 & 0.1 & 0.2             & 0.2 & 0.2 & 0.2 & 0.2 & 0.2 & 0.2 \\
        \midrule
        \rowcolor{gray!15}DAF-T     & 70.2 & 50.7 & 56.1 & 70.7 & 66.0 & 62.7       & 27.4 & 13.3 & 27.8 & 27.7 & 21.4 & 23.5 \\
        \rowcolor{gray!15}DAF-M     & 73.2 & 47.9 & 59.7 & 74.0 & 69.3 & 64.8       & 28.1 & 11.9 & 28.4 & 27.2 & 19.3 & 23.0 \\
        \bottomrule
    \end{tabular}
    \end{center}
\end{table*}
\begin{table*}[htbp]
    \caption{Domain generalization (forward transfer) using NACLIP\cite{Hajimiri2024PayAT} ViT-B/16. We perform CTTA on PASCAL VOC20-C \cite{Everingham2014ThePV} and LoveDA-C \cite{Wang2021LoveDAAR} domains except for the last 5 domains. We directly evaluate on them using the adapted weights. Entries report mIoU (\%). "-" indicates OOM.}
    \label{tab:ctta_domain_gen_vitb16}
    \begin{center}
    \footnotesize
    \setlength\tabcolsep{4pt}
    \begin{tabular}{l|ccccc|c|ccccc|c}
        \toprule
        \multirow{3}{*}{Method} & \multicolumn{6}{c|}{VOC20-C} & \multicolumn{6}{c}{LoveDA-C} \\
        \cline{2-13}
         & \rotatebox[origin=c]{0}{B} & \rotatebox[origin=c]{0}{C} & \rotatebox[origin=c]{0}{ET} & \rotatebox[origin=c]{0}{P} & \rotatebox[origin=c]{0}{JC} & Mean$\uparrow$ & \rotatebox[origin=c]{0}{B} & \rotatebox[origin=c]{0}{C} & \rotatebox[origin=c]{0}{ET} & \rotatebox[origin=c]{0}{P} & \rotatebox[origin=c]{0}{JC} & Mean$\uparrow$ \\
        \midrule
        SOURCE      & 73.3 & 60.3 & 50.1 & 75.5 & 69.1 & 65.6 & 20.3 & 7.0 & 19.2 & 13.8 & 6.7 & 13.4 \\
        TENT        & 0.3 & 1.3 & 0.1 & 0.2 & 0.2 & 0.4 & 1.2 & 1.0 & 1.0 & 1.0 & 1.0 & 1.1 \\
        WATT        & 75.1 & 59.5 & 47.1 & 76.1 & 69.6 & 65.5 & - & - & - & - & - & - \\
        CLIPArTT    & 1.6 & 1.8 & 0.7 & 1.0 & 1.0 & 1.2 & - & - & - & - & - & - \\
        MLMP        & 1.2 & 1.6 & 1.0 & 1.0 & 1.0 & 1.2 & 5.2 & 5.2 & 5.2 & 5.2 & 5.2 & 5.2 \\
        \midrule
        \rowcolor{gray!15}DAF-T     & 76.9 & 62.5 & 55.2 & 78.0 & 71.6 & 68.9 & 23.0 & 10.8 & 23.0 & 15.8 & 10.7 & 16.7 \\
        \rowcolor{gray!15}DAF-M     & 78.4 & 66.6 & 56.9 & 78.3 & 72.4 & 70.5 & 25.3 & 11.9 & 24.7 & 18.5 & 12.9 & 18.7 \\
        \bottomrule
    \end{tabular}
    \end{center}
\end{table*}

\subsection{Forward transfer.}
\cref{tab:ctta_domain_gen_vitb32} and \cref{tab:ctta_domain_gen_vitb16} evaluate forward transfer by adapting on all but the last five domains and directly evaluating on the held-out domains. On VOC20-C with ViT-B/32, DAF-M reaches 64.8 mean mIoU versus the source of 60.4, while TENT and MLMP collapse to 0.4 and 0.2 before reaching the held-out domains. On Cityscapes-C, DAF-T and DAF-M exceed the source of 22.7. The same pattern holds with ViT-B/16: DAF-M reaches 70.5 on VOC20-C (source 65.6) and 18.7 on LoveDA-C (source 13.4). Methods that collapse during adaptation have nothing to transfer, confirming the error accumulation concern raised in the main paper. DAF's CMAC loss prevents drift accumulation so that adapted weights remain useful for unseen domains, turning continual adaptation into a net positive for forward transfer rather than a source of degradation.

\begin{table}[htbp]
    \caption{Learning rate sweep on LoveDA\cite{Wang2021LoveDAAR} with NACLIP\cite{Hajimiri2024PayAT} ViT-B/32. Entries report mIoU (\%).}
    \label{tab:loveda_lr_sweep}
    \begin{center}
    \footnotesize
    \setlength\tabcolsep{4pt}
    \begin{tabular}{l|cccc|c}
        \toprule
        \multirow{2}{*}{METHOD} & \multicolumn{4}{c|}{Learning Rate} & \multirow{2}{*}{Mean$\uparrow$} \\
        \cline{2-5}
         & 1e-2 & 1e-3 & 1e-4 & 1e-5 & \\
        \midrule
        SOURCE      & 22.7 & 22.7 & 22.7 & 22.7 & 22.7 \\
        TENT        & 4.5 & 20.1 & -- & 22.8 & 15.8 \\
        COTTA       & 21.0 & 22.1 & 22.6 & 22.7 & 22.1 \\
        SAR         & 16.6 & 20.2 & 23.1 & 22.8 & 20.7 \\
        SegTTO      & 5.7 & 7.8 & 18.2 & 28.4 & 15.0 \\
        MLMP        & 6.6 & 15.2 & 27.9 & 27.7 & 19.4 \\
        \midrule
        \rowcolor{gray!15}DAF-T & 25.3 & 24.1 & 23.1 & 22.8 & 23.8 \\
        \rowcolor{gray!15}DAF-M & 31.9 & 31.8 & 28.4 & 27.7 & 30.0 \\
        \bottomrule
    \end{tabular}
    \end{center}
\end{table}

\subsection{Learning rate robustness on LoveDA.}
\cref{tab:loveda_lr_sweep} repeats the learning rate sweep on LoveDA with ViT-B/32. The pattern mirrors the VOC20-C findings: TENT and MLMP collapse at higher learning rates, while DAF-T and DAF-M stay above source across all learning rates. At $1{\times}10^{-2}$, DAF-M reaches 31.9 while TENT drops to 4.5 and MLMP to 6.6, confirming the same robustness gap observed on VOC20-C.

\begin{table}[htbp]
    \caption{OVSS variant comparison on PASCAL VOC20-C \cite{Everingham2014ThePV} with backbone ViT-B/32. Entries report mIoU and mDice (\%).}
    \label{tab:voc20_ovss}
    \begin{center}
    \footnotesize
    \setlength\tabcolsep{3pt}
    \begin{tabular}{l|cc|cc}
        \toprule
        \multirow{2}{*}{Method} & \multicolumn{2}{c|}{SCLIP\cite{Wang2023SCLIPRS}} & \multicolumn{2}{c}{NACLIP\cite{Hajimiri2024PayAT}} \\
        \cmidrule(lr){2-3} \cmidrule(lr){4-5}
         & mIoU & mDice & mIoU & mDice \\
        \midrule
        SOURCE      & 51.0 & 65.8 & 54.4 & 68.9 \\
        TENT        & 2.2 & 3.4 & 3.3 & 5.0 \\
        WATT        & 47.5 & 62.5 & 45.7 & 60.7 \\
        CLIPArTT    & 2.7 & 4.5 & 2.1 & 3.7 \\
        MLMP        & 1.5 & 2.6 & 13.5 & 18.2 \\
        \midrule
        \rowcolor{gray!15}DAF-T & 55.1 & 69.5 & 57.4 & 71.6 \\
        \rowcolor{gray!15}DAF-M & 20.9 & 31.5 & 62.8 & 76.0 \\
        \bottomrule
    \end{tabular}
    \end{center}
\end{table}
\begin{table}[htbp]
    \caption{OVSS variant comparison on LoveDA-C\cite{Wang2021LoveDAAR} with backbone ViT-B/32. Entries report mIoU and mDice (\%).}
    \label{tab:loveda_ovss}
    \begin{center}
    \footnotesize
    \setlength\tabcolsep{3pt}
    \begin{tabular}{l|cc|cc}
        \toprule
        \multirow{2}{*}{Method} & \multicolumn{2}{c|}{SCLIP\cite{Wang2023SCLIPRS}} & \multicolumn{2}{c}{NACLIP\cite{Hajimiri2024PayAT}} \\
        \cmidrule(lr){2-3} \cmidrule(lr){4-5}
         & mIoU & mDice & mIoU & mDice \\
        \midrule
        SOURCE      & 10.8 & 18.4 & 9.2 & 15.9 \\
        TENT        & 1.6 & 3.0 & 1.2 & 2.2 \\
        MLMP        & 1.3 & 2.5 & 5.2 & 7.6 \\
        \midrule
        \rowcolor{gray!15}DAF-T & 12.9 & 21.7 & 12.3 & 20.6 \\
        \rowcolor{gray!15}DAF-M & 5.3 & 8.2 & 14.9 & 24.5 \\
        \bottomrule
    \end{tabular}
    \end{center}
\end{table}

\subsection{OVSS variant analysis.}
\cref{tab:voc20_ovss} and \cref{tab:loveda_ovss} compare DAF across SCLIP and NACLIP OVSS backbones on VOC20-C and LoveDA-C. On NACLIP, DAF-M reaches 62.8 on VOC20-C and 14.9 on LoveDA-C, both exceeding source by significant margins. DAF-T is more consistent across backbones, exceeding source on SCLIP for both datasets (55.1 vs 51.0 on VOC20-C, 12.9 vs 10.8 on LoveDA-C). DAF-M excels on NACLIP but can degrade on SCLIP, where the weaker initial vision-language alignment provides a less stable anchor for CMAC. The collapse pattern persists across both OVSS variants, confirming that class collapse is intrinsic to CTTA rather than specific to one backbone. These results suggest that DAF's stabilization is most effective when the source model has stronger cross-modal alignment to serve as an anchor, consistent with the CMAC design in the main paper.

\begin{table}[htbp]
    \caption{Ablation study on DAF components and adaptation efficiency on LoveDA \cite{Wang2021LoveDAAR} using NACLIP\cite{Hajimiri2024PayAT} ViT-B/32. Each input image has a size of $1024{\times}1024$ and is divided into $224{\times}224$ samples as input to the CLIP model. Each domain has an average of $S{=}107008$ samples with a batch size of 8 input images, evaluated on an AMD Instinct MI200 GPU. \#Fward and \#Bward denote the total number of forward and backward passes per domain, expressed in terms of $S$. Time (s) is the per-domain adaptation time in seconds.}
    \label{tab:love_efficiency}
    \begin{center}
    \footnotesize
    \setlength\tabcolsep{3pt}
    \begin{tabular}{l|ccc|ccc|c}
        \toprule
         Method & $\mathcal{L}_{\text{div}}$ & $\mathcal{L}_{\text{cmac}}$ & $\rho_{\text{safs}}$ & \#Fward$\downarrow$ & \#Bward$\downarrow$ & Time(s)$\downarrow$ & mIoU$\uparrow$ \\
        \midrule
        SOURCE      & $\times$ & $\times$ & $\times$ & $1.0S$ & $0$       & 57.5  & 22.7 \\
        TENT        & $\times$ & $\times$ & $\times$ & $1.0S$ & $1.0S$    & 67.0  & 20.5 \\
        MLMP        & $\times$ & $\times$ & $\times$ & $1.0S$ & $1.0S$    & 79.0  & 18.2 \\
        \midrule
        \rowcolor{gray!15}DAF-T      & $\checkmark$ & $\checkmark$ & $\checkmark$ & $2.0S$ & $0.7S$     & 71.5  & 24.1 \\
        \rowcolor{gray!15}DAF-M      & $\checkmark$ & $\times$ & $\times$ & $2.0S$ & $1.0S$             & 83.0  & 30.3 \\
        \rowcolor{gray!15}DAF-M      & $\times$ & $\checkmark$ & $\times$ & $2.0S$ & $1.0S$             & 92.0  & 27.4 \\
        \rowcolor{gray!15}DAF-M      & $\checkmark$ & $\checkmark$ & $\times$ & $2.0S$ & $1.0S$         & 99.5  & 31.1 \\
        \rowcolor{gray!15}DAF-M      & $\checkmark$ & $\checkmark$ & $\checkmark$ & $2.0S$ & $0.7S$     & 81.5  & 31.8 \\
        \bottomrule
    \end{tabular}
    \end{center}
\end{table}

\subsection{Cross-dataset ablation.}
\cref{tab:love_efficiency} repeats the component ablation on LoveDA with ViT-B/32. The findings mirror the VOC20-C results: MDIV alone reaches 30.3 and CMAC alone 27.4, both improving over MLMP's 18.2 but neither matching the combined 31.1 without SAFS. This confirms they address complementary failure modes, with MDIV resisting class collapse and CMAC constraining feature drift, as designed in \cref{sec:methodology}. Comparing MDIV+CMAC with and without SAFS shows that filtering reduces backward passes by approximately 30\% (from $1.0S$ to $0.7S$ per domain) and lowers DAF-M adaptation time from 99.5 s to 81.5 s, at the cost of less than one mIoU point (31.8 vs 31.1), confirming its role as an efficiency tool rather than a stability mechanism. This cross-dataset consistency confirms that the complementary design of MDIV and CMAC in the main paper generalizes across datasets, though the degree of complementarity varies: on VOC20-C the combination gains roughly 19 points over the better single component, while on LoveDA MDIV alone captures most of the stabilization benefit and CMAC adds only 0.8 points, possibly because the 7-class LoveDA vocabulary is less prone to the cross-modal drift that CMAC targets.

\subsection{Consistency loss variants.}
\label{sec:consistency_variants}
To isolate the contribution of text-anchored consistency, we compare CMAC against two feature-only consistency losses that penalize drift between adapted and source patch tokens without referencing text prototypes. Both operate on the L2-normalized patch-token features $\mathbf{f}^{\text{adapt}}_{q,i}$ and $\mathbf{f}^{\text{src}}_{q,i}$ defined in \cref{sec:preliminaries_full}, excluding the CLS token. The cosine variant penalizes angular drift on the unit hypersphere:
\begin{equation}
\label{eq:feat_cosine}
\mathcal{L}_{\text{cosine}} = \frac{1}{QN} \sum_{q=1}^{Q} \sum_{i=1}^{N} \left(1 - \mathbf{f}^{\text{adapt}}_{q,i} \cdot \mathbf{f}^{\text{src}}_{q,i}\right),
\end{equation}
where the dot product is the cosine similarity between L2-normalized vectors. The L2 variant penalizes the squared Euclidean distance:
\begin{equation}
\label{eq:feat_l2}
\mathcal{L}_{\text{L2}} = \frac{1}{QN} \sum_{q=1}^{Q} \sum_{i=1}^{N} \|\mathbf{f}^{\text{adapt}}_{q,i} - \mathbf{f}^{\text{src}}_{q,i}\|_2^2.
\end{equation}
Because CLIP features are L2-normalized, the squared L2 distance and cosine distance are proportional on the unit hypersphere: $\|\mathbf{f}^{\text{adapt}}_{q,i} - \mathbf{f}^{\text{src}}_{q,i}\|_2^2 = 2(1 - \cos\theta_{q,i})$, where $\theta_{q,i}$ is the angle between the two vectors. Both losses therefore constrain the same angular drift, though $\mathcal{L}_{\text{L2}}$ provides a smoother gradient near zero drift. In practice, each loss is weighted by $\lambda_{\text{feat\_cons}} = 0.5$ and added to the base adaptation loss in place of CMAC, activated via the flags \texttt{-{}-loss\_feat\_cons True}, \texttt{-{}-lamb\_feat\_cons 0.5}, and \texttt{-{}-feat\_cons\_type \{cosine,l2\}}. The source model $g_{\theta_s}$ is the same frozen deep copy used by CMAC. The key distinction from CMAC is that these losses penalize all directional drift equally, while CMAC uses text prototypes to distinguish harmful drift (away from the assigned class, toward unassigned classes) from benign reorientation that preserves the discriminative structure of the vision-language alignment. This text-anchored directionality is what allows CMAC to preserve segmentation accuracy under continual adaptation, as demonstrated by the 11-point mIoU gap between CMAC and $\mathcal{L}_{\text{cosine}}$ on VOC20-C reported in \cref{tab:ctta_consistency_type_vitb32}.

\begin{figure*}[htbp]
    \centering
    \begin{subfigure}{0.24\textwidth}
        \centering
        \includegraphics[width=\textwidth]{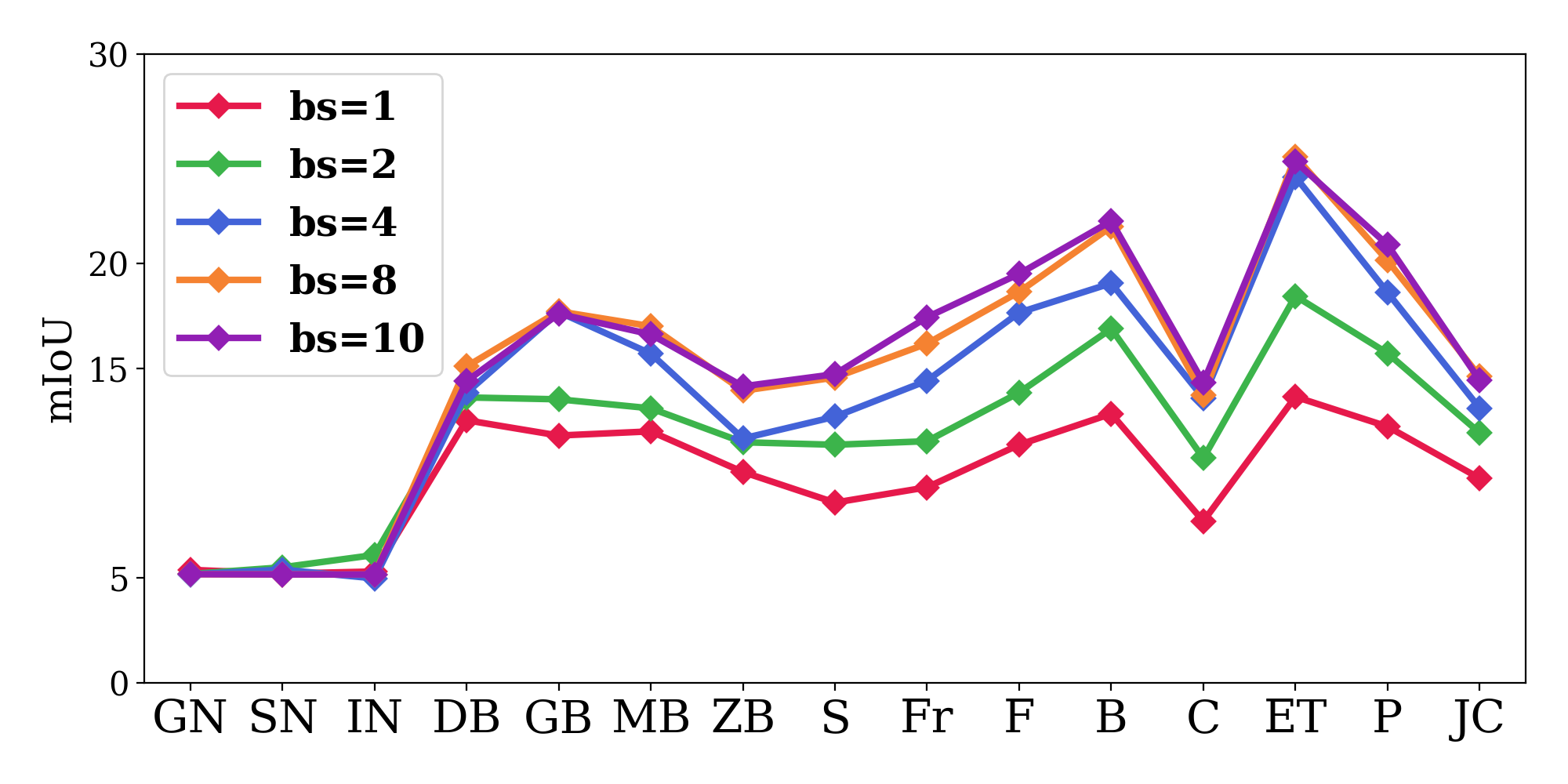}
        \caption{Batch Size}
    \end{subfigure}
    \hfill
    \begin{subfigure}{0.24\textwidth}
        \centering
        \includegraphics[width=\textwidth]{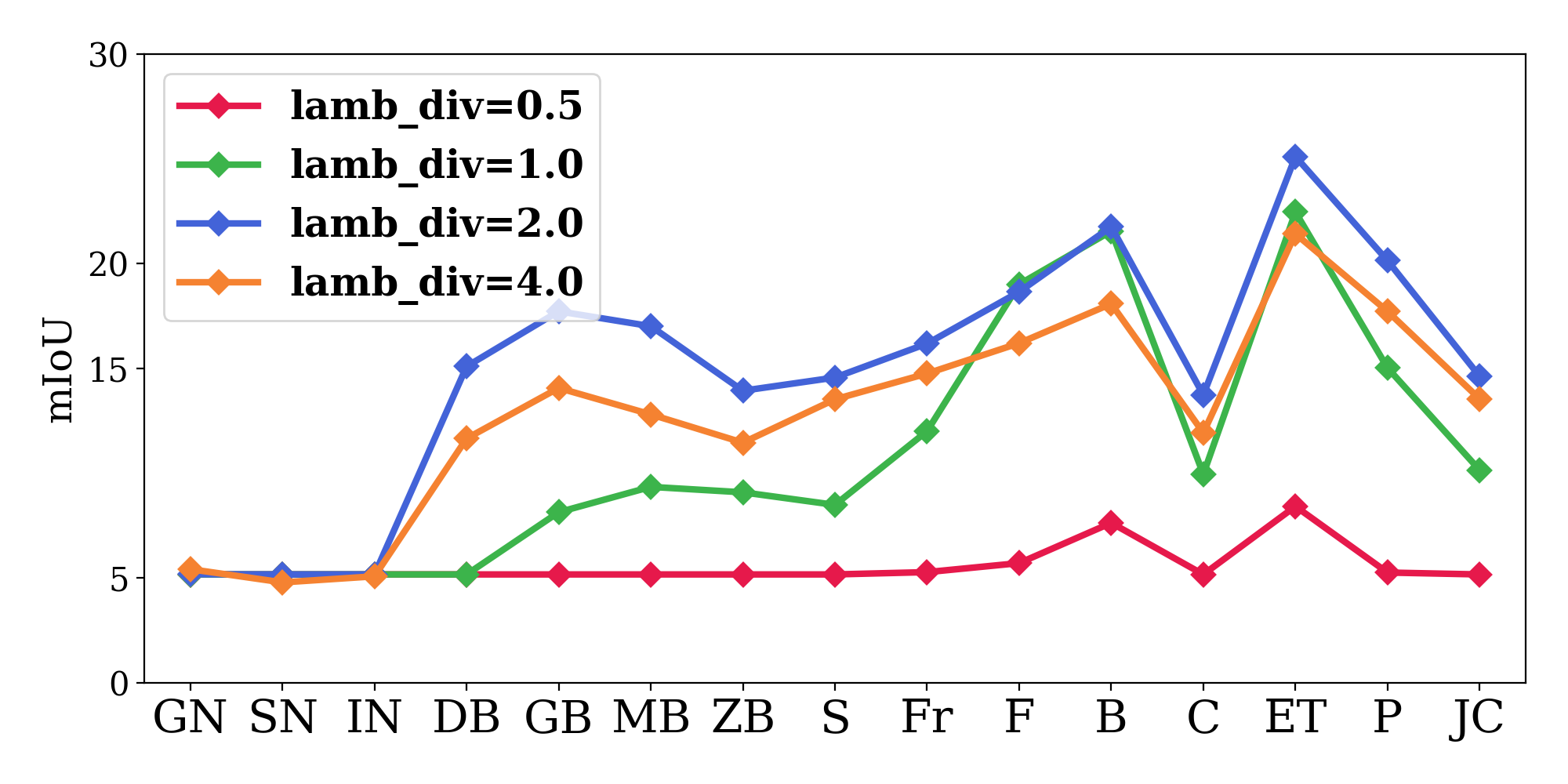}
        \caption{$\lambda_{\text{div}}$}
    \end{subfigure}
    \hfill
    \begin{subfigure}{0.24\textwidth}
        \centering
        \includegraphics[width=\textwidth]{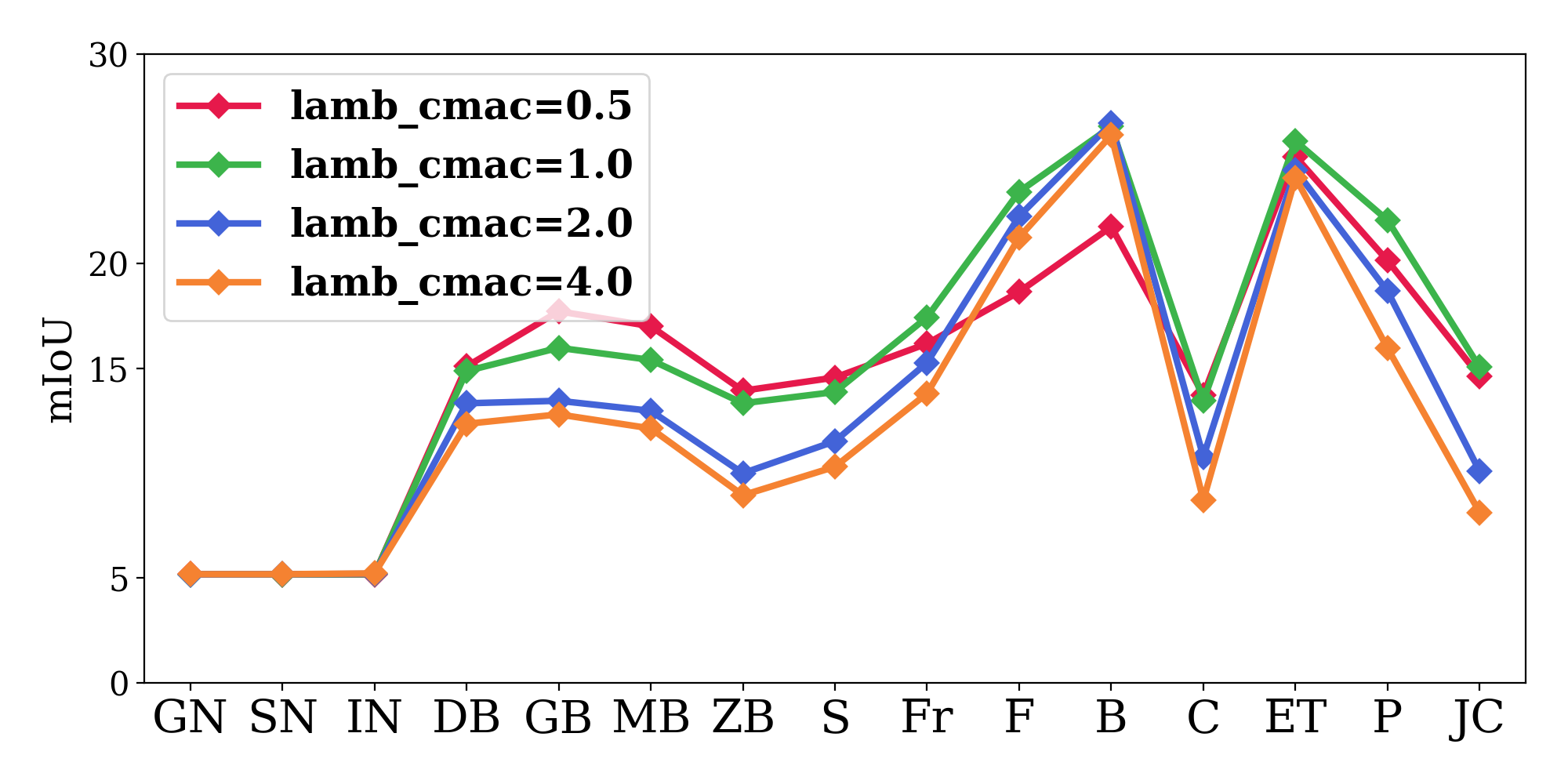}
        \caption{$\lambda_{\text{cmac}}$}
    \end{subfigure}
    \hfill
    \begin{subfigure}{0.24\textwidth}
        \centering
        \includegraphics[width=\textwidth]{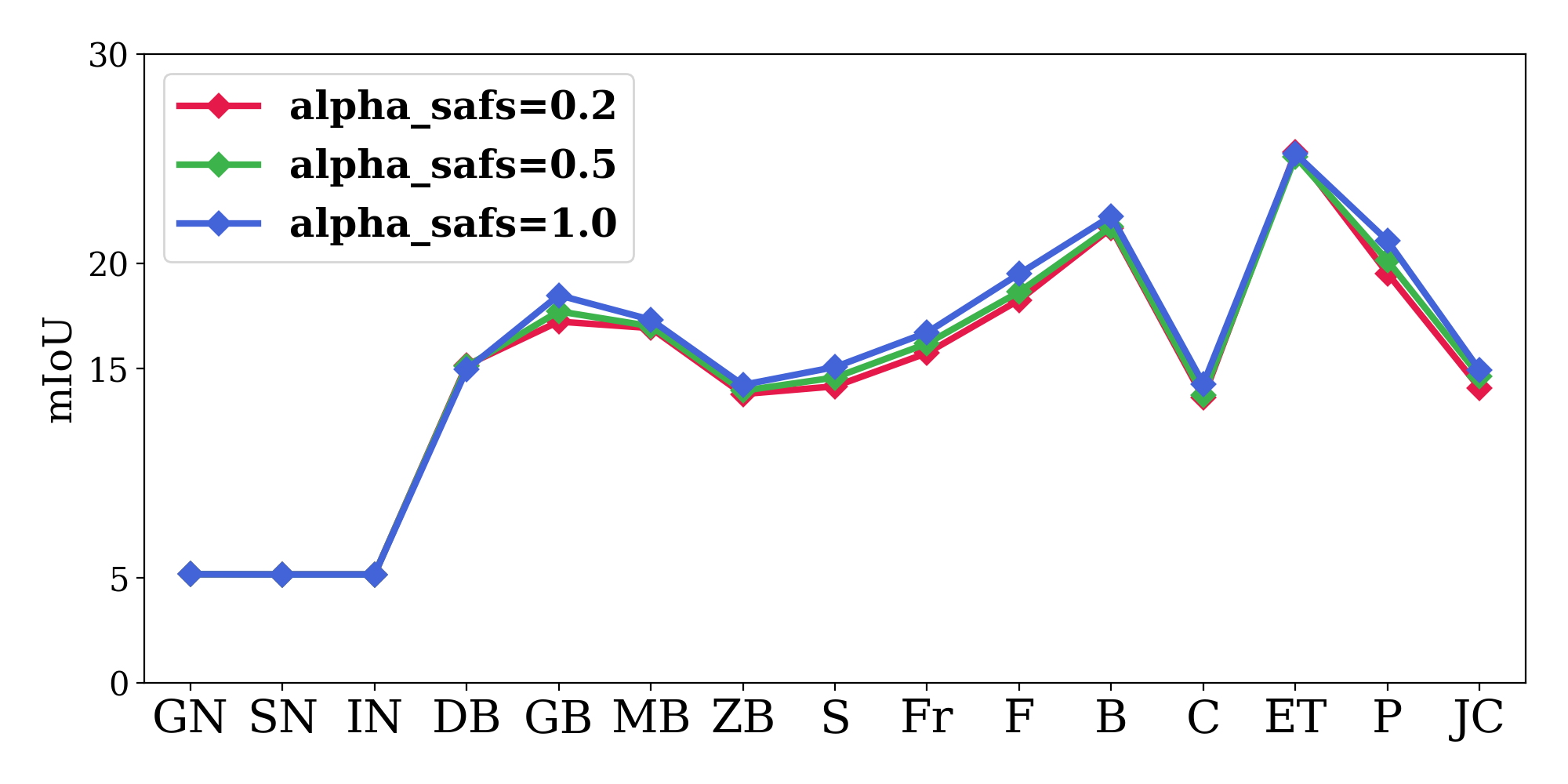}
        \caption{$\alpha_{\text{safs}}$}
    \end{subfigure}
    \caption{Ablation results of DAF-M with NACLIP~\cite{Hajimiri2024PayAT} ViT-B/32 under CTTA in LoveDA-C\cite{Wang2021LoveDAAR}. Entries show mIoU (\%). Each panel shows the effect of a single hyperparameter: (a) batch size, (b) $\lambda_{\text{div}}$, (c) $\lambda_{\text{cmac}}$, and (d) $\alpha_{\text{safs}}$.}
    \label{fig:loveda_hyperparams}
\end{figure*}

\subsection{LoveDA hyperparameter analysis.}
\cref{fig:loveda_hyperparams} sweeps the same four hyperparameters of DAF-M on LoveDA-C. As on VOC20-C, all four exhibit broad stable plateaus, confirming that DAF does not require per-dataset tuning.

\section{Diagnostic Analysis}
\label{sec:diagnostic_analysis}

\begin{figure*}[htbp]
    \centering
    \begin{subfigure}{0.24\textwidth}
        \centering
        \includegraphics[width=\textwidth]{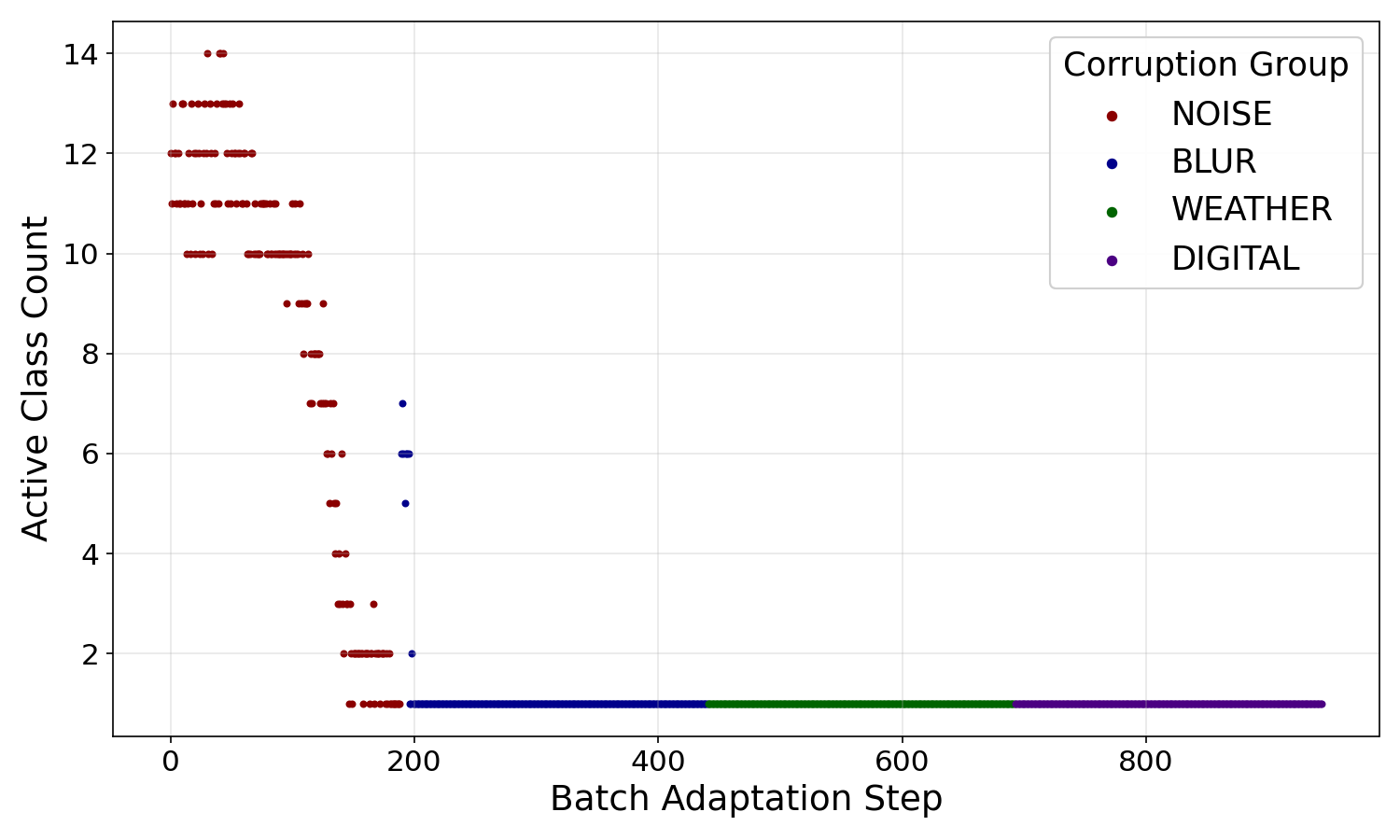}
        \caption{Class Collapse}
    \end{subfigure}
    \hfill
    \begin{subfigure}{0.24\textwidth}
        \centering
        \includegraphics[width=\textwidth]{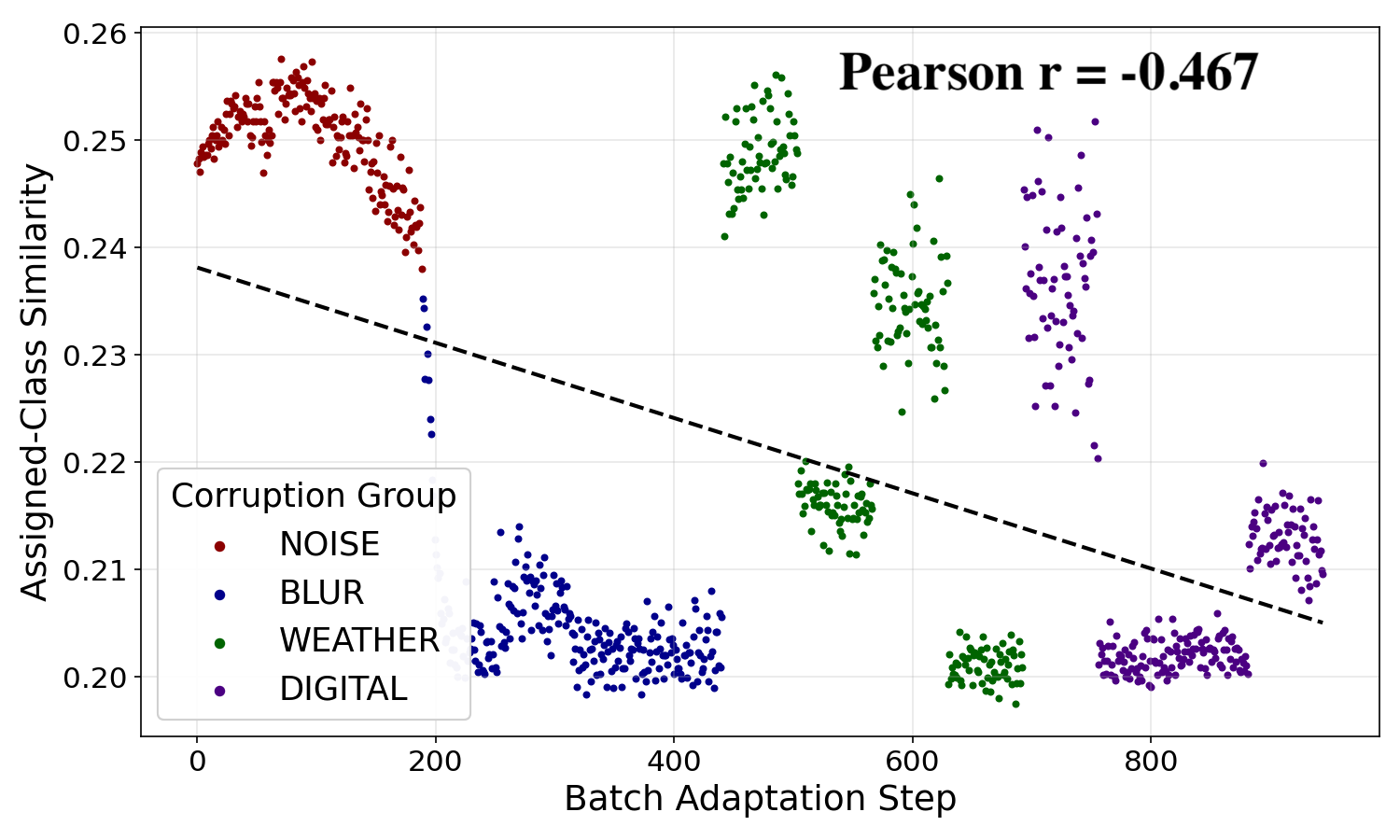}
        \caption{Drift Away}
    \end{subfigure}
    \hfill
    \begin{subfigure}{0.24\textwidth}
        \centering
        \includegraphics[width=\textwidth]{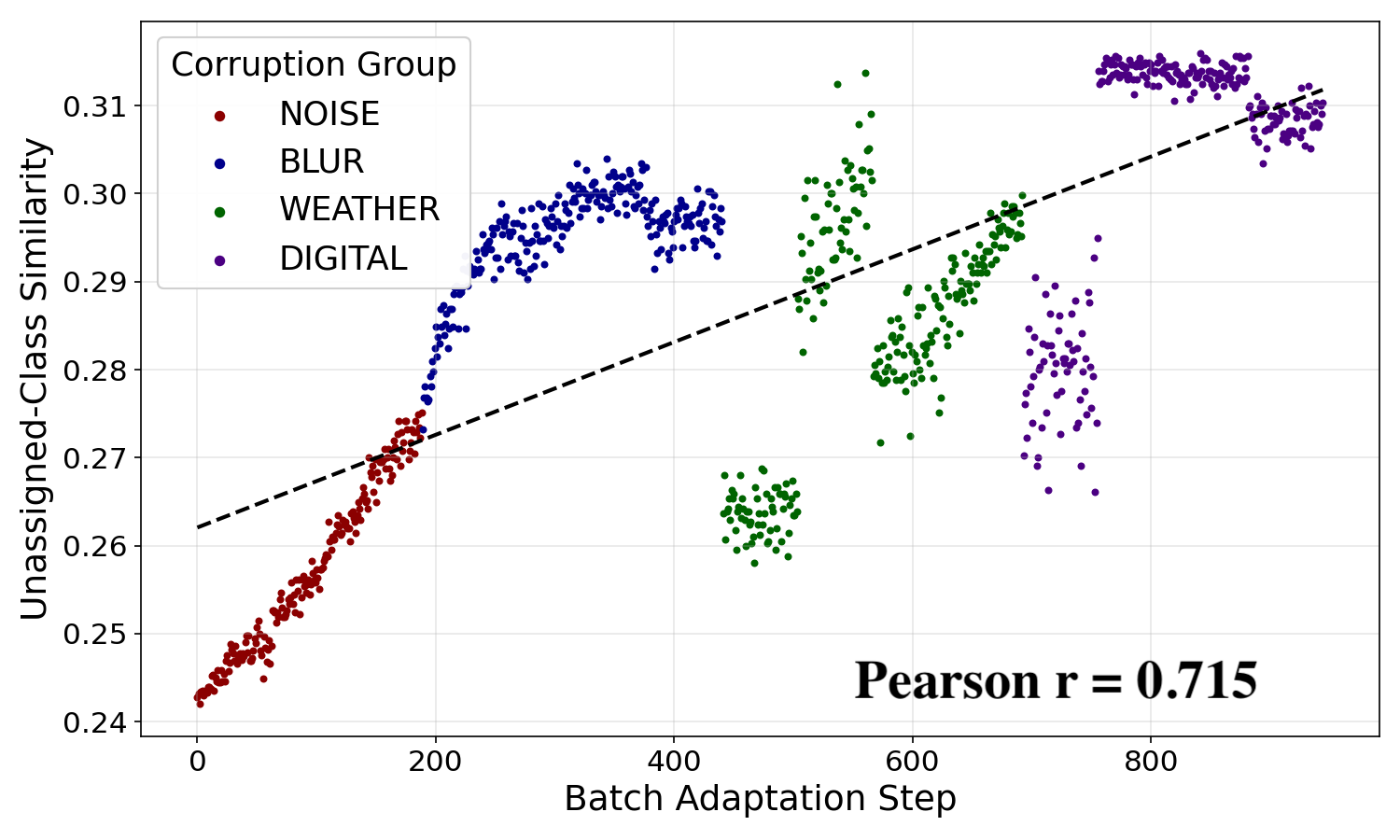}
        \caption{Drift Toward}
    \end{subfigure}
    \hfill
    \begin{subfigure}{0.192\textwidth}
        \centering
        \includegraphics[width=\textwidth]{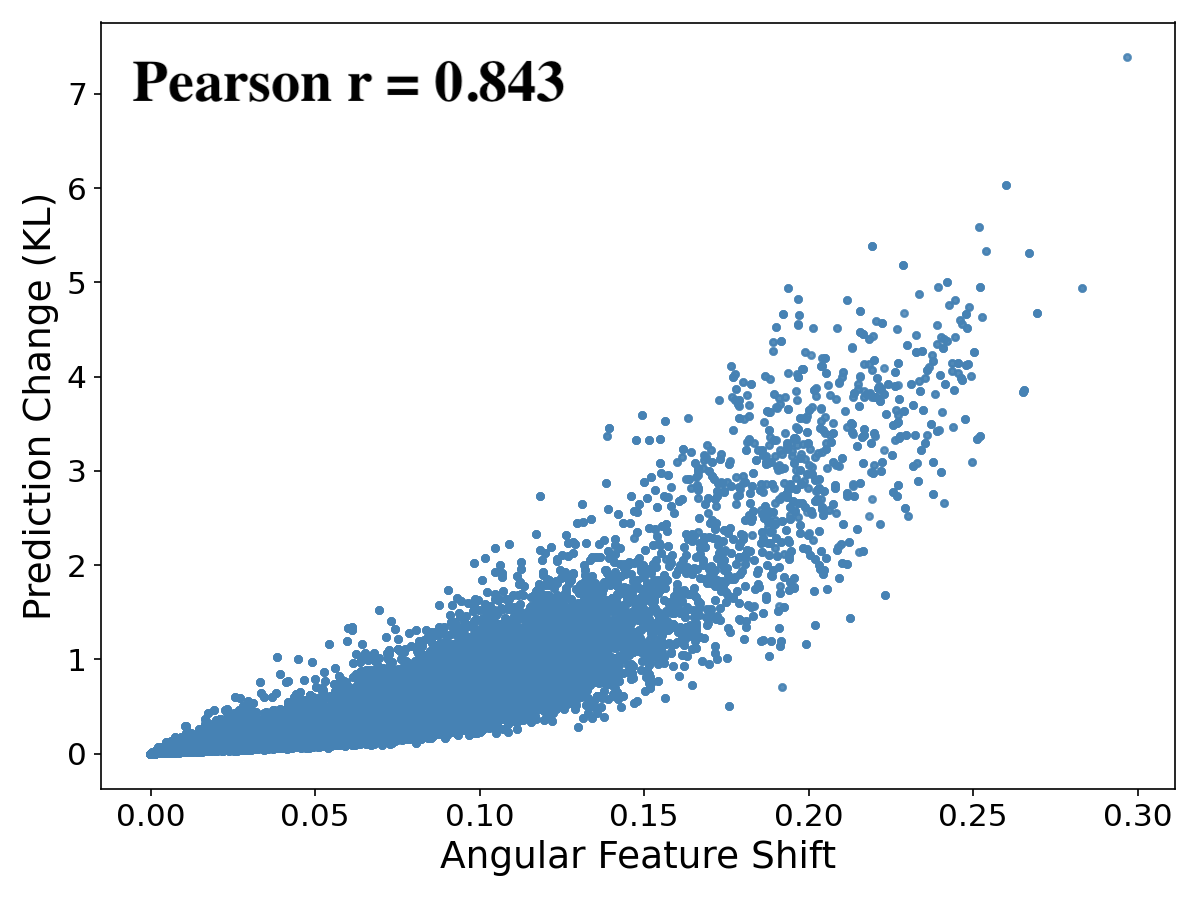}
        \caption{Shift vs Prediction}
    \end{subfigure}
    \caption{Diagnostic analysis of MLMP~\cite{Noori2025TestTimeAO} under CTTA in OVSS with NACLIP~\cite{Hajimiri2024PayAT} ViT-B/32 on Cityscapes-C~\cite{Cordts2016TheCD}. (a) Entropy minimization progressively shrinks the active class vocabulary. (b) Adapted features drift away from their source assigned text prototypes ($r = -0.467$). (c) Features simultaneously drift toward unassigned prototypes ($r = +0.715$). (d) Per sample feature shift correlates with prediction change ($r = +0.843$), indicating that source-relative feature change can identify samples whose representations have changed little. These observations motivate our diversity and anchor consistency components for stability, and our filtering component for adaptation efficiency.}
    \label{fig:city_diagnostics_pre}
\end{figure*}

\begin{figure*}[htbp]
    \centering
    \begin{subfigure}{0.30\textwidth}
        \centering
        \includegraphics[width=\textwidth]{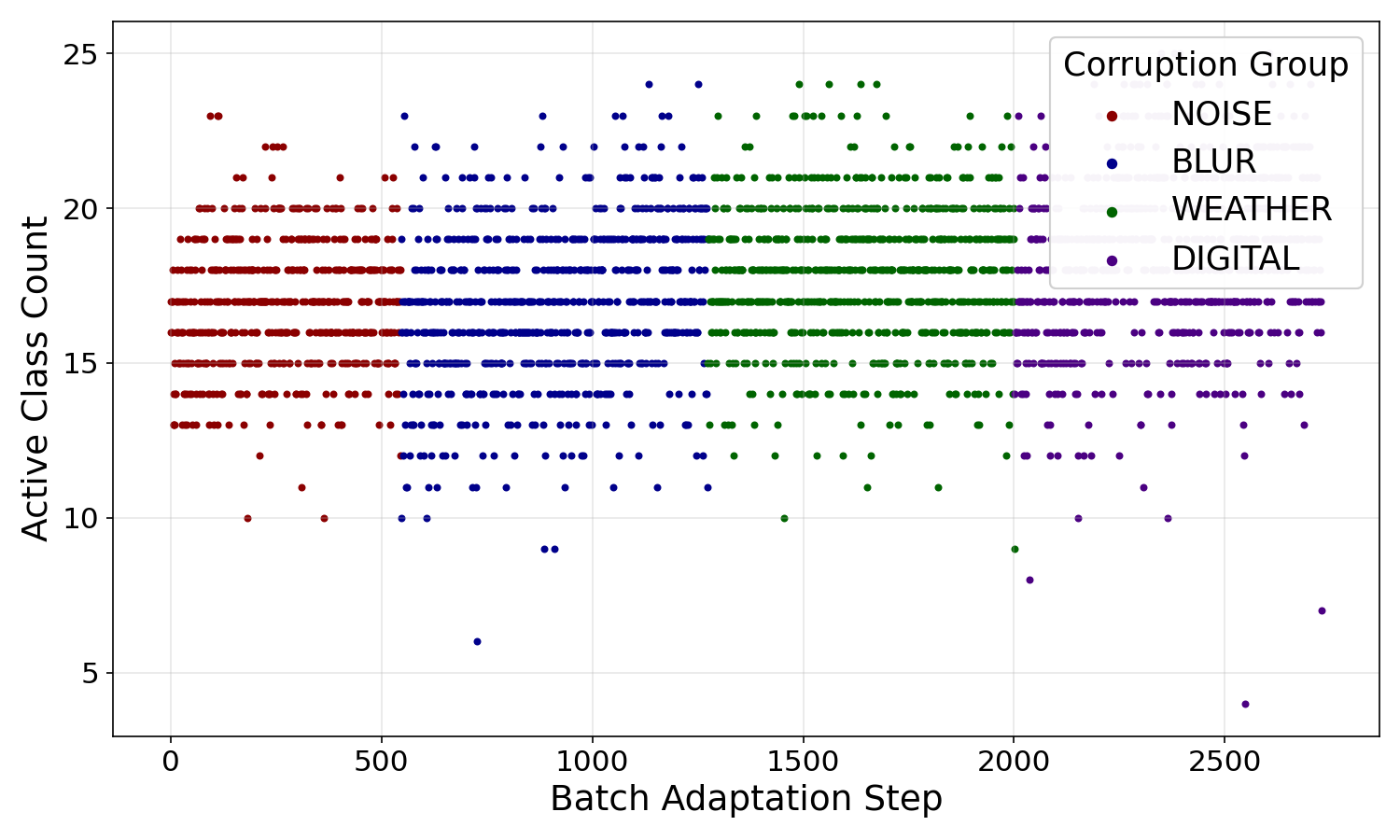}
        \caption{Active Classes}
    \end{subfigure}
    \hfill
    \begin{subfigure}{0.30\textwidth}
        \centering
        \includegraphics[width=\textwidth]{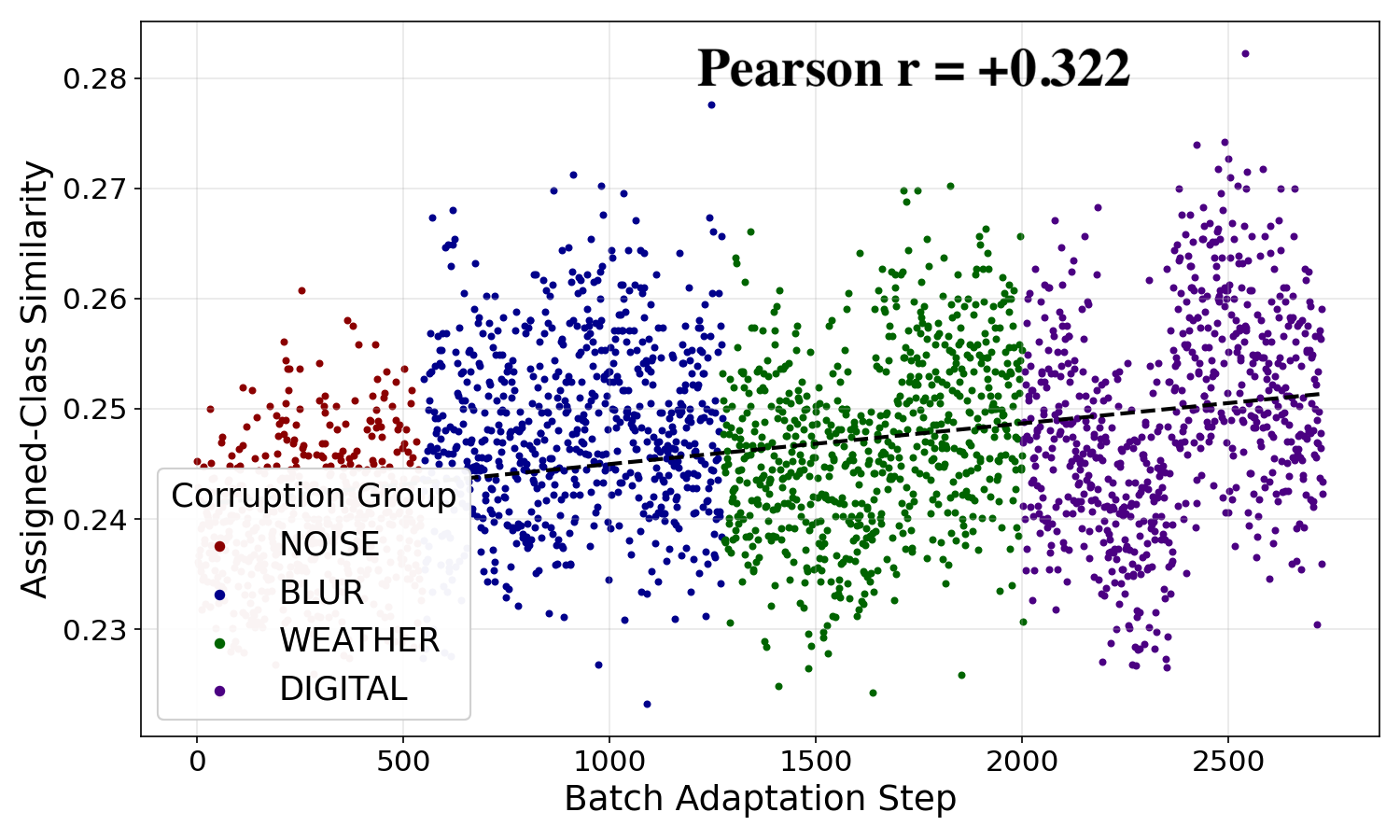}
        \caption{Assigned Similarity}
    \end{subfigure}
    \hfill
    \begin{subfigure}{0.30\textwidth}
        \centering
        \includegraphics[width=\textwidth]{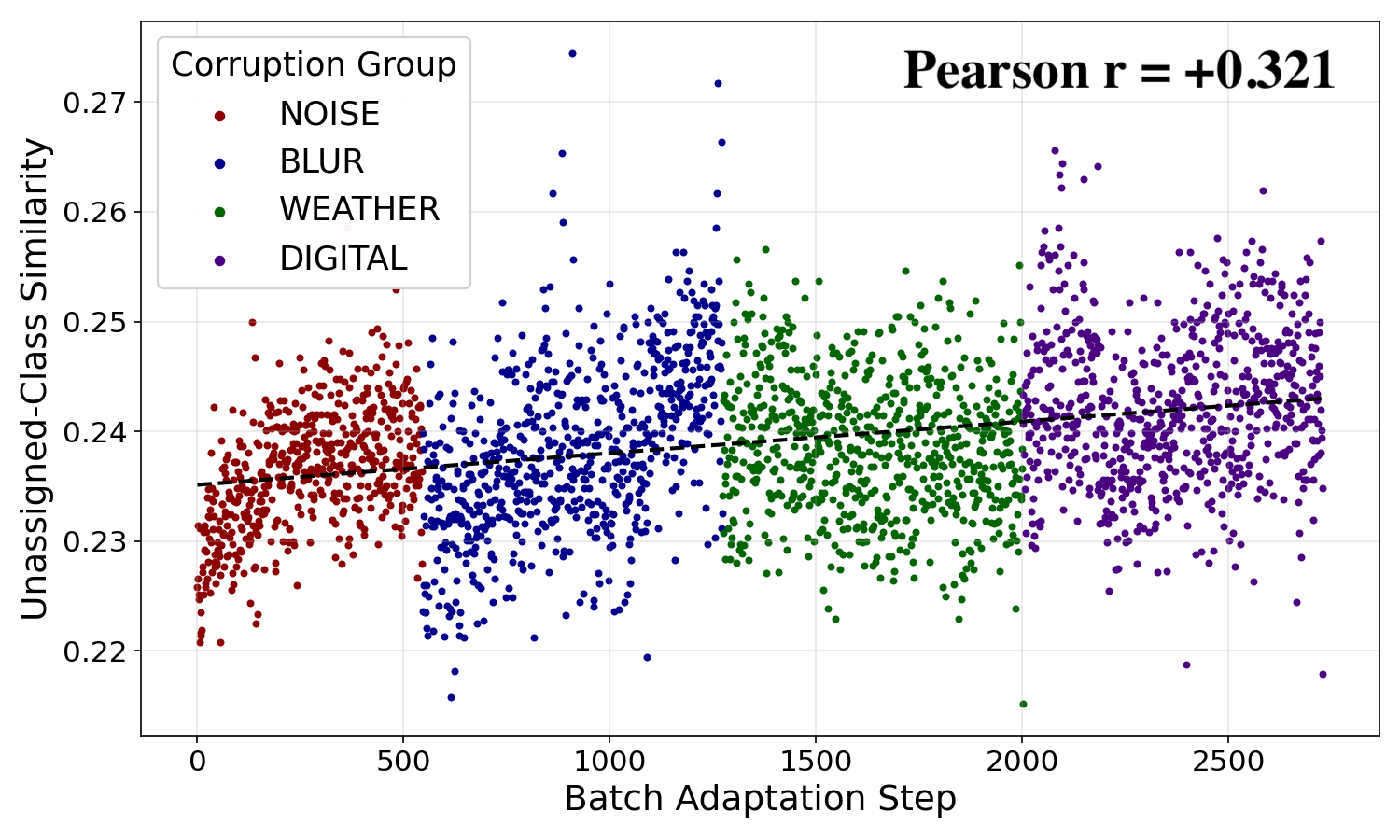}
        \caption{Unassigned Similarity}
    \end{subfigure}
    \caption{Effect of the marginal diversity loss and cross-modal anchor consistency from DAF on MLMP~\cite{Noori2025TestTimeAO} under CTTA for OVSS with NACLIP~\cite{Hajimiri2024PayAT} ViT-B/32 on PASCAL VOC20-C~\cite{Everingham2014ThePV}. (a) The active class vocabulary remains stable, counteracting collapse. (b) Adapted features has near neutral alignment with source assigned text prototypes ($r = +0.322$). (c) Similarity to unassigned prototypes stays near neutral ($r = +0.321$), preventing drift toward wrong classes.}
    \label{fig:voc20_diagnostics_post}
\end{figure*}

\subsection{Diagnostic methodology.}
All diagnostics are computed at each adaptation step using the adapted model $g_\theta$ and the frozen source model $g_{\theta_s}$, both evaluated on the same input batch. We use the notation defined in \cref{sec:preliminaries_full}.

\textbf{Class collapse.} From the patch-level softmax $p_r(c \mid \mathbf{x}_{q,i})$, we take the argmax over classes for each patch token to obtain a per-patch prediction. We count the fraction of patch tokens assigned to each class across the entire sample batch and define active classes as those receiving more than 1\% of predictions. The active class vocabulary is the number of active classes, tracked over adaptation steps.

\textbf{Cross-modal drift.} Using the L2-normalized image patch features $\mathbf{f}^{\text{adapt}}_{q,i}$ and $\mathbf{f}^{\text{src}}_{q,i}$ (excluding the CLS token) and the prompt-averaged text prototype $\mathbf{T}_c$, we compute cosine similarities
\begin{equation}
s^{\text{adapt}}_{q,i,c} = \mathrm{cos}(\mathbf{f}^{\text{adapt}}_{q,i}, \mathbf{T}_c), \quad
s^{\text{src}}_{q,i,c} = \mathrm{cos}(\mathbf{f}^{\text{src}}_{q,i}, \mathbf{T}_c).
\end{equation}
Each source patch is assigned to its nearest prototype $a_{q,i} = \arg\max_c s^{\text{src}}_{q,i,c}$. Drift-away is the change in similarity to the assigned prototype, measured as $s^{\text{adapt}}_{q,i,a_{q,i}} - s^{\text{src}}_{q,i,a_{q,i}}$, averaged over all patch tokens. Drift-toward is the change in maximum similarity to any unassigned prototype, measured as $\max_{c \neq a_{q,i}} s^{\text{adapt}}_{q,i,c} - \max_{c \neq a_{q,i}} s^{\text{src}}_{q,i,c}$, also averaged over all patch tokens. Both quantities are tracked over adaptation steps.

\textbf{Feature shift and prediction change.} Per-sample feature shift is
\begin{equation}
\Delta_q = 1 - \frac{1}{N} \sum_{i=1}^{N} \mathrm{cos}(\mathbf{f}^{\text{adapt}}_{q,i}, \mathbf{f}^{\text{src}}_{q,i}),
\end{equation}
the average cosine distance between adapted and source patch features within each sample. Prediction change is the symmetric KL divergence between the adapted and source patch-level softmax distributions, averaged over patch tokens within each sample. The correlation between $\Delta_q$ and prediction change across samples motivates the SAFS filtering criterion.

\subsection{Motivating diagnostics.}
\cref{fig:city_diagnostics_pre} and \cref{fig:voc20_diagnostics_post} provide the empirical foundation for DAF's design. Under MLMP without stabilization on Cityscapes-C, entropy minimization progressively shrinks the active class vocabulary, confirming the class collapse mechanism. Adapted features drift away from their source assigned text prototypes ($r = -0.467$) and toward unassigned ones ($r = +0.715$), directly visualizing the alignment erosion diagnosed in the main paper. Feature shift also correlates with prediction change ($r = +0.843$), indicating that samples with small source-relative shift need no gradient update, which motivates the SAFS filtering component. With DAF applied on VOC20-C, the active vocabulary remains stable, assigned prototype alignment improves to near-neutral ($r = +0.332$), and unassigned similarity stays near-neutral ($r = +0.321$). This before-and-after comparison validates that the MDIV counteracts collapse and CMAC constrains drift, as designed in the main paper. The shift signal underlying SAFS is a reliable predictor of adaptation necessity.

\subsection{MDIV effect on VOC21-C.}
\begin{figure}[htbp]
    \centering
    \begin{subfigure}{0.48\columnwidth}
        \centering
        \includegraphics[width=\columnwidth]{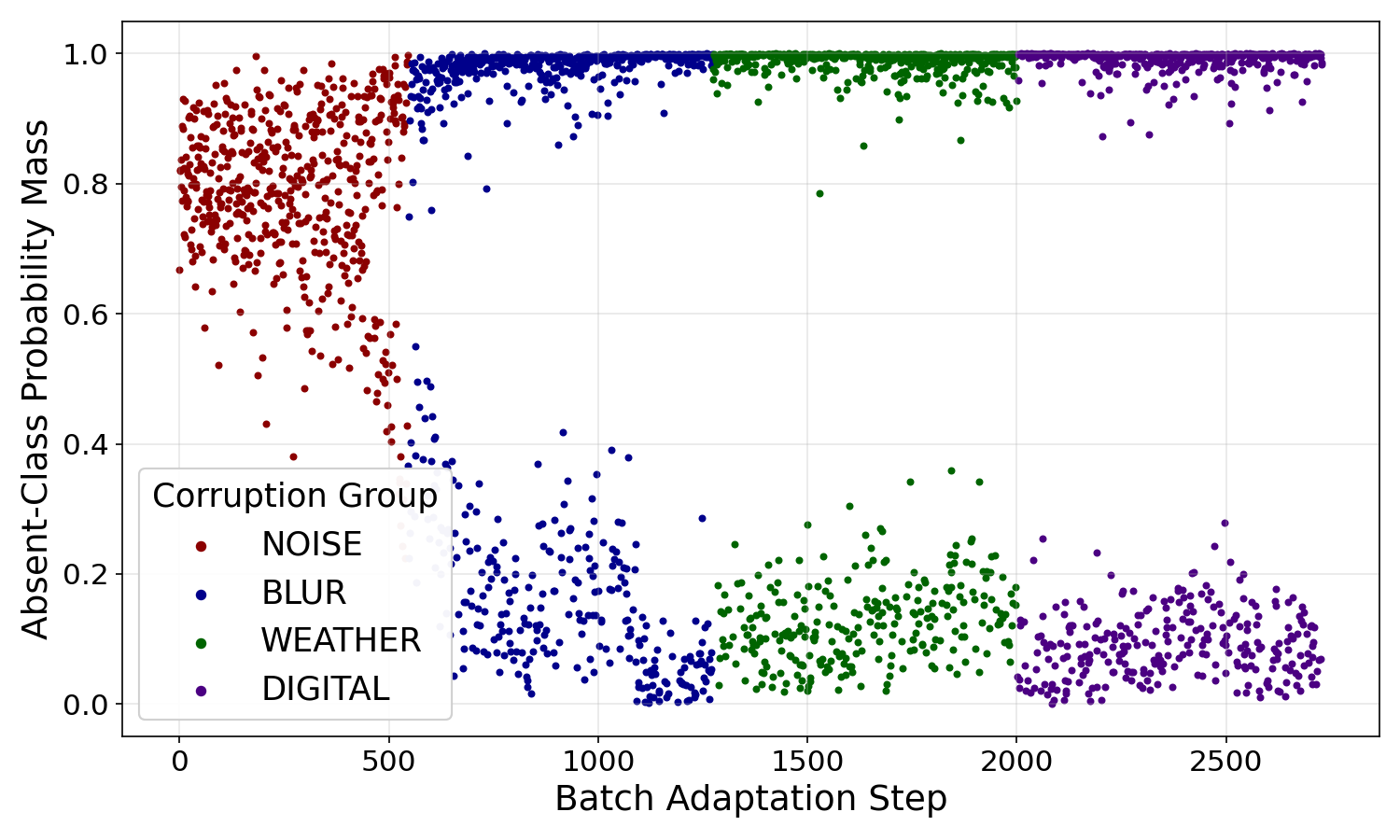}
        \caption{$\mathcal{L}_{\text{div}} = \text{False}$}
    \end{subfigure}
    \hfill
    \begin{subfigure}{0.48\columnwidth}
        \centering
        \includegraphics[width=\columnwidth]{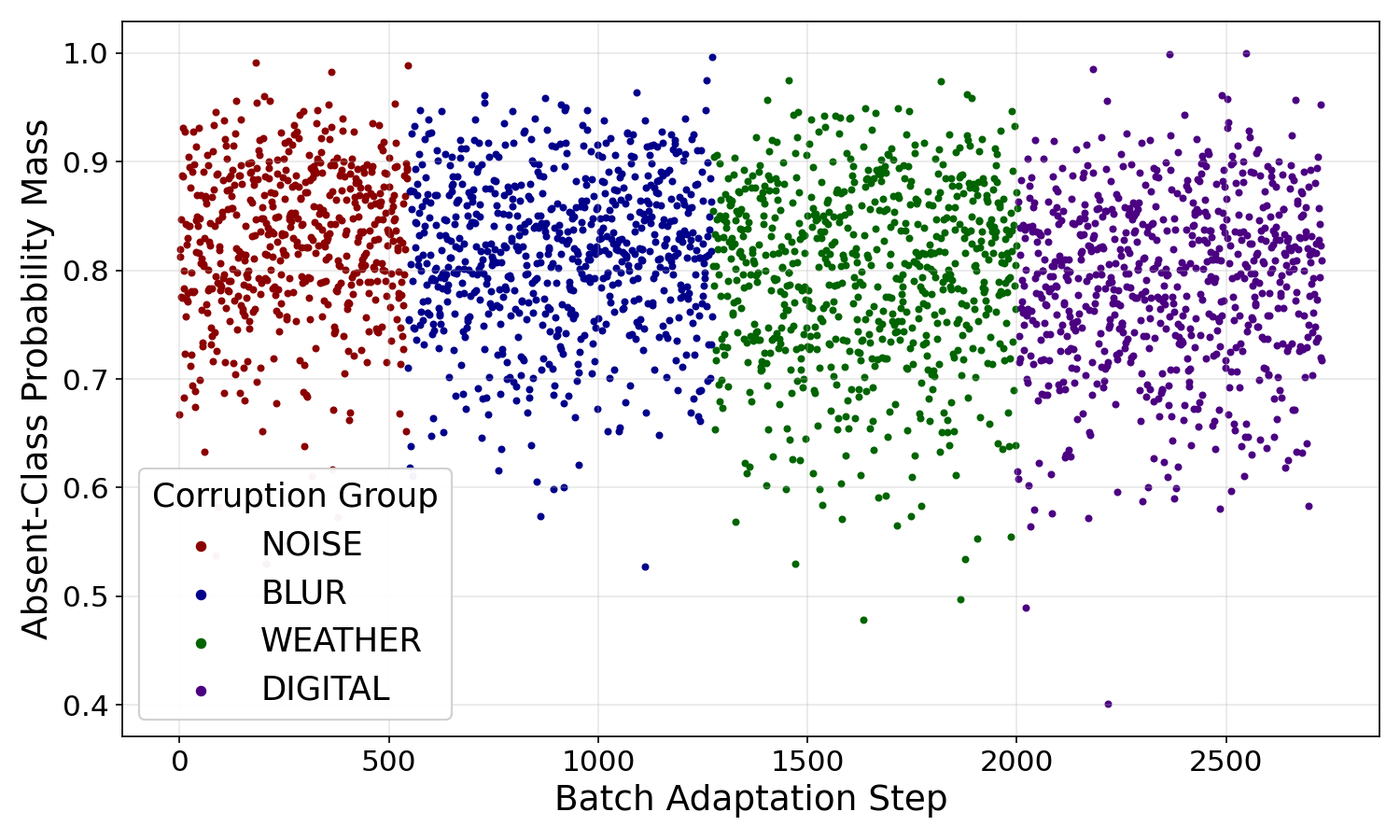}
        \caption{$\mathcal{L}_{\text{div}} = \text{True}$}
    \end{subfigure}
    \caption{Absent-class probability mass on PASCAL VOC21-C~\cite{Everingham2014ThePV} with NACLIP~\cite{Hajimiri2024PayAT} ViT-B/32, comparing adaptation without~(a) and with~(b) $\mathcal{L}_{\text{div}}$.}
    \label{fig:voc21_probability_mass}
\end{figure}

\cref{fig:voc21_probability_mass} extends the absent-class probability mass analysis to VOC21-C, which adds a background class with 26 scene element synonyms to the VOC20 vocabulary. Without $\mathcal{L}_{\text{div}}$, the model exhibits the same bimodal behavior observed on VOC20-C, swinging between stable predictions and near-collapse where almost all probability shifts to absent classes. With $\mathcal{L}_{\text{div}}$ enabled, these extreme swings are reduced and absent-class probability mass remains stable on average. This confirms that the MDIV loss generalizes to the extended vocabulary setting, where the larger and more diverse label space provides more opportunities for probability mass to leak into absent classes.

{
    \small
    \bibliographystyle{ieeenat_fullname}
    \bibliography{main}
}